\documentclass[12pt]{report}
\usepackage {graphicx}
\usepackage{authblk}

\begin{document}

\title{
New Evolutionary Computation Models and their Applications to Machine Learning\\

\textbf{PhD thesis}
}

\author{Mihai Oltean}

\affil{Department of Computer Science\\
Faculty of Mathematics and Computer Science\\
Babe\c s-Bolyai University, Kog\u alniceanu 1\\
Cluj-Napoca, 3400, Romania.\\
mihai.oltean@gmail.com\\
}

\date{December 2004}

\maketitle

\tableofcontents

\listoffigures

\listoftables

\chapter{Introduction}

This chapter serves as an introduction in the field of Machine Learning and Genetic Programming. The last section presents the goals and the achievements of the thesis.

\section{Machine Learning and Genetic Programming}

Automatic Programming is one of the most important areas of computer science research today. Hardware speed and capability has increased exponentially, but the software is years behind. The demand for software has also increased significantly, but it is still written in old-fashion: by using humans.

There are multiple problems when the work is done by humans: cost, time, quality. It is costly to pay humans, it is hard to keep them satisfied for long time, it takes a lot of time to teach and train them and the quality of their output is in most cases low (in software, mostly due to bugs).

The real advances in the human civilization appeared during the industrial revolutions. Before the first revolution most people worked in agriculture. Today, very few percent of people work in this field.

A similar revolution must appear in the computer programming field. Otherwise we will have so many people working in this field as we had in the past working in agriculture.

How people know how to write computer programs? Very simple: by learning. Can we do the same for software? Can we put the software to learn how to write software?

It seems that is possible (to some degree) and the term is called Machine Learning. It was first coined in 1959 by the first person who made a computer perform a serious learning task, namely Arthur Samuel \cite{samuel1}.

However, things are not so easy as in humans (well, truth to be said - for some humans it is impossible to learn how to write software). So far we do not have a software which can learn perfectly to write software. We have some particular cases where some programs do better than humans, but the examples are sporadic at best. Learning from experience is difficult for computer programs.

Instead of trying to simulate how humans teach humans how to write computer programs, we can simulate nature. What is the advantage of nature when solving problems? Answer: Nature is huge and can easily solve problems by brute force. It can try a huge number of possibilities and choose the best one (by a mechanism called survival).

Genetic algorithms are inspired by nature. They have random generation of solutions, they have crossover, mutation, selection and so on. However, genetic algorithms cannot simulate the nature perfectly because they run on computers which have limited resources.

Can genetic algorithms learn programming? Yes. The subfield is called Genetic Programming \cite{koza1} and is quite popular today.

In Genetic Programming, we evolve computer programs by using ideas inspired from nature.

Because Genetic Programming belongs to Machine learning, it should be able to learn. It can do that if it has a set of so called training data which is nothing but a set of pairs (input;output). The most important aspect of learning in Genetic Programming is how the fitness is computed. That measurement tells us how great is learning so far.

After training we usually use a so called test set to see if the obtained program is able to generalize on unseen data. If we still get minimum errors on test set, it means that we have obtained an intelligent program.

However, one should not imagine that things are so easy. Currently Genetic Programming has generated some good results, but in general it cannot generate large computer programs. It cannot generate an operating system. Not even an word editor, not even a game. The programmer of such system should know its limitations and should help the system to reduce the search space. It is very important to ask Genetic Programming to do what humans cannot do.

There are several questions to be answered when solving a problem using a GP/ML technique. Some of the questions are:\\

\textit{How are solutions represented in the algorithm?}\\

\textit{What search operators does the algorithm use to move in the solution space?}\\

\textit{What type of search is conducted?}\\

\textit{Is the learning supervised or unsupervised?}\\

Shortly speaking, Genetic Programming (GP) \cite{koza5,koza1} is a special sub-domain of the ML. GP individuals are computer programs evolved by using specific genetic operators. The GP search is guided by a fitness function and the learning process is usually unsupervised.

\section {Thesis structure and achievements}

This thesis describes several evolutionary models developed by the author during his PhD program.

Chapter \ref{gp_chap} provides an introduction to the field of Evolutionary Code Generation. The most important evolutionary technique addressing the problem of code generation is Genetic Programming (GP) \cite{koza1}. Several linear variants of the GP technique namely Gene Expression Programming (GEP) \cite{ferreira1}, Grammatical Evolution (GE) \cite{oneill1}, Linear Genetic Programming (LGP) \cite{brameier1} and Cartesian Genetic Programming (CGP) \cite{miller2}, are briefly described.

Chapter \ref{mep_chap} describes Multi Expression Programming (MEP), an original evolutionary 
paradigm intended for solving computationally difficult problems. MEP 
individuals are linear entities that encode complex computer programs. MEP 
chromosomes are represented in the way C or Pascal compilers 
translate mathematical expressions into machine code. A unique feature of MEP is its ability of storing multiple solutions in a chromosome. This ability proved to be crucial for improving the search process. The chapter is entirely original and is based on the paper \cite{oltean_mep}.

In chapters \ref{mep_dm}, \ref{mep_digital_circuits} MEP technique is used for 
solving certain difficult problems such as symbolic regression, 
classification, multiplexer, designing digital circuits. MEP is compared 
to Genetic Programming, Gene Expression Programming, Linear 
Genetic Programming and Cartesian Genetic Programming by using 
several well-known test problems. This chapter is entirely original and it is based on the papers \cite{oltean_mep,oltean_parity_fea,oltean_knapsack,oltean_circuits_nasa,oltean_parity_trail}.

Chapter \ref{mep_heuristics} describes the way in which MEP may be used for evolving more complex computer programs such as heuristics for the Traveling Salesman Problem and winning strategies for the Nim-like games and Tic-Tac-Toe. This chapter is entirely original and it is based on the papers \cite{oltean_tsp,oltean_nim}.

Chapter \ref{ifgp_chap} describes a new evolutionary technique called Infix Form Genetic Programming (IFGP). IFGP individuals encodes mathematical expressions in infix form. Each IFGP individual is an array of integer numbers that are translated into mathematical expressions. IFGP is used for solving several well-known symbolic regression and classification problems. IFGP is compared to standard Genetic Programming, Linear Genetic Programming and Neural Networks approaches. This chapter is entirely original and it is based on the paper \cite{oltean_ifgp}.

Chapter \ref{mslgp_chap} describes Multi Solution Linear Genetic Programming (MSLGP), an improved technique based on Linear Genetic Programming. Each MSLGP individual stores multiple solutions of the problem being solved in a single chromosome. 
MSLGP is used for solving symbolic regression problems. Numerical experiments show that Multi Solution Linear Genetic Programming performs significantly better than the standard Linear Genetic Programming for all the considered test problems. This chapter is entirely original and it is based on the paper \cite{oltean_mslgp,oltean_improving}.

Chapter \ref{eea_chap} describes two evolutionary models used for evolving evolutionary algorithms. The models are based on Multi Expression Programming and Linear Genetic Programming. The output of the programs implementing the proposed models are some full-featured evolutionary algorithms able to solve a given problem. Several evolutionary algorithms for function optimization, Traveling Salesman Problem and for the Quadratic Assignment Problem are evolved using the proposed models. This chapter is entirely original and it is based on the papers \cite{oltean_lgp_eea,oltean_mep_eea}.

Chapter \ref{nfl_chap} describes an attempt to provide a practical evidence of the No Free Lunch (NFL) theorems. NFL theorems state that all black box algorithms perform equally well over the space of all optimization problems, when averaged over all the optimization problems. An important question related to NFL is finding problems for which a given algorithm $A$ is better than another specified algorithm $B$. The problem of finding mathematical functions for which an evolutionary algorithm is better than another evolutionary algorithm is addressed in this chapter. This chapter is entirely original and it is based on the papers \cite{oltean_nfl,oltean_nfl_search}.

\section{Other ML results not included in this Thesis}

During the PhD program I participated to the development of other evolutionary paradigms and algorithms:

\textit{Traceless Genetic Programming} (TGP) \cite{oltean_tgp,oltean_tgp_class} - a Genetic Programming variant that does not store the evolved computer programs. Only the outputs of the program are stored and recombined using specific operators.

\textit{Evolutionary Cutting Algorithm} (ECA) \cite{oltean_2d} - an algorithm for the 2-dimensional cutting stock problem. 

\textit{DNA Theorem Proving} \cite{oltean_dna} - a technique for proving theorems using DNA computers.

\textit{Genetic Programming Theorem Prover} \cite{oltean_proof} - an algorithm for proving theorems using Genetic Programming.

\textit{Adaptive Representation Evolutionary Algorithm} (AREA) \cite{oltean_area} - an evolutionary model that allow dynamic representation, i.e. encodes solutions over several alphabets. The encoding alphabet is adaptive and it may be changes during the search process in order to accomodate to the fitness landscape.

\textit{Evolving Continuous Pareto Regions} \cite{oltean_c_pareto} - a technique for detecting continuous Pareto regions (when they exist).

\chapter{Genetic Programming and related techniques}\label{gp_chap}

This chapter describes Genetic Programming (GP) and several related techniques. The chapter is organized as follows: Genetic Programming is described in section \ref{gp}. Cartesian Genetic Programming is presented in section \ref{cgp}. Gene Expression Programming is described in section \ref{gep}. Linear Genetic Programming is described in section \ref{lgp}. Grammatical Evolution is described in section \ref{ge}.

\section{Genetic Programming} \label{gp}

Genetic Programming (GP) \cite{koza1,koza2,koza3} is an evolutionary technique used for breeding a population of computer programs.
Whereas the evolutionary schemes employed by GP are similar to those used by 
other techniques (such as Genetic Algorithms \cite{holland1}, Evolutionary Programming \cite{yao1}, Evolution Strategies \cite{rechenberg1}), the individual representation and the 
corresponding genetic operators are specific only to GP. Due to its 
nonlinear individual representation (GP individuals are usually represented 
as trees) GP is widely known as a technique that creates computer programs. 

Each GP individual is usually represented as a tree encoding a 
complex computer program. The genetic operators used with GP are crossover 
and mutation. The crossover operator takes two parents and generates two 
offspring by exchanging sub-trees between the parents. The mutation operator 
generates an offspring by changing a sub-tree of a parent into another 
sub-tree. 

\begin{figure}[htbp]
\centerline{\includegraphics[width=4.81in,height=2.70in]{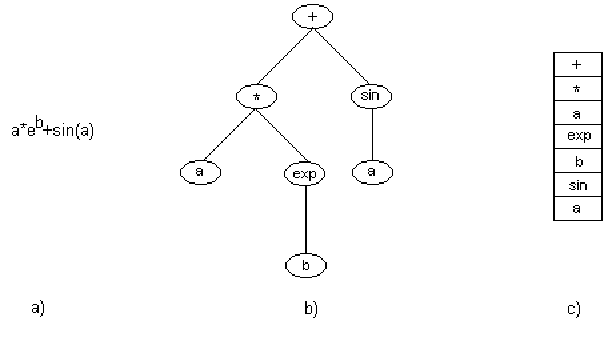}}
\caption{A mathematical expression in infix form (a), Polish form (c) and the corresponding program tree (b).}
\label{gp_tree}
\end{figure}

For efficiency reasons, each GP program tree is stored as a vector using the 
Polish form (see \cite{koza3} chapter 63). A mathematical expression in Infix and 
Polish notations and the corresponding GP program tree are depicted in 
Figure \ref{gp_tree}.

Each element in this vector contains a function or a terminal symbol. Since 
each function has a unique arity we can clearly interpret each vector that 
stores an expression in Polish notation. In this notation, a sub-tree of a 
program tree corresponds to a particular contiguous subsequence of the 
vector. When applying the crossover or the mutation operator, the exchanged 
or changed subsequences can easily be identified and manipulated.

\section{Cartesian Genetic Programming} \label{cgp}

\textit{Cartesian Genetic Programming} (CGP) \cite{miller2} is a GP technique that encodes chromosomes in graph structures 
rather than standard GP trees. The motivation for this representation is 
that the graphs are more general than the tree structures, thus allowing the 
construction of more complex computer programs \cite{miller2}.

CGP is Cartesian in the sense that the graph nodes are represented in a 
Cartesian coordinate system. This representation was chosen due to the node 
connection mechanism, which is similar to GP mechanism. A CGP node contains 
a function symbol and pointers towards nodes representing function 
parameters. Each CGP node has an output that may be used as input for 
another node.

\begin{figure}[htbp]
\centerline{\includegraphics[width=4.46in,height=2.96in]{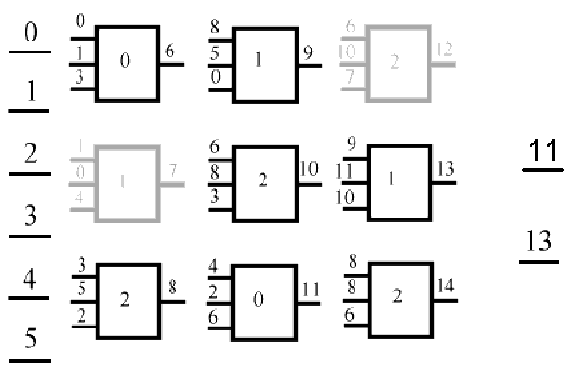}}
\caption{A CGP program with 5 inputs, 2 outputs and 3 functions (0, 1, 2 inside square nodes). The grey squares represent unconnected nodes.}
\label{cgp_program}
\end{figure}

An example of CGP program is depicted in Figure \ref{cgp_program}.

Each CGP program (graph) is defined by several parameters: number of rows 
($n_{r})$, number of columns ($n_{c})$, number of inputs, number of outputs, 
and number of functions. The nodes interconnectivity is defined as being the 
number ($l)$ of previous columns of cells that may have their outputs connected 
to a node in the current column (the primary inputs are treated as node 
outputs).

CGP chromosomes are encoded as strings by reading the graph columns top down 
and printing the input nodes and the function symbol for each node. The CGP 
chromosome depicted in Figure \ref{cgp_program} is encoded as:\\

$C$ = \textbf{0 1 3 0} 1 0 4 1 \textbf{3 5 2 2 8 5 0 1 6 8 3 2 4 2 6 0} 6 10 7 
2 \textbf{9 11 10 1 8 8 6 2 11 13}\\

Standard string genetic operators (crossover and mutation) are used within 
CGP system. Crossover may be applied without any restrictions. Mutation 
operator requires that some conditions are met. Nodes supplying the outputs 
are not fixed as they are also subject to crossover and mutation.

\section{Gene Expression Programming} \label{gep}

\textit{Gene Expression Programming} (GEP) \cite{ferreira1} uses linear chromosomes. A chromosome is composed of genes 
containing terminal and function symbols. Chromosomes are modified by 
mutation, transposition, root transposition, gene transposition, gene 
recombination, one-point and two-point recombination.

GEP genes are composed of a \textit{head} and a \textit{tail}. The head contains both functions and 
terminals symbols. The tail contains only terminal symbols.

For each problem the head length ($h$) is chosen by the user. The tail length 
(denoted $t)$ is calculated using the formula:\\

$t=(n-1)*h + 1$, \\

\noindent
where $n$ is the number of arguments of the function with more arguments.

A tree-program is translated into a GEP gene is by means of breadth-first 
parsing.

Let us consider a gene made up of symbols in the set $S$:\\

$S$ = {\{}*, /, +, -, $a$, $b${\}}. \\

In this case we have $n$ = 2. If we choose $h$ = 10 we get $t$ = 11, and the length 
of the gene is 10 + 11 = 21. Such a gene is given below:\\

$C$ = +*\textit{ab}/+\textit{aab}+\textit{ababbbababb}.\\

The expression encoded by the gene $C$ is:\\

$E=(a+b)/a*b+a$.\\

The expression $E$ represents the phenotypic transcription of a chromosome 
having $C$ as its unique gene.

Usually, a GEP gene is not entirely used for phenotypic transcription. If 
the first symbol in the gene is a terminal symbol, the expression tree 
consists of a single node. If all symbols in the head are function symbols, 
the expression tree uses all the symbols of the gene.

GEP genes may be linked by a function symbol in order to obtain a fully 
functional chromosome. In the current version of GEP, the linking functions 
for algebraic expressions are addition and multiplication. A single type of 
function is used for linking multiple genes.

This seems to be enough in some situation \cite{ferreira1}. But, generally, it is not a 
good idea to assume that the genes may be linked either by addition or by 
multiplication. If the functions {\{}+, -, *, /{\}} are used as linking 
operators then, the complexity of the problem grows substantially (since the 
problem of determining how to mixed these operators with the genes is as 
hard as the initial problem).

When solving computationally difficult problems (like automated code 
generation) one should not assume that a unique kind of function symbol 
(like \textbf{for,} \textbf{while} or \textbf{if} instructions) is necessary 
for inter-connecting different program parts.

Moreover, the success rate of GEP increases with the number of genes in the 
chromosome \cite{ferreira1}. However, after a certain value, the success rate decreases 
if the number of genes in the chromosome increases. This is because one can 
not force a complex chromosome to encode a less complex expression.

Thus, when using GEP one must be careful with the number of genes that form 
the chromosome. The number of genes in the chromosome must be somehow 
related to the complexity of the expression that one wants to discover.

According to \cite{ferreira1}, GEP performs better than standard GP for several 
particular problems.

\section{Linear Genetic Programming}\label{lgp}

\textit{Linear Genetic Programming} (LGP) \cite{brameier1,brameier2,brameier3} uses a specific linear representation of computer programs. 
Instead of the tree-based GP expressions of a functional programming 
language (like \textbf{\textit{LISP}}), programs of an imperative language 
(like \textbf{\textit{C}}) are evolved.

A LGP individual is represented by a variable-length sequence of simple 
\textbf{\textit{C}} language instructions. Instructions operate on one or 
two indexed variables (registers) $r$ or on constants $c$ from predefined sets. 
The result is assigned to a destination register, e.g. $r_{i}=r_{j}$ * 
$c$.

An example of the LGP program is the following one:

\textsf{\textbf{void}}\textsf{ LGP(}\textsf{\textbf{double}}\textsf{ v[8])}

\textsf{{\{}}

\textsf{v[0] = v[5] + 73;}

\textsf{v[7] = v[3] - 59;}

\textsf{if (v[1] $>$ 0)}

\textsf{if (v[5] $>$ 21)}

\textsf{v[4] = v[2] * v[1];}

\textsf{v[2] = v[5] + v[4];}

\textsf{v[6] = v[7] * 25;}

\textsf{v[6] = v[4] - 4;}

\textsf{v[1] = sin(v[6]);}

\textsf{if (v[0] $>$ v[1])}

\textsf{v[3] = v[5] * v[5];}

\textsf{v[7] = v[6] * 2;}

\textsf{v[5] = [7] + 115;}

\textsf{if (v[1] $<=$ v[6])}

\textsf{v[1] = sin(v[7]);}

\textsf{{\}}}

A linear genetic program can be turned into a functional representation by 
successive replacements of variables starting with the last effective 
instruction. The variation operators used here are crossover and mutation. 
By crossover, continuous sequences of instructions are selected and 
exchanged between parents. Two types of mutations are used: micro mutation 
and macro mutation. By micro mutation an operand or an operator of an 
instruction is changed. Macro mutation inserts or deletes a random 
instruction.

\section{Grammatical Evolution}\label{ge}

\textit{Grammatical Evolution }(GE) \cite{oneill1} uses Backus - Naur Form (BNF) in order to express 
computer programs. BNF is a notation that allows a computer program to 
be expressed as a grammar. 

A BNF grammar consists of terminal and non-terminal symbols. Grammar symbols 
may be re-written in other terminal and non-terminal symbols. 

Each GE individual is a variable length binary string that contains the 
necessary information for selecting a production rule from a BNF grammar in 
its \textit{codons} (groups of 8 bits).

An example from a BNF grammar is given by the following production rules:\\

$S$ ::= \textit{expr} (0)

$\vert $\textit{if}-\textit{stmt} (1)

$\vert $\textit{loop} (2)\\

These production rules state that the start symbol $S$ can be replaced 
(re-written) either by one of the non-terminals \textit{expr}, \textit{if}-\textit{stmt}, or by \textit{loop}.

The grammar is used in a generative process to construct a program by 
applying production rules, selected by the genome, beginning with the start 
symbol of the grammar.

In order to select a GE production rule, the next codon value on the genome 
is generated and placed in the following formula:\\

\textit{Rule} = \textit{Codon}{\_}\textit{Value} \textbf{MOD} \textit{Num}{\_}\textit{Rules}.\\

If the next \textit{Codon} integer value is 4, knowing that we have 3 rules to select 
from, as in the example above, we get 4 \textbf{MOD} 3 = 1.

Therefore, $S$ will be replaced with the non-terminal \textit{if-stmt}, corresponding to the 
second production rule.

Beginning from the left side of the genome codon, integer values are 
generated and used for selecting rules from the BNF grammar, until one of 
the following situations arises:

\begin{itemize}

\item[{\it (i)}]{A complete program is generated. This occurs when all the non-terminals in 
the expression being mapped, are turned into elements from the terminal set 
of the BNF grammar.}

\item[{\it (i)}]{The end of the genome is reached, in which case the \textit{wrapping} operator is invoked. 
This results in the return of the genome reading frame to the left side of 
the genome once again. The reading of the codons will then continue unless a 
higher threshold representing the maximum number of wrapping events has 
occurred during this individual mapping process.}

\end{itemize}

In the case that a threshold on the number of wrapping events is exceeded 
and the individual is still incompletely mapped, the mapping process is 
halted, and the individual is assigned the lowest possible fitness value.\\

\textbf{Example}\\

Consider the grammar:\\

$G$ = {\{}$N$, $T$, $S$, $P${\}},\\

\noindent
where the terminal set is: \\

$T$ = {\{}+, -, *, /, \textit{sin}, \textit{exp}, \textit{var}, (, ){\}}, \\

\noindent\\
and the nonterminal symbols are: \\

$N$ = {\{}\textit{expr}, \textit{op}, \textit{pre{\_}op}{\}}.\\

The start symbol is $S$ = $<$\textit{expr}$>$.\\

The production rules $P$ are:\\

$<$\textit{expr}$>$ :: $<$\textit{expr}$>$ $<$\textit{op}$>$ $<$\textit{expr}$>$ $\vert $ (0)

($<$\textit{expr}$>$ $<$\textit{op}$>$ $<$\textit{expr}$>$) $\vert $ (1)

$<$\textit{pre{\_}op}$>$ ($<$\textit{expr}$>$) $\vert $ (2)

$<$\textit{var}$>$. (3)

$<$\textit{op}$>$ ::= + $\vert $ (0)

- $\vert $ (1)

* $\vert $ (2)

/. (3)

$<$\textit{pre{\_}op}$>$ ::= \textit{sin} $\vert $ (0)

\textit{exp}. (1)\\

An example of a GE chromosome is the following:\\

$C_{GE}$ = 000000000000001000000001000000110000001000000011.\\

Translated into GE codons, the chromosome is:\\

$C_{GE}$ = 0, 2, 1, 3, 2, 3.\\

This chromosome is translated into the expression:\\

$E$ = \textit{exp}($x)$ * $x$.\\

Using the BNF grammars for specifying a chromosome provides a natural way of 
evolving programs written in programming languages whose instructions may be expressed as BNF rules.

The wrapping operator provides a very original way of translating short 
chromosomes into very long expressions. Wrapping also provides an efficient way to avoid the obtaining of invalid expressions.

The GE mapping process also has some disadvantages. Wrapping may never end 
in some situations. For instance consider the G$_{GE}$ grammar defined 
above. In these conditions the chromosome \\

$C'_{GE}$ = 0, 0, 0, 0, 0\\

\noindent
cannot be translated into a valid expression as it does not contain 
operands. To prevent infinite cycling a fixed number of wrapping occurrences 
is allowed. If this threshold is exceeded the obtained expression is 
incorrect and the corresponding individual is considered to be invalid.

Since the debate regarding the supremacy of binary encoding over integer 
encoding has not finished yet we cannot say which one is better. However, as the 
translation from binary representations to integer/real representations 
takes some time we suspect that the GE system is a little slower than other 
GP techniques that use integer representation.
GE uses a steady-state \cite{syswerda1} algorithm.

\chapter{Multi Expression Programming}\label{mep_chap}

The chapter is organized as follows: MEP algorithm is given in section \ref{MEP_algorithm}. Individual representation is described in section \ref{MEP_representation}. The way in which MEP individuals are translated in computer programs is presented in section \ref{mep1}. The search operators used in conjunction with MEP are given in section \ref{Search_operators}. The way in which MEP handles exceptions raised during the fitness assignment process is presented in section \ref{exceptions}. MEP complexity is computed in section \ref{complexity}. 

The chapter is entirely original and it is based on the papers \cite{oltean_mep,oltean_parity_fea,oltean_mep_eea,oltean_knapsack,oltean_circuits_nasa,oltean_improving}.

\section{MEP basic ideas}

\textit{Multi Expression Programming} (MEP) \cite{oltean_mep} is a GP variant that uses a linear representation of chromosomes. MEP individuals are strings of genes encoding complex computer programs.

When MEP individuals encode expressions, their representation is similar to 
the way in which compilers translate $C$ or \textit{Pascal} expressions into machine code \cite{aho1}. 
This may lead to very efficient implementation into assembler languages. The 
ability of evolving machine code (leading to very important speedups) has 
been considered by others researchers, too. For instance Nordin \cite{nordin1} evolves 
programs represented in machine code. Poli and Langdon \cite{poli1} proposed 
\textit{Sub-machine code GP}, which exploits the processor ability to perform 
some operations simultaneously. Compared to these approaches, MEP has the 
advantage that it uses a representation that is more compact, simpler, and 
independent of any programming language.

A salient MEP feature is the ability of storing multiple solutions of a 
problem in a single chromosome. Usually, the best solution is chosen for 
fitness assignment. When solving symbolic regression or classification 
problems (or any other problems for which the training set is known before 
the problem is solved) MEP has the same complexity as other techniques 
storing a single solution in a chromosome (such as GP, CGP, GEP or GE).

Evaluation of the expressions encoded into a MEP individual can be performed 
by a single parsing of the chromosome.

Offspring obtained by crossover and mutation are 
always syntactically correct MEP individuals (computer programs). Thus, no 
extra processing for repairing newly obtained individuals is needed.

\section{MEP algorithm}\label{MEP_algorithm}

Standard MEP algorithm uses steady-state evolutionary model \cite{syswerda1} as its underlying 
mechanism. 

The MEP algorithm starts by creating a random population of individuals. The 
following stpng are repeated until a given number of generations is reached: 
Two parents are selected using a standard selection procedure. The parents 
are recombined in order to obtain two offspring. The offspring are 
considered for mutation. The best offspring $O$ replaces the worst individual 
$W$ in the current population if $O$ is better than $W$. 

The variation operators ensure that the chromosome length is a constant of 
the search process. The algorithm returns as its answer the best expression 
evolved along a fixed number of generations.

The standard MEP algorithm is outlined below:\\

\textbf{Standard MEP Algorithm}\\

\textsf{S}$_{1}$\textsf{. Randomly create the initial population 
}\textsf{\textit{P}}\textsf{(0)}

\textsf{S}$_{2}$\textsf{. }\textsf{\textbf{for}}\textsf{ 
}\textsf{\textit{t}}\textsf{ = 1 }\textsf{\textbf{to}}\textsf{ 
}\textsf{\textit{Max}}\textsf{{\_}}\textsf{\textit{Generations}}\textsf{ 
}\textsf{\textbf{do}}

\textsf{S}$_{3}$\textsf{. }\textsf{\textbf{for}}\textsf{ 
}\textsf{\textit{k}}\textsf{ = 1 }\textsf{\textbf{to}}\textsf{ $\vert 
$}\textsf{\textit{P}}\textsf{(}\textsf{\textit{t}}\textsf{)$\vert $ / 2 
}\textsf{\textbf{do}}

\textsf{S}$_{4}$\textsf{. }\textsf{\textit{p}}$_{1}$\textsf{ = 
}\textsf{\textit{Select}}\textsf{(}\textsf{\textit{P}}\textsf{(}\textsf{\textit{t}}\textsf{)); 
// select one individual from the current }

\textsf{// population}

\textsf{S}$_{5}$\textsf{. }\textsf{\textit{p}}$_{2}$\textsf{ = 
}\textsf{\textit{Select}}\textsf{(}\textsf{\textit{P}}\textsf{(}\textsf{\textit{t}}\textsf{)); 
// select the second individual }

\textsf{S}$_{6}$\textsf{. }\textsf{\textit{Crossover}}\textsf{ 
(}\textsf{\textit{p}}$_{1}$\textsf{, }\textsf{\textit{p}}$_{2}$\textsf{, 
}\textsf{\textit{o}}$_{1}$\textsf{, }\textsf{\textit{o}}$_{2}$\textsf{); // 
crossover the parents p}$_{1}$\textsf{ and p}$_{2}$

\textsf{// the offspring o}$_{1}$\textsf{ and o}$_{2}$\textsf{ are obtained}

\textsf{S}$_{7}$\textsf{. }\textsf{\textit{Mutation}}\textsf{ 
(}\textsf{\textit{o}}$_{1}$\textsf{); // mutate the offspring o}$_{1}$

\textsf{S}$_{8}$\textsf{. }\textsf{\textit{Mutation}}\textsf{ 
(}\textsf{\textit{o}}$_{2}$\textsf{); // mutate the offspring o}$_{2}$

\textsf{S}$_{9}$\textsf{. }\textsf{\textbf{if}}\textsf{ 
}\textsf{\textit{Fitness}}\textsf{(}\textsf{\textit{o}}$_{1}$\textsf{) $<$ 
Fitness(}\textsf{\textit{o}}$_{2}$\textsf{)}

\textsf{S}$_{10}$\textsf{. }\textsf{\textbf{then}}\textsf{ 
}\textsf{\textbf{if}}\textsf{ 
}\textsf{\textit{Fitness}}\textsf{(}\textsf{\textit{o}}$_{1}$\textsf{) $<$ the 
fitness of the worst individual }

\textsf{in the current population}

\textsf{S}$_{11}$\textsf{. }\textsf{\textbf{then}}\textsf{ 
}\textsf{\textit{Replace}}\textsf{ the worst individual with 
}\textsf{\textit{o}}$_{1}$\textsf{;}

\textsf{S}$_{12}$\textsf{. }\textsf{\textbf{else}}\textsf{ 
}\textsf{\textbf{if}}\textsf{ 
}\textsf{\textit{Fitness}}\textsf{(}\textsf{\textit{o}}$_{2}$\textsf{) $<$ the 
fitness of the worst individual }

\textsf{in the current population}

\textsf{S}$_{13}$\textsf{. }\textsf{\textbf{then}}\textsf{ 
}\textsf{\textit{Replace}}\textsf{ the worst individual with 
}\textsf{\textit{o}}$_{2}$\textsf{;}

\textsf{S}$_{14}$\textsf{. }\textsf{\textbf{endfor}}

\textsf{S}$_{15}$\textsf{. }\textsf{\textbf{endfor }}

\section{MEP representation}\label{MEP_representation}

MEP genes are (represented by) substrings of a variable length. The number 
of genes per chromosome is constant. This number defines the length of the 
chromosome. Each gene encodes a terminal or a function symbol. A gene that 
encodes a function includes pointers towards the function arguments. 
Function arguments always have indices of lower values than the position of 
the function itself in the chromosome.

The proposed representation ensures that no cycle arises while the 
chromosome is decoded (phenotypically transcripted). According to the 
proposed representation scheme, the first symbol of the chromosome must be a 
terminal symbol. In this way, only syntactically correct programs (MEP 
individuals) are obtained.\\

\textbf{Example}\\

Consider a representation where the numbers on the left positions stand for 
gene labels. Labels do not belong to the chromosome, as they are provided 
only for explanation purposes.

For this example we use the set of functions:\\

$F$ = {\{}+, *{\}},\\

\noindent
and the set of terminals\\

$T$ = {\{}$a$, $b$, $c$, $d${\}}.\\

An example of chromosome using the sets $F$ and $T$ is given below:\\

1: $a$

2: $b$

3: + 1, 2

4: $c$

5: $d$

6: + 4, 5

7: * 3, 6\\

The maximum number of symbols in MEP chromosome is given by the formula:\\

\textit{Number{\_}of{\_}Symbols} = ($n + $1) * (\textit{Number{\_}of{\_}Genes} -- 1) + 1, \\

\noindent
where $n$ is the number of arguments of the function with the greatest number 
of arguments. 

The maximum number of effective symbols is achieved when each gene 
(excepting the first one) encodes a function symbol with the highest number 
of arguments. The minimum number of effective symbols is equal to the number 
of genes and it is achieved when all genes encode terminal symbols only.

\section{MEP phenotypic transcription. Fitness assignment}\label{mep1}

Now we are ready to describe how MEP individuals are translated into 
computer programs. This translation represents the phenotypic transcription 
of the MEP chromosomes.

Phenotypic translation is obtained by parsing the chromosome top-down. A 
terminal symbol specifies a simple expression. A function symbol specifies a 
complex expression obtained by connecting the operands specified by the 
argument positions with the current function symbol.

For instance, genes 1, 2, 4 and 5 in the previous example encode simple 
expressions formed by a single terminal symbol. These expressions are:\\

$E_{1} = a$,

$E_{2} = b$,

$E_{4} = c$,

$E_{5} = d$,\\

Gene 3 indicates the operation + on the operands located at positions 1 and 
2 of the chromosome. Therefore gene 3 encodes the expression:\\

$E_{3}=a+b$.\\

Gene 6 indicates the operation + on the operands located at positions 4 and 
5. Therefore gene 6 encodes the expression:\\

$E_{6}=c+d$.\\

Gene 7 indicates the operation * on the operands located at position 3 and 
6. Therefore gene 7 encodes the expression:\\

$E_{7}=(a+b)*(c+d)$.\\

$E_{7}$ is the expression encoded by the whole chromosome.

There is neither practical nor theoretical evidence that one of these 
expressions is better than the others. Moreover, Wolpert and McReady \cite{wolpert1,wolpert2} 
proved that we cannot use the search algorithm's behavior so far for a 
particular test function to predict its future behavior on that function. 
This is why each MEP chromosome is allowed to encode a number of expressions 
equal to the chromosome length (number of genes). The chromosome described 
above encodes the following expressions:\\

$E_{1}=a$,

$E_{2}=b$,

$E_{3}=a+b$,

$E_{4}=c$,

$E_{5}=d$,

$E_{6}=c+d$,

$E_{7}$ = ($a+b)$ * ($c+d)$.\\

The value of these expressions may be computed by reading the chromosome top 
down. Partial results are computed by dynamic programming \cite{bellman1} and are stored 
in a conventional manner.

Due to its multi expression representation, each MEP chromosome may be 
viewed as a forest of trees rather than as a single tree, which is the case 
of Genetic Programming.

As MEP chromosome encodes more than one problem solution, it is interesting 
to see how the fitness is assigned. 

The chromosome fitness is usually defined as the fitness of the best 
expression encoded by that chromosome.

For instance, if we want to solve symbolic regression problems, the fitness 
of each sub-expression $E_{i}$ may be computed using the formula:

\begin{equation}
\label{eq1}
f(E_i ) = \sum\limits_{k = 1}^n {\left| {o_{k,i} - w_k } \right|} ,
\end{equation}

where $o_{k,i}$ is the result obtained by the expression $E_{i}$ for the 
fitness case $k$ and $w_{k}$ is the targeted result for the fitness case $k$. In 
this case the fitness needs to be minimized.

The fitness of an individual is set to be equal to the lowest fitness of the 
expressions encoded in the chromosome:

\begin{equation}
\label{eq2}
f(C) = \mathop {\min }\limits_i f(E_i ).
\end{equation}

When we have to deal with other problems, we compute the fitness of each 
sub-expression encoded in the MEP chromosome. Thus, the fitness of the 
entire individual is supplied by the fitness of the best expression encoded 
in that chromosome. 

\section{MEP representation revisited}

Generally a GP chromosome encodes a single expression (computer program). 
This is also the case for GEP and GE chromosomes. By contrast, a MEP 
chromosome encodes several expressions (as it allows a multi-expression 
representation). Each of the encoded expressions may be chosen to represent 
the chromosome, i.e. to provide the phenotypic transcription of the 
chromosome. Usually, the best expression that the chromosome encodes 
supplies its phenotypic transcription (represents the chromosome).

Therefore, the MEP technique is based on a special kind of implicit 
parallelism. A chromosome usually encodes several well-defined expressions. 
The ability of MEP chromosome to encode several syntactically correct 
expressions in a chromosome is called \textit{strong implicit parallelism} (SIP).

Although, the ability of storing multiple solutions in a single chromosome 
has been suggested by others authors, too (see for instance \cite{lones1}), and 
several attempts have been made for implementing this ability in GP 
technique. For instance Handley \cite{handley1} stored the entire population of GP 
trees in a single graph. In this way a lot of memory is saved. Also, if 
partial solutions are efficiently stored, we can get a considerable speed 
up.

Linear GP \cite{brameier1} is also very suitable for storing multiple solutions in a 
single chromosome. In that case the multi expression ability is given by the 
possibility of choosing any variable as the program output.

It can be seen that the effective length of the expression may increases 
exponentially with the length of the chromosome. This is happening because 
some sub-expressions may be used more than once to build a more complex (or 
a bigger) expression. Consider, for instance, that we want to obtain a 
chromosome that encodes the expression$a^{2^n}$, and only the operators 
{\{}+, -, *, /{\}} are allowed. If we use a GEP representation the 
chromosome has to contain at least (2$^{n + 1 }$-- 1) symbols since we need 
to store 2$^{n}$ terminal symbols and (2$^{n}$ -- 1) function operators. A 
GEP chromosome that encodes the expression $E=a^{8}$ is given below:\\

$C$ = *******\textit{aaaaaaaa}.\\

A MEP chromosome uses only (3$n$ + 1) symbols for encoding the 
expression $a^{2^n}$. A MEP chromosome that encodes expression $E = a^{8}$ is 
given below:\\

1: $a$

2: * 1, 1

3: * 2, 2

4: * 3, 3\\

As a further comparison, when $n$ = 20, a GEP chromosome has to have 2097151 
symbols, while MEP needs only 61 symbols. 

MEP representation is similar to GP and CGP, in the sense that each function 
symbol provides pointers towards its parameters. Whereas both GP and CGP 
have complicated representations (trees and graphs), MEP provides an easy 
and effective way to connect (sub) parts of a computer program. Moreover, 
the motivation for MEP was to provide an individual representation close to 
the way in which C or Pascal compilers interpret mathematical expressions 
\cite{aho1}. That code is also called three addresses code or intermediary code.

Some GP techniques, like Linear GP, remove non-coding sequences of 
chromosome during the search process. As already noted \cite{brameier1} this strategy 
does not give the best results. The reason is that sometimes, a part of the 
useless genetic material has to be kept in the chromosome in order to 
maintain population diversity.

\section{Search operators}\label{Search_operators}

The search operators used within MEP algorithm are crossover and mutation. 
These search operators preserve the chromosome structure. All offspring are 
syntactically correct expressions. 

\subsection{Crossover}

By crossover two parents are selected and are recombined.

Three variants of recombination have been considered and tested within our 
MEP implementation: one-point recombination, two-point recombination and 
uniform recombination.\\

\textbf{One-point recombination}\\

One-point recombination operator in MEP representation is similar to the 
corresponding binary representation operator \cite{dumitrescu1}. One crossover point is 
randomly chosen and the parent chromosomes exchange the sequences at the 
right side of the crossover point.\\

\textbf{Example}\\

Consider the parents $C_{1}$ and $C_{2}$ given below. Choosing the crossover 
point after position 3 two offspring, $O_{1}$ and $O_{2}$ are obtained as 
given in Table \ref{mep_one_cut_point}.

\begin{table}[htbp]
\begin{center}
\caption{MEP one-point recombination.}
\begin{tabular}
{|p{55pt}|p{49pt}|p{49pt}|p{53pt}|}
\hline
\multicolumn{2}{|p{104pt}|}{Parents } & 
\multicolumn{2}{|p{103pt}|}{Offspring}  \\
\hline
$C_{1}$& 
$C_{2}$& 
$O_{1}$& 
$O_{2}$ \\
\hline
1: \textbf{\textit{b}} \par 2: \textbf{* 1, 1} \par 3: \textbf{+ 2, 1} \par 4: \textbf{\textit{a}} \par 5: \textbf{* 3, 2} \par 6: \textbf{\textit{a}} \par 7: \textbf{- 1, 4}& 
1: $a$ \par 2: $b$ \par 3: + 1, 2 \par 4: $c$ \par 5: $d$ \par 6: + 4, 5 \par 7: * 3, 6& 
1: \textbf{\textit{b}} \par 2: \textbf{* 1, 1} \par 3: \textbf{+ 2, 1} \par 4: $c$ \par 5: $d$ \par 6: + 4, 5 \par 7: * 3, 6& 
1: $a$ \par 2: $b$ \par 3: + 1, 2 \par 4: \textbf{\textit{a}} \par 5: \textbf{* 3, 2} \par 6: \textbf{\textit{a}} \par 7: \textbf{- 1, 4} \\
\hline
\end{tabular}
\end{center}
\label{mep_one_cut_point}
\end{table}

\textbf{Two-point recombination}\\

Two crossover points are randomly chosen and the chromosomes exchange 
genetic material between the crossover points.\\

\textbf{Example}\\

Let us consider the parents $C_{1}$ and $C_{2}$ given below. Suppose that the 
crossover points were chosen after positions 2 and 5. In this case the 
offspring $O_{1}$ and $O_{2}$ are obtained as given in Table \ref{mep_two_cut_point}.

\begin{table}[htbp]
\begin{center}
\caption{MEP two-point recombination.}
\begin{tabular}
{|p{55pt}|p{49pt}|p{49pt}|p{49pt}|}
\hline
\multicolumn{2}{|p{104pt}|}{Parents } & 
\multicolumn{2}{|p{98pt}|}{Offspring}  \\
\hline
$C_{1}$& 
$C_{2}$& 
$O_{1}$& 
$O_{2}$ \\
\hline
1: \textbf{\textit{b}} \par 2: \textbf{* 1, 1} \par 3: \textbf{+ 2, 1} \par 4: \textbf{\textit{a}} \par 5: \textbf{* 3, 2} \par 6: \textbf{\textit{a}} \par 7: \textbf{- 1, 4}& 
1: $a$ \par 2: $b$ \par 3: + 1, 2 \par 4: $c$ \par 5: $d$ \par 6: + 4, 5 \par 7: * 3, 6& 
1: \textbf{\textit{b}} \par 2: \textbf{* 1, 1} \par 3: + 1, 2 \par 4: $c$ \par 5: $d$ \par 6: \textbf{\textit{a}} \par 7: \textbf{- 1, 4}& 
1: $a$ \par 2: $b$ \par 3:\textbf{ + 2, 1} \par 4: \textbf{\textit{a}} \par 5: \textbf{* 3, 2} \par 6: + 4, 5 \par 7: * 3, 6 \\
\hline
\end{tabular}
\end{center}
\label{mep_two_cut_point}
\end{table}

\textbf{Uniform recombination}\\

During the process of uniform recombination, offspring genes are taken 
randomly from one parent or another.\\

\textbf{Example}\\

Let us consider the two parents $C_{1}$ and $C_{2}$ given below. The two 
offspring $O_{1}$ and $O_{2}$ are obtained by uniform recombination as 
given in Table \ref{mep_uniform}.

\begin{table}[htbp]
\caption{MEP uniform recombination.}
\label{mep_uniform}
\begin{center}
\begin{tabular}
{|p{55pt}|p{49pt}|p{49pt}|p{49pt}|}
\hline
\multicolumn{2}{|p{104pt}|}{Parents } & 
\multicolumn{2}{|p{98pt}|}{Offspring}  \\
\hline
$C_{1}$& 
$C_{2}$& 
$O_{1}$& 
$O_{2}$ \\
\hline
1: \textbf{\textit{b}} \par 2: \textbf{* 1, 1} \par 3: \textbf{+ 2, 1} \par 4: \textbf{\textit{a}} \par 5: \textbf{* 3, 2} \par 6: \textbf{\textit{a}} \par 7: \textbf{- 1, 4}& 
1: $a$ \par 2: $b$ \par 3: + 1, 2 \par 4: $c$ \par 5: $d$ \par 6: + 4, 5 \par 7: * 3, 6& 
1: $a$ \par 2: \textbf{* 1, 1} \par 3: \textbf{+ 2, 1} \par 4: $c$ \par 5: \textbf{* 3, 2} \par 6: + 4, 5 \par 7: \textbf{- 1, 4}& 
1: \textbf{\textit{b}} \par 2: $b$ \par 3: + 1, 2 \par 4: \textbf{\textit{a}} \par 5: $d$ \par 6: \textbf{\textit{a}} \par 7: * 3, 6 \\
\hline
\end{tabular}
\end{center}
\end{table}

It is easy to derive - by analogy with standard GA several recombination operators.

\subsection{Mutation}

Each symbol (terminal, function of function pointer) in the chromosome may 
be the target of the mutation operator. Some symbols in the chromosome are 
changed by mutation. To preserve the consistency of the chromosome, its 
first gene must encode a terminal symbol.

We may say that the crossover operator occurs between genes and the mutation 
operator occurs inside genes.

If the current gene encodes a terminal symbol, it may be changed into 
another terminal symbol or into a function symbol. In the later case, the 
positions indicating the function arguments are randomly generated. If the 
current gene encodes a function, the gene may be mutated into a terminal 
symbol or into another function (function symbol and pointers towards 
arguments).\\

\textbf{Example}\\

Consider the chromosome $C$ given below. If the boldfaced symbols are selected 
for mutation an offspring $O$ is obtained as given in Table \ref{mep_mutation}.

\begin{table}[htbp]
\caption{MEP mutation.}
\label{mep_mutation}
\begin{center}
\begin{tabular}
{|p{55pt}|p{56pt}|}
\hline
$C$& 
$O$ \\
\hline
1: $a$ \par 2: * 1, 1 \par 3: \textbf{\textit{b}} \par 4: * 2, 2 \par 5: $b$ \par 6: +\textbf{ 3}, 5 \par 7: $a$& 
1: $a$ \par 2: * 1, 1 \par 3: \textbf{+ 1, 2} \par 4: * 2, 2 \par 5: $b$ \par 6: \textbf{+ 1}, 5 \par 7: $a$ \\
\hline
\end{tabular}
\end{center}
\end{table}

\section{Handling exceptions within MEP}\label{exceptions}

Exceptions are special situations that interrupt the normal flow of 
expression evaluation (program execution). An example of exception is 
\textit{division by zero,} which is raised when the divisor is equal to zero.

\textit{Exception handling} is a mechanism that performs special processing when an exception is 
thrown.

Usually, GP techniques use a \textit{protected exception} handling mechanism \cite{koza1}. For instance, if a 
division by zero exception is encountered, a predefined value (for instance 
1 or the numerator) is returned.

GEP uses a different mechanism: if an individual contains an expression that 
generates an error, this individual receives the lowest fitness possible \cite{ferreira1}.

MEP uses a new and specific mechanism for handling exceptions. When an 
exception is encountered (which is always generated by a gene containing a 
function symbol), the gene that generated the exception is mutated into a 
terminal symbol. Thus, no infertile individual appears in a population.

\section{MEP complexity}\label{complexity}

Let \textit{NG} be the number of genes in a chromosome.

When solving symbolic regression, classification or any other problems for 
which the training set is known in advance (before computing the fitness), 
the fitness of an individual can be computed in O(\textit{NG}) stpng by dynamic 
programming \cite{bellman1}. In fact, a MEP chromosome needs to be read once for 
computing the fitness. 

Thus, MEP decoding process does not have a higher complexity than other GP - 
techniques that encode a single computer program in each chromosome.

\section{Conclusions}

Multi Expression Programming has been described in this chapter. A detailed description of the representation and of the fitness assignment has been given. 

A distinct feature of MEP is its ability to encode multiple solutions in the same chromosome. It has been shown that the complexity of decoding process is the same as in the case of other GP techniques encoding a single solution in the same chromosome.

\chapter{MEP for Data Mining}\label{mep_dm}

\section{Introduction}

Multi Expression Programming is used for solving several symbolic regression and classification problems. A comparison of MEP with standard GP, GEP and CGP is also provided.

The chapter is organized as follows: Symbolic regression problems are addressed in section \ref{Symbolic_regression}. For this problem MEP is compared to GP, GEP and CGP. Even Parity problems are addressed in section \ref{Even_parity}. The Multiplexer is addressed in section \ref{Multiplexer}. The way in which MEP may be used for designing digital circuits is described in sections \ref{digital_circuits} and \ref{dc_np}. In sections \ref{game_strategies} and \ref{Nim_game} MEP is used to evolve winning strategies for the Tic-Tac-Toe and Nim-like games. Another interesting application of MEP is the discovery of heuristics for NP-Complete problems. In section \ref{NP_Complete} MEP is used for evolving an heuristic for the Traveling Salesman Problem.

The chapter is based on the author's papers \cite{oltean_mep}.

\section{Symbolic regression}\label{Symbolic_regression}

In this section, MEP technique is used for solving symbolic regression problems. Results are reported in the papers \cite{oltean_mep}.

\subsection{Problem statement}

The aim of symbolic regression is to discover a function that 
satisfies a set of fitness cases.

Two well-known problems are used for testing the MEP ability of solving 
symbolic regression problems. The problems are:

The quartic polynomial \cite{koza1}. Find a mathematical expression that satisfies 
best a set of fitness cases generated by the function:\\

$f(x)=x^{4}+x^{3}+x^{2}+x$.\\

The sextic polynomial \cite{koza2}. Find a mathematical expression that satisfies 
best a set of fitness cases generated by the function:\\

$f(x)=x^{6}-2x^{4}+x^{2}$.\\

A set of 20 fitness cases was randomly generated over the interval [-1.0, 1.0] and used in the experiments performed.

\subsection{Numerical experiments}

In this section several numerical experiments with Multi Expression Programming for solving symbolic regression problems are performed.\\

\textbf{Experiment 1}\\

The success rate of the MEP algorithm is analyzed in this experiment. Success rate is computed as the number of successful runs over the total number of runs. The 
chromosome length is gradually increased. MEP algorithm parameters are given 
in Table \ref{sr1}.

\begin{table}[htbp]
\caption{Algorithm parameters for the Experiment 1.}
\label{sr1}
\begin{center}
\begin{tabular}
{|p{130pt}|p{130pt}|}
\hline
\textbf{Parameter}& 
\textbf{Value} \\
\hline
Population size& 
30 \\
\hline
Number of generations& 
50 \\
\hline
Mutation& 
2 symbols / chromosome \\
\hline
Crossover type& 
Uniform-crossover \\
\hline
Crossover probability& 
0.9 \\
\hline
Selection& 
Binary tournament \\
\hline
Terminal Set& 
$T$ = {\{}$x${\}} \\
\hline
Function Set& 
$F$ = {\{}+, -, *, /{\}} \\
\hline
\end{tabular}
\end{center}
\end{table}

\begin{figure}[htbp]
\centerline{\includegraphics[width=3.43in,height=2.93in]{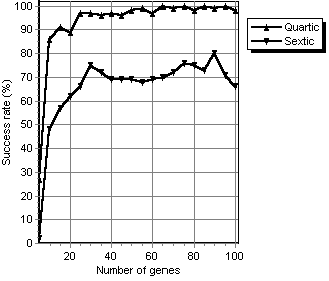}}
\caption{The relationship between the success rate and the number of symbols in a MEP chromosome. The number of symbols in chromosome varies between 5 and 100. The results are summed over 100 runs.}
\label{teza3}
\end{figure}

The success rate of the MEP algorithm depending on the number of symbols in 
the chromosome is depicted in Figure \ref{teza3}.

The success rate of the MEP algorithm increases with the chromosome length 
and never decreases towards very low values. When the search space 
(chromosome length) increases, an increased number of expressions are 
encoded by MEP chromosomes. Very large search spaces (very long chromosomes) 
are extremely beneficial for MEP technique due to its multi expression 
representation. This behavior is different from those obtained with the GP 
variants that encode a single solution in a chromosome (such as GEP). Figure 
\ref{teza3} also shows that the sextic polynomial is more difficult to solve with MEP 
(with the parameters given in Table \ref{sr1}) than the quatic polynomial.\\

\textbf{Experiment 2}\\

From Experiment 1 we may infer that for the considered problem, the MEP 
success rate never decreases to very low values as the number of genes 
increases. To obtain an experimental evidence for this assertion longer 
chromosomes are considered. We extend chromosome length up to 300 genes (898 
symbols).

\begin{figure}[htbp]
\centerline{\includegraphics[width=5.38in,height=2.94in]{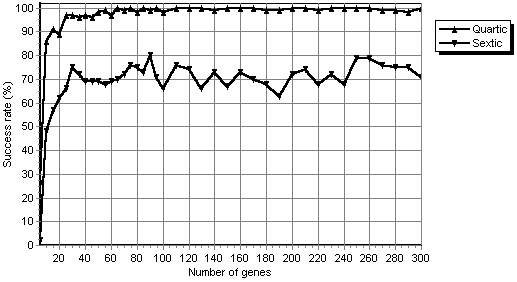}}
\caption{The relationship between the success rate and the number of symbols in a MEP chromosome. The number of symbols in chromosome varies between 5 and 300. The results are summed over 100 runs.}
\label{teza4}
\end{figure}

The success rate of MEP is depicted in Figure \ref{teza4}.

Figure \ref{teza4} shows that, when solving the quartic (sextic) 
polynomial problem, the MEP success rate, lies in the interval [90, 100] 
([60, 80]) for the chromosome length larger than 20.

One may note that after that the chromosome length becomes 10, the success 
rate never decrease more than 90{\%} (for the quartic polynomial) and never 
decrease more than 60{\%} (for the sextic polynomial). It also seems that, 
after a certain value of the chromosome length, the success rate does not 
improve significantly.\\

\textbf{Experiment 3}\\

In this experiment the relationship between the success rate and the 
population size is analyzed. Algorithm parameters for this experiment are 
given in Table \ref{sr2}.

\begin{table}[htbp]
\caption{Algorithm parameters for Experiment 3.}
\label{sr2}
\begin{center}
\begin{tabular}
{|p{130pt}|p{130pt}|}
\hline
\textbf{Parameter}& 
\textbf{Value} \\
\hline
Number of generations& 
50 \\
\hline
Chromosome length& 
10 genes \\
\hline
Mutation& 
2 symbols / chromosome \\
\hline
Crossover type& 
Uniform-crossover \\
\hline
Crossover probability& 
0.9 \\
\hline
Selection& 
Binary tournament \\
\hline
Terminal Set& 
$T$ = {\{}$x${\}} \\
\hline
Function Set& 
$F$ = {\{}+, -, *, /{\}} \\
\hline
\end{tabular}
\end{center}
\end{table}

\begin{figure}[htbp]
\centerline{\includegraphics[width=3.45in,height=2.94in]{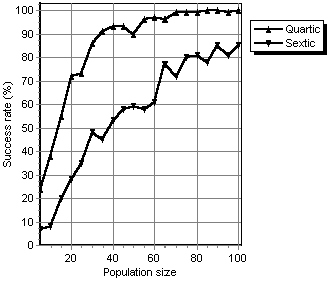}}
\caption{Success rate of the MEP algorithm. Population size varies between 5 and 100. The results are summed over 100 runs.}
\label{teza5}
\end{figure}

Experiment results are given in Figure \ref{teza5}.

For the quartic problem and for the MEP algorithm parameters given in Table 
\ref{sr2}, the optimal population size is 70 (see Figure \ref{teza5}). The corresponding success rate is 
99{\%}. A population of 100 individuals yields a success rate of 88{\%} for 
the sextic polynomial. This result suggests that even small MEP populations 
may supply very good results.\\

\textbf{Experiment 4}\\

In this experiment the relationship between the MEP success rate and the 
number of generations used in the search process is analyzed.

MEP algorithm parameters are given in Table \ref{sr3}.

\begin{table}[htbp]
\caption{Algorithm parameters for Experiment 4.}
\label{sr3}
\begin{center}
\begin{tabular}
{|p{117pt}|p{117pt}|}
\hline
\textbf{Parameter}& 
\textbf{Value} \\
\hline
Population size& 
20 \\
\hline
Chromosome length& 
12 genes \\
\hline
Mutation& 
2 genes / chromosome \\
\hline
Crossover type& 
Uniform-crossover \\
\hline
Crossover probability& 
0.9 \\
\hline
Selection& 
Binary tournament \\
\hline
Terminal Set& 
$T$ = {\{}$x${\}} \\
\hline
Function Set& 
$F$ = {\{}+, -, *, /{\}} \\
\hline
\end{tabular}
\end{center}
\end{table}

\begin{figure}[htbp]
\centerline{\includegraphics[width=3.45in,height=2.98in]{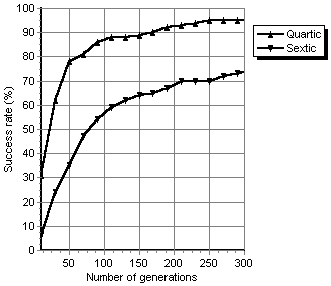}}
\caption{Relationship between the success rate of MEP algorithm and the number of generations used in the search process. The number of generations varies between 10 and 300. The results are summed over 100 runs.}
\label{teza6}
\end{figure}

Experiment results are given in Figure \ref{teza6}.

Figure \ref{teza6} shows that the success rate of the MEP algorithm rapidly increases from 34{\%}, 
respectively 8{\%} (when the number of generations is 10) up to 95{\%}, and 
74{\%} respectively (when the number of generations is 300).

\subsection{MEP vs. GEP}

In \cite{ferreira1} GEP has been used for solving the quartic polynomial based on a set 
of 10 fitness cases randomly generated over the interval [0, 20]. Several 
numerical experiments analyzing the relationship between the success rate 
and the main parameters of the GEP algorithm have been performed in \cite{ferreira1}. In 
what follows we will perform similar experiments with MEP.

The first experiment performed in \cite{ferreira1} analyses the relationship between the 
GEP chromosome length and the success rate. GEP success rate increases up to 
80{\%} (obtained when the GEP chromosome length is 30) and then decreases. 
This indicates that very long GEP chromosomes cannot encode short 
expressions efficiently. The length of the GEP chromosome must be somehow 
related to the length of the expression that must be discovered.

Using the same parameter setting (i.e. Population Size = 30, Number of 
Generations = 50; Crossover probability = 0.7, Mutations = 2 / chromosome), 
the MEP obtained a success rate of 98{\%} when the chromosome length was set 
to 20 genes (58 symbols).

The second experiment performed in \cite{ferreira1} analyses the relationship between the 
population size used by the GEP algorithm and the success rate. For a 
population of size 30, the GEP success rate reaches 80{\%}, and for a 
population of 50 individuals the GEP success rate reaches 90{\%}.

Using the same parameter setting (i.e. Number of Symbols in Chromosome = 49 
(17 MEP genes), Number of Generations = 50; Crossover probability = 0.7, 
Mutations = 2 / chromosome), MEP obtained a success rate of 99{\%} (in 99 
runs out of 100 MEP has found the correct solution) using a population of 
only 30 individuals.

In another experiment performed in \cite{ferreira1}, the relationship between the number 
of generations used by GEP and the rate of success is analysed. The success 
rate of 69{\%} was obtained by GEP when the number of generations was 70 and 
a success rate of 90{\%} was obtained only when the number of generations 
reached 500. For the considered generation range GEP success rate never 
reached 100{\%}.

Using the same parameter setting (i.e. Number of Symbols in Chromosome = 80 
(27 MEP genes), Population Size = 30; Crossover probability = 0.7, Mutations 
= 2 / chromosome), the MEP obtained a success rate of 97{\%} (in 97 runs out 
of 100 MEP has found the correct solution) using 30 generations only. This 
is an improvement (regarding the number of generations used to obtain the 
same success rate) with more than one order of magnitude.

We may conclude that for the quartic polynomial, the MEP has a higher 
success rate than GEP using the previously given parameter settings.

\subsection{MEP vs. CGP}

CGP has been used \cite{miller2} for symbolic regression of the sextic polynomial 
problem.

In this section, the MEP technique is used to solve the same problem using 
parameters settings similar to those of CGP. To provide a fair comparison, 
all experimental conditions described in \cite{miller2} are carefully reproduced for 
the MEP technique.

CGP chromosomes were characterized by the following parameters: $n_{r}$ = 1, 
$n_{c}$ = 10, $l$ = 10. MEP chromosomes are set to contain 12 genes (in addition 
MEP uses two supplementary genes for the terminal symbols {\{}1.0, $x${\}}).

MEP parameters (similar to those used by CGP) are given in Table \ref{sr4}.

\begin{table}[htbp]
\caption{Algorithm parameters for the MEP vs. CGP experiment.}
\label{sr4}
\begin{center}
\begin{tabular}
{|p{117pt}|p{117pt}|}
\hline
\textbf{Parameter}& 
\textbf{Value} \\
\hline
Chromosome length& 
12 genes \\
\hline
Mutation& 
2 genes / chromosome \\
\hline
Crossover type& 
One point crossover \\
\hline
Crossover probability& 
0.7 \\
\hline
Selection& 
Binary tournament \\
\hline
Elitism size& 
1 \\
\hline
Terminal set& 
$T$ = {\{}$x$, 1.0{\}} \\
\hline
Function set& 
$F$ = {\{}+, -, *, /{\}} \\
\hline
\end{tabular}
\end{center}
\end{table}

In the experiment with CGP a population of 10 individuals and a number of 
8000 generations have been used. We performed two experiments. In the first 
experiment, the MEP population size is set to 10 individuals and we compute 
the number of generations needed to obtain the success rate (61 {\%}) 
reported in \cite{miller2} for CGP.

When the MEP run for 800 generations, the success rate was 60{\%} (in 60 
runs (out of 100) MEP found the correct solution). Thus MEP requires 10 
times less generations than CGP to solve the same problem (the sextic 
polynomial problem in our case). This represents an improvement of one order 
of magnitude.

In the second experiment, the number of generations is kept unchanged (8000) 
and a small MEP population is used. We are interested to see which is the 
optimal population size required by MEP to solve this problem.

After several trials, we found that MEP has a success rate of 70{\%} when a 
population of 3 individuals is used and a success rate of 46{\%} when a 
population of 2 individuals is used. This means that MEP requires 3 times 
less individuals than CGP for solving the sextic polynomial problem.

\subsection{MEP vs. GP}

In \cite{koza1} GP was used for symbolic regression of the quartic polynomial 
function.

GP parameters are given in Table \ref{sr5}.

\begin{table}[htbp]
\caption{GP parameters used for solving the quartic problem.}
\label{sr5}
\begin{center}
\begin{tabular}
{|p{130pt}|p{200pt}|}
\hline
\textbf{Parameter}& 
\textbf{Value} \\
\hline
Population Size& 
500 \\
\hline
Number of generations& 
51 \\
\hline
Crossover probability& 
0.9 \\
\hline
Mutation probability& 
0 \\
\hline
Maximum tree depth & 
17 \\
\hline
Maximum initial tree depth& 
6 \\
\hline
Terminal set& 
$T$ = {\{}$x${\}} \\
\hline
Function set& 
$F$ = {\{}+, - , *, {\%}, \textit{Sin}, \textit{Cos}, \textit{Exp}, \textit{RLog}{\}} \\
\hline
\end{tabular}
\end{center}
\end{table}

It is difficult to compare MEP with GP since the experimental conditions 
were not the same. The main difficulty is related to the number of symbols 
in chromosome. While GP individuals may increase, MEP chromosomes have fixed 
length. Individuals in the initial GP population are trees having a maximum 
depth of 6 levels. The number of nodes in the largest tree containing 
symbols from $F \quad  \cup  \quad T$ and having 6 levels is 2$^{6}$ -- 1 = 63 nodes. The 
number of nodes in the largest tree containing symbols from $F \quad  \cup  \quad T$ and 
having 17 levels (maximum depth allowed for a GP tree) is 2$^{17}$ -- 1 = 
131071 nodes.

Due to this reason we cannot compare MEP and GP relying on the number of 
genes in a chromosome. Instead, we analyse different values for the MEP 
chromosome length.

\begin{figure}[htbp]
\centerline{\includegraphics[width=3.02in,height=2.97in]{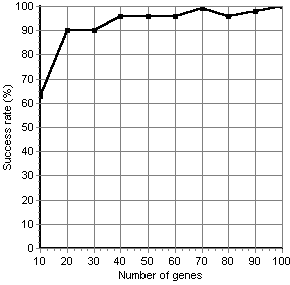}}
\caption{The relationship between the number of genes in a chromosome and the MEP success rate. The number of genes in a chromosome varies between 10 and 100. The results are averaged over 100 runs.}
\label{teza7}
\end{figure}

MEP algorithm parameters are similar to those used by GP \cite{koza1}. The results 
are depicted in Figure \ref{teza7}.

For this problem, the GP cumulative probability of success is 35{\%} (see 
\cite{koza1}). Figure \ref{teza7} shows that the lowest success rate for MEP is 
65{\%}, while the highest success rate is 100{\%} (for the considered 
chromosome length domain). Thus, MEP outperforms GP on the quartic 
polynomial problem (when the parameters given in Table \ref{sr1} are used).

\section{Conclusions}

In this chapter, MEP has been used for solving various symbolic regression problems. MEP has been compared with other Genetic Programming techniques. Numerical results indicate that MEP performs better than the compared methods. 

\chapter{Designing Digital Circuits with MEP}\label{mep_digital_circuits}

MEP is used for designing digital circuits based on the truth table. Four problems are addressed: even-parity, multiplexer, arithmetic circuits and circuits for NP-complete problems.
This chapter is entirely original and it is based on the papers \cite{oltean_parity_fea,oltean_circuits_nasa,oltean_parity_trail}.

\section{Even-parity problem}\label{Even_parity}

\subsection{Problem statement}

The Boolean even-parity function of $k$ Boolean arguments returns \textbf{T} 
(\textbf{True}) if an even number of its arguments are \textbf{T}. Otherwise 
the even-parity function returns \textbf{NIL} (\textbf{False}) \cite{koza1}.

In applying MEP to the even-parity function of $k$ arguments, the terminal set 
$T$ consists of the $k$ Boolean arguments $d_{0}$, $d_{1}$, $d_{2}$, ... $d_{k - 1}$. 
The function set $F$ consists of four two-argument primitive Boolean functions: 
AND, OR, NAND, NOR. According to \cite{koza1} the Boolean even-parity functions 
appear to be the most difficult Boolean functions to be detected via a blind 
random search.

The set of fitness cases for this problem consists of the 2$^{k}$ combinations 
of the $k$ Boolean arguments. The fitness of an MEP chromosome is the sum, over 
these 2$^{k}$ fitness cases, of the Hamming distance (error) between the 
returned value by the MEP chromosome and the correct value of the Boolean 
function. Since the standardized fitness ranges between 0 and 2$^{k}$, a 
value closer to zero is better (since the fitness is to be minimized).

\subsection{Numerical experiments}

The parameters for the numerical experiments with MEP for even-parity problems are given in Table \ref{ep1}.

\begin{table}[htbp]
\caption{The MEP algorithm parameters for the numerical experiments with even-parity problems.}
\label{ep1}
\begin{center}
\begin{tabular}
{|p{130pt}|p{210pt}|}
\hline
\textbf{Parameter}& 
\textbf{Value} \\
\hline
Number of generations& 
51 \\
\hline
Crossover type& 
Uniform \\
\hline
Crossover probability& 
0.9 \\
\hline
Mutation probability& 
0.2 \\
\hline
Terminal set& 
$T_{3}$ = {\{}$D_{0}, D_{1}, D_{2}${\}} for even-3-parity \par $T_{4}$ = {\{}$D_{0}, D_{1}, D_{2}$, $D_{3}${\}} for even-4-parity \\
\hline
Function set& 
$F$ = {\{}AND, OR, NAND, NOR{\}} \\
\hline
\end{tabular}
\end{center}
\end{table}

In order to reduce the length of the chromosome all the terminals are kept 
on the first positions of the MEP chromosomes. The selection pressure is 
also increased by using higher values (usually 10{\%} of the population 
size) for the $q$-tournament size.

Several numerical experiments with MEP have been performed for solving the 
even-3-parity and the even-4-parity problems. After several trials we have 
found that a population of 100 individuals having 300 genes was enough to 
yield a success rate of 100{\%} for the even-3-parity problem and a 
population of 400 individuals with 200 genes yielded a success rate of 
43{\%} for the even-4-parity problem. GP without Automatically Defined 
Functions has been used for solving the even-3 and even-4 parity problems 
using a population of 4000 individuals \cite{koza1}. The cumulative probability of 
success was 100{\%} for the even-3-parity problem and 42{\%} for the 
even-4-parity problem \cite{koza1}. Thus, MEP outperforms GP for the even-3 and 
even-4 parity problems with more than one order of magnitude. However, we 
already mentioned, a perfect comparison between MEP and GP cannot be drawn 
due to the incompatibility of the respective representations.

\begin{figure}[htbp]
\centerline{\includegraphics[width=3.97in,height=2.04in]{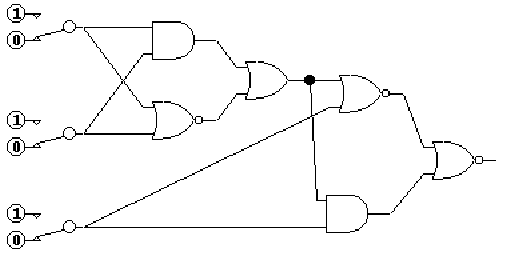}}
\caption{An evolved circuit for the even-3-parity problem.}
\label{fig:16_2}
\end{figure}

\begin{figure}[htbp]
\centerline{\includegraphics[width=5.53in,height=3.06in]{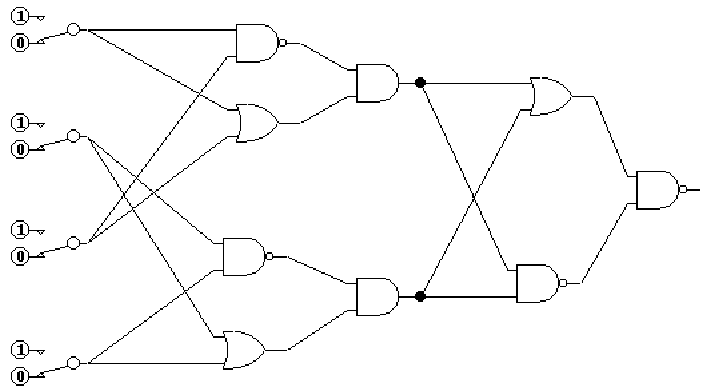}}
\caption{An evolved circuit for the even-4-parity problem.}
\label{fig:16_4}
\end{figure}

One of the evolved circuits for the even-3-parity problem is given in Figure \ref{fig:16_2} and one of the evolved circuits for the even-4-parity is given in Figure \ref{fig:16_4}.

\section{Multiplexer problem}\label{Multiplexer}

In this section, the MEP technique is used for solving the 6-multiplexer and 
the 11-multiplexer problems \cite{koza1}. Numerical experiments obtained by applying MEP to multiplexer problem are reported in \cite{oltean_mep}.

\subsection{Problem statement}

The input to the Boolean $N$-multiplexer function consists of $k$ address bits 
$a_{i}$ and 2$^{k}$ data bits $d_{i}$, where

$N=k$ + 2$^{k}$. That is, the input consists of the $k$+2$^{k}$ bits $a_{k - 1}$, 
... , $a_{1}$, $a_{0}$, $d_{2}k_{ - 1}$, ... , $d_{1}$, $d_{0}$. The value of 
the Boolean multiplexer function is the Boolean value (0 or 1) of the 
particular data bit that is singled out by the $k$ address bits of the 
multiplexer. Another way to look at the search space is that the Boolean 
multiplexer function with $k$+2$^{k}$ arguments is a particular function of 
$2^{k+2^k}$ possible Boolean functions of $k$+2$^{k}$ arguments. For example, 
when $k$=3, then $k$+2$^{k}$ = 11 and this search space is of size 22$^{11}$. That 
is, the search space is of size 2$^{2048}$, which is approximately 
10$^{616}$.

The terminal set for the 6-multiplexer problem consists of the 6 Boolean 
inputs, and for the 11-multiplexer problem consists of the 11 Boolean 
inputs. Thus, the terminal set $T$ for the 6-multiplexer is of $T$= {\{}$a_{0}$, 
$a_{1}$, $d_{0}$, $d_{1}$, ... , $d_{4}${\}} and for the 11-multiplexer is of $T$= 
{\{}$a_{0}$, $a_{1}$, $a_{2}$, $d_{0}$, $d_{1}$, ... , $d_{7}${\}}.

The function set $F$ for this problem is $F$ = {\{}AND, OR, NOT, IF{\}} taking 2, 2, 
1, and 3 arguments, respectively \cite{koza1}. The function IF returns its 3$^{rd}$ 
argument if its first argument is set to 0. Otherwise it returns its second 
argument.

There are 2$^{11}$ = 2,048 possible combinations of the 11 arguments 
$a_{0} a_{1} a_{2} d_{0} d_{1} d_{2} d_{3} d_{4} d_{5} d_{6} d_{7}$ along 
with the associated correct value of the 11-multiplexer function. For this 
particular problem, we use the entire set of 2048 combinations of arguments 
as the fitness cases for evaluating fitness.

\subsection{Numerical experiments}

Several numerical experiments with the 6-multiplexer and 11-multiplexer are performed in this section.

\subsubsection{Experiments with 6-multiplexer}

Two main statistics are of high interest: the relationship between the success 
rate and the number of genes in a MEP chromosome and the relationship 
between the success rate and the size of the population used by the MEP 
algorithm. For these experiments the parameters are given in Table \ref{mp1}.

\begin{table}[htbp]
\caption{MEP algorithm parameters for the numerical experiments with 6-multiplexer problem.}
\label{mp1}
\begin{center}
\begin{tabular}
{|p{130pt}|p{150pt}|}
\hline
\textbf{Parameter}& 
\textbf{Value} \\
\hline
Number of generations& 
51 \\
\hline
Crossover type& 
Uniform \\
\hline
Crossover probability& 
0.9 \\
\hline
Mutation probability& 
0.1 \\
\hline
Terminal set& 
$T $= {\{}$a_{0}$, $a_{1}$, $d_{0}$, $d_{1}$, ... , $d_{4}${\}} \\
\hline
Function set& 
$F $= {\{}AND, OR, NOT, IF{\}} \\
\hline
\end{tabular}
\end{center}
\end{table}

A population of 100 individuals is used when the influence of the number of 
genes is analysed and a code length of 100 genes is used when the influence 
of the population size is analysed. For reducing the chromosome length we 
keep all the terminals on the first positions of the MEP chromosomes. We 
also increased the selection pressure by using larger values (usually 10{\%} 
of the population size) for the tournament sample.

\begin{figure}[htbp]
\centerline{\includegraphics[width=6in,height=3in]{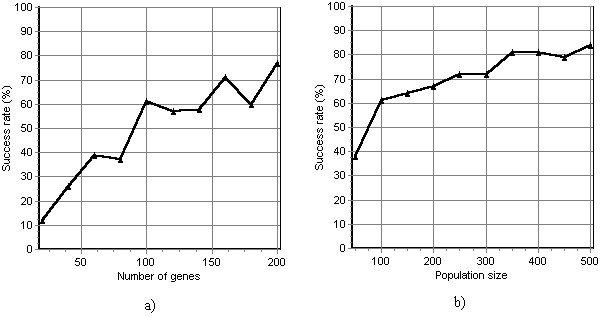}}
\caption{The success rate of the MEP algorithm for solving the 6-multiplexer problem. (a) The relationship between the success rate and the chromosome length. (b) The relationship between the success rate and the population size. Results are summed over 100 runs.}
\label{teza8}
\end{figure}

The results of these experiments are given in Figure \ref{teza8}.

Figure \ref{teza8} shows that MEP is able to solve the 6-multiplexer 
problem very well. A population of 500 individuals yields a success rate of 
84{\%}. A similar experiment using the GP technique with a population of 500 
individuals has been reported in \cite{oreilly1}. The reported probability of success 
is a little less (79,5{\%}) than the one obtained with MEP (84{\%}).

\subsubsection{Experiments with 11-multiplexer}

We also performed several experiments with the 11-multiplexer problem. We 
have used a population of 500 individuals and three values (100, 200 and 
300) for the number of genes in a MEP chromosome. In all these experiments, 
MEP was able to find a perfect solution (out of 30 runs), thus yielding a 
success rate of 3.33{\%}. When the number of genes was set to 300, the 
average of the best fitness of each run taken as a percentage of the perfect 
fitness was 91.13{\%}, with a standard deviation of 4.04. As a comparison, 
GP was not able to obtain a perfect solution by using a population of 500 
individuals and the average of the best fitness of each run taken as a 
percentage of the perfect fitness was 79.2{\%} (as reported in \cite{oreilly1}).

\section{Designing digital circuits for arithmetic functions}\label{digital_circuits}

The problem of evolving digital circuits has been intensely analyzed in the 
recent past \cite{miller1,miller3,miller4,miller5,stoica1}. A considerable effort has been spent on 
evolving very efficient (regarding the number of gates) digital circuits. J. 
Miller, one of the pioneers in the field of the evolvable digital circuits, 
used a special technique called Cartesian Genetic Programming (CGP) \cite{miller2} 
for evolving digital circuits. CGP architecture consists of a network of 
gates (placed in a grid structure) and a set of wires connecting them. For 
instance this structure has been used for evolving digital circuits for the 
multiplier problem \cite{miller5}. The results \cite{miller5} shown that CGP was able to evolve 
digital circuits better than those designed by human experts.

In this section, we use Multi Expression Programming for 
evolving digital circuits with multiple outputs. We present the way in which MEP may be efficiently applied for 
evolving digital circuits. We show the way in which multiple digital 
circuits may be stored in a single MEP chromosome and the way in which the 
fitness of this chromosome may be computed by traversing the MEP chromosome 
only once. 

Several numerical experiments are performed with MEP for evolving arithmetic 
circuits. The results show that MEP significantly outperforms CGP for the 
considered test problems.

Numerical results are reported in the papers \cite{oltean_circuits_nasa}.

\subsection{Problem statement}

The problem that we are trying to solve here may be briefly stated 
as follows:

\begin{center}
\textit{Find a digital circuit that implements a function given by its truth table.}
\end{center}

The gates that are usually used in the design of digital circuits along with 
their description are given in Table \ref{dc_tb1}.

\begin{table}[htbp]
\caption{Function set (gates) used in numerical experiments. Some functions are independent on the input (functions 0 and 1), other depend on only one of the input variables (functions 2-5), other functions depend on two input variables (functions 6-15) and the other functions depends on three input variables (functions 16-19). These functions are taken from 
\cite{miller5}.}
\label{dc_tb1}
\begin{center}
\begin{tabular}
{|p{32pt}|p{58pt}|p{36pt}|p{67pt}|}
\hline
{\#}& 
Function& 
{\#}& 
Function \\
\hline
0& 
0& 
10& 
$a \oplus b$ \\
\hline
1& 
1& 
11& 
$a \oplus \bar {b}$ \\
\hline
2& 
$a$& 
12& 
$a + b$ \\
\hline
3& 
$b$& 
13& 
$a + \bar {b}$ \\
\hline
4& 
$\bar {a}$& 
14& 
$\bar {a} + b$ \\
\hline
5& 
$\bar {b}$& 
15& 
$\bar {a} + \bar {b}$ \\
\hline
6& 
$a \cdot b$& 
16& 
$a \cdot \bar {c} + b \cdot c$ \\
\hline
7& 
$a \cdot \bar {b}$& 
17& 
$a \cdot \bar {c} + \bar {b} \cdot c$ \\
\hline
8& 
$\bar {a} \cdot b$& 
18& 
$\bar {a} \cdot \bar {c} + b \cdot c$ \\
\hline
9& 
$\bar {a} \cdot \bar {b}$& 
19& 
$\bar {a} \cdot \bar {c} + \bar {b} \cdot c$ \\
\hline
\end{tabular}
\end{center}
\end{table}

\begin{figure}[htbp]
\centerline{\includegraphics[width=3.50in,height=0.90in]{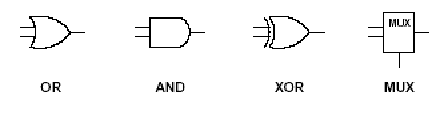}}
\caption{The symbols used to represent some of the logical gates in Table \ref{dc_tb1} (OR is function 12, AND is function 6, XOR is function 10 and MUX is function 16). In some pictures a small circle may appear on these symbols indicating the negation (inversion) of the respective results.}
\label{circuits1}
\end{figure}

The symbols used to represent some of the logical gates are given in Figure \ref{circuits1}.

The MUX gate may be also represented using 2 ANDs and 1 OR \cite{miller5}. However some 
modern devices use the MUX gate as an atomic device in that all other gates 
are synthesized using this one.

Gates may also be represented using the symbols given in Table \ref{dc_tb2}.

\begin{table}[htbp]
\caption{Representation of some functions given in Table \ref{dc_tb1}.}
\label{dc_tb2}
\begin{center}
\begin{tabular}
{|p{86pt}|p{94pt}|}
\hline
Gate& 
Representation \\
\hline
AND& 
\textbf{$ \cdot $}  \\
\hline
OR& 
+ \\
\hline
XOR& 
$ \oplus $ \\
\hline
NOT& 
- \\
\hline
\end{tabular}
\end{center}
\end{table}

\subsection{CGP for evolving digital circuits}

\begin{figure}[htbp]
\centerline{\includegraphics[width=5.06in,height=1.76in]{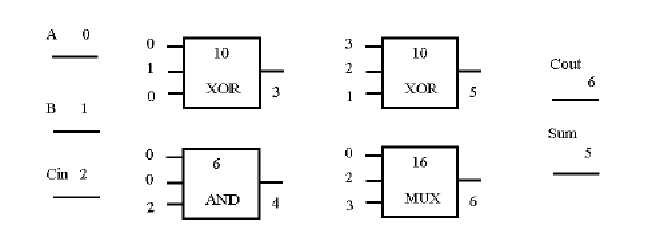}}
\caption{A Cartesian Genetic Programming program for 1-bit adder problem.}
\label{Circuits2}
\end{figure}

An example of CGP program encoding a digital circuit is depicted in Figure \ref{Circuits2}.

In Figure \ref{Circuits2}, a gate array representation of a one-bit adder is given. $A$, $B$, 
and \textit{Cin} are the binary inputs. The outputs \textit{Sum} and \textit{Cout} are the binary outputs. \textit{Sum} 
represents the sum bit of the addition of $A+B$+\textit{Cin}, and \textit{Cout} the carry bit. The 
chromosome representation of the circuit in Figure 2 is the following 
(function symbols are given in bold):

0 1 0 \textbf{10 }0 0 2 \textbf{6 }3 2 1 \textbf{10 }0 2 3 \textbf{16 }6 5.

The evolutionary algorithm used in \cite{miller5} to evolve digital circuits is a 
simple form of (1+$\lambda )$-ES \cite{dumitrescu1,miller5}, where $\lambda $ was set to 4. This 
algorithm seems to perform very well in conjunction to CGP representation. 
However, a Genetic Algorithm (GA) \cite{goldberg1} may also be used as underlying 
mechanism for CGP.

\subsection{MEP for evolving digital circuits}

In this section we describe the way in which Multi Expression Programming 
may be efficiently used for evolving digital circuits.

Each circuit has one or more inputs (denoted by \textit{NI}) and one or more outputs 
(denoted \textit{NO}). In section \ref{mep1} we presented the way in which is the fitness of a 
chromosome with a single output is computed. When multiple outputs are 
required for a problem, we have to choose \textit{NO} genes which will provide the 
desired output (it is obvious that the genes must be distinct unless the 
outputs are redundant).

In CGP, the genes providing the program's output are evolved just like all 
other genes. In MEP, the best genes in a chromosome are chosen to provide 
the program's outputs. When a single value is expected for output we simply 
choose the best gene (see section \ref{mep1}). When multiple 
genes are required as outputs we have to select those genes which minimize 
the difference between the obtained result and the expected output. 

We have to compute first the quality of a gene (sub-expression) for a given 
output:

\begin{equation}
\label{eq3}
f(E_i ,q) = \sum\limits_{k = 1}^n {\left| {o_{k,i} - w_{k,q} } \right|} ,
\end{equation}

\noindent
where $o_{k,i}$ is the obtained result by the expression (gene) $E_{i}$ for 
the fitness case $k$ and $w_{k,q}$ is the targeted result for the fitness case 
$k$ and for the output $q$. The values $f(E_{i}$, $q)$ are stored in a matrix (by using 
dynamic programming \cite{bellman1} for latter use (see formula (\ref{eq4})).

Since the fitness needs to be minimized, the quality of a MEP chromosome is 
computed by using the formula:

\begin{equation}
\label{eq4}
f(C) = \mathop {\min }\limits_{i_1 ,i_2 ,...,i_{NO} } \sum\limits_{q = 1}^{NO} {f(E_{i_q } ,q)} .
\end{equation}

In equation (\ref{eq4}) we have to choose numbers $i_{1}$, $i_{2}$, \ldots , 
$i_{NO}$ in such way to minimize the program's output. For this we shall use 
a simple heuristic which does not increase the complexity of the MEP 
decoding process: for each output $q$ (1 $ \le  \quad q \quad  \le $ \textit{NO}) we choose the gene 
$i$ that minimize the quantity $f(E_{i}$, $q)$. Thus, to an output is assigned the 
best gene (which has not been assigned before to another output). The 
selected gene will provide the value of the $q^{th}$ output.\\

\textbf{Remark}\\
Formulas (\ref{eq3}) and (\ref{eq4}) are the generalization of formulas (\ref{eq1}) and (\ref{eq2}) for the case of multiple outputs of a MEP chromosome.

The complexity of the heuristic used for assigning outputs to genes is \\

O(\textit{NG $ \cdot $ NO}) \\

where \textit{NG} is the number of genes and \textit{NO} is the number of outputs.

We may use another procedure for selecting the genes that will provide the 
problem's outputs. This procedure selects, at each step, the minimal value 
in the matrix $f(E_{i}$, $q)$ and assign the corresponding gene $i$ to its paired 
output $q$. Again, the genes already used will be excluded from the search. 
This procedure will be repeated until all outputs have been assigned to a 
gene. However, we did not used this procedure because it has a higher 
complexity -- $O$(\textit{NO}$ \cdot $\textit{log}$_{2}$(\textit{NO})$ \cdot $\textit{NG}) - than the previously 
described procedure which has the complexity $O$(\textit{NO}$ \cdot $\textit{NG}).

\subsection{Numerical experiments}

In this section, several numerical experiments with MEP for evolving digital 
circuits are performed. For this purpose several well-known test problems 
\cite{miller5} are used.

For reducing the chromosome length and for preventing input redundancy we 
keep all the terminals on the first positions of the MEP chromosomes.

For assessing the performance of the MEP algorithm three statistics are of 
high interest: 

\begin{itemize}

\item[{\it (i)}]{The relationship between the success rate and the number of genes in a MEP chromosome.}

\item[{\it (ii)}]{The relationship between the success rate and the size of the population 
used by the MEP algorithm.}

\item[{\it (iii)}]{The computation effort.}
\end{itemize}

The success rate is computed using the equation (\ref{eq5}).

\begin{equation}
\label{eq5}
Success\,rate = 
\frac{The\,number\,of\,successful\,runs}{The\,total\,number\,of\,runs}.
\end{equation}

The method used to assess the effectiveness of an algorithm has been 
suggested by Koza \cite{koza1}. It consists of calculating the number of chromosomes, 
which would have to be processed to give a certain probability of success. 
To calculate this figure one must first calculate the cumulative probability 
of success $P(M, i)$, where $M $represents the population size, and $i $the generation 
number. The value $R(z)$ represents the number of independent runs required for a 
probability of success (given by $z)$ at generation $i$. The quantity \textit{I(M, z, i) }represents 
the minimum number of chromosomes which must be processed to give a 
probability of success $z$, at generation $i$. \textit{Ns}($i)$ represents the number of successful runs at 
generation $i$, and $N_{total }$, represents the total number of runs:

The formulae are given below:

\begin{equation}
\label{eq6}
P(M,i) = \frac{Ns(i)}{N_{total} }.
\end{equation}

\begin{equation}
\label{eq7}
R(z) = ceil\left\{ {\frac{\log (1 - z)}{\log (1 - P(M,i)}} \right\}.
\end{equation}

\begin{equation}
\label{eq8}
I(M,i,z) = M \cdot R(z) \cdot i.
\end{equation}

Note that when $z$ = 1.0 the formulae are invalid (all runs successful). In the 
tables and graphs of this section $z$ takes the value 0.99.

In the numerical experiments performed in this section the number of symbols 
in a MEP chromosome is usually larger than the number of symbols in a CGP 
chromosome because in a MEP the problem's inputs are also treated as a 
normal gene and in a CGP the inputs are treated as being isolated from the 
main CGP chromosome. Thus, the number of genes in a MEP chromosome is equal 
to the number of genes in CGP chromosome + the number of problem's inputs.

\subsubsection{Two-bit multiplier: a MEP vs. CGP experiment}

The two-bit multiplier \cite{miller1} implements the binary multiplication of two 
two-bit numbers to produce a possible four-bit number. The training set for 
this problem consist of 16 fitness cases, each of them having 4 inputs and 4 
outputs.

Several experiments for evolving a circuit that implements the two-bit 
multiplier are performed. In the first experiment we want to compare the 
computation effort spent by CGP and MEP for solving this problem. Gates 6, 7 
and 10 (see Table \ref{dc_tb1}) are used in this experiment.

The parameters of CGP are given in Table \ref{dc_tb3} and the parameters of the MEP 
algorithm are given in Table \ref{dc_tb4}.

\begin{table}[htbp]
\caption{Parameters of the CGP algorithm.}
\label{dc_tb3}
\begin{center}
\begin{tabular}
{|p{122pt}|p{160pt}|}
\hline
\textbf{Parameter}& 
\textbf{Value} \\
\hline
Number of rows& 
1 \\
\hline
Number of columns& 
10 \\
\hline
Levels back& 
10 \\
\hline
Mutation & 
3 symbols / chromosome \\
\hline
Evolutionary scheme& 
(1+4) ES \\
\hline
\end{tabular}
\end{center}
\end{table}

\begin{table}[htbp]
\caption{Parameters of the MEP algorithm.}
\label{dc_tb4}
\begin{center}
\begin{tabular}
{|p{122pt}|p{160pt}|}
\hline
\textbf{Parameter}& 
\textbf{Value} \\
\hline
Code length& 
14 (10 gates + 4 inputs) \\
\hline
Crossover& 
Uniform \\
\hline
Crossover probability& 
0.9 \\
\hline
Mutation & 
3 symbols / chromosome \\
\hline
Selection& 
Binary Tournament \\
\hline
\end{tabular}
\end{center}
\end{table}

One hundred runs of 150000 generations are performed for each population 
size. Results are given in Table \ref{dc_tb5}.

\begin{table}[htbp]
\caption{Computation effort spent for evolving two-bit multipliers for different population sizes. CGP results are taken from \cite{miller5}. The differences $\Delta $ in percent considering the values of MEP as a baseline are given in the last column. Results are averaged over 100 runs.}
\label{dc_tb5}
\begin{center}
\begin{tabular}
{|p{62pt}|p{110pt}|p{110pt}|p{40pt}|}
\hline
\textbf{Population size}& 
\textbf{Cartesian Genetic Programming}& 
\textbf{Multi Expression Programming}& 
$\Delta $ \\
\hline
2& 
148808& 
53352& 
178.91 \\
\hline
3& 
115224& 
111600& 
3.24 \\
\hline
4& 
81608& 
54300& 
50.29 \\
\hline
5& 
126015& 
59000& 
113.58 \\
\hline
6& 
100824& 
68850& 
46.44 \\
\hline
7& 
100821& 
39424& 
155.73 \\
\hline
8& 
96032& 
44160& 
117.46 \\
\hline
9& 
108036& 
70272& 
53.73 \\
\hline
10& 
108090& 
28910& 
273.88 \\
\hline
12& 
115248& 
25536& 
351.31 \\
\hline
14& 
117698& 
26544& 
343.40 \\
\hline
16& 
120080& 
21216& 
465.98 \\
\hline
18& 
145854& 
17820& 
718.48 \\
\hline
20& 
120100& 
21120& 
468.65 \\
\hline
25& 
180075& 
23500& 
666.27 \\
\hline
30& 
162180& 
19440& 
734.25 \\
\hline
40& 
216360& 
16000& 
1252.25 \\
\hline
50& 
225250& 
13250& 
1600.00 \\
\hline
\end{tabular}
\end{center}
\end{table}

MEP outperforms CGP for all considered population sizes as shown in Table \ref{dc_tb5}. The differences range from 3.24{\%} (for 3 individuals in the population) up to 1600{\%} (for 50 individuals in the population). From 
this experiment we also may infer that large populations are better for MEP 
than for CGP. The computational effort decrease for MEP as the population 
size is increased.

We are also interested in computing the relationship between the success 
rate and the chromosome length and the population size. 

The number of genes in each MEP chromosome is set to 20 genes when the 
relationship between the success rate and the population size is analyzed. 
When the relationship between the success rate and the population size is 
analyzed a population consisting of 20 MEP chromosomes is used. Gates 6, 7 
and 10 are used in this experiment. Other MEP parameters are given in Table \ref{dc_tb4}.

\begin{figure}[htbp]
\centerline{\includegraphics[width=6.17in,height=3.21in]{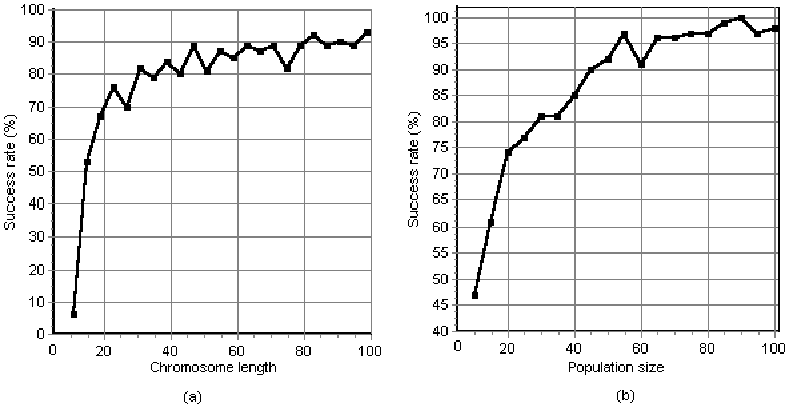}}
\caption{The relationship between the success rate of the MEP algorithm and (a) number of genes in a chromosome, (b) the number of individuals in population. Results are averaged over 100 runs.}
\label{Circuits3}
\end{figure}

Results are depicted in Figure \ref{Circuits3}.

Figure \ref{Circuits3} shows that MEP is able to find a correct digital 
circuit in multiple runs. A population consisting of 90 individuals with 20 
genes yields a success rate of 100{\%} (see Figure \ref{Circuits3}(b)) and a population 
with 20 individuals with 85 genes yields a success rate of 92{\%} (see 
Figure \ref{Circuits3}(a)).

From Figure \ref{Circuits3}(a) we may infer that larger MEP chromosomes are better than 
the shorter ones. The minimum number of gates for this circuit is 7. This 
number has been achieved by Miller during his numerical experiments (see 
\cite{miller5}). A MEP chromosome implementing Miller's digital circuit has 11 genes 
(the actual digital circuit + 4 input genes). From Figure \ref{Circuits3}(a) we can see 
that, for a MEP chromosome with 11 genes, only 6 correct solutions have been 
evolved. As the chromosome length increases the number of correct solutions 
evolved by also increases. If the chromosome has more than 21 genes the 
success rate never decreases below than 70{\%}. 

Even if the chromosome length is larger than the minimum required (11 genes) 
the evolved solutions usually have no more than 14 genes. This is due to the 
multi expression ability of MEP which acts like a provider of variable 
length chromosomes \cite{oltean_mep}. The length of the obtained circuits could be 
reduced by adding another feature to our MEP algorithm. This feature has 
been suggested by C. Coello in \cite{coello1} and it consists of a multiobjective 
fitness function. The first objective is to minimize the differences between 
the expected output and the actual output (see formulas (\ref{eq3}) and (\ref{eq4})). The 
second objective is to minimize the number of gates used by the digital 
circuit. Note that he first objective is more important than the second one. 
We also have to modify the algorithm. Instead of stopping the MEP algorithm 
when an optimal solution (regarding the first objective) is found we 
continue to run the program until a fixed number of generations have been 
elapsed. In this way we hope that also the number of gates (the second 
objective) will be minimized.

\subsubsection{Two-bit adder with carry}

A more complex situation is the \textit{Two Bit Adder with Carry} problem \cite{miller5}. The circuit implementing this 
problem adds 5 bits (two numbers represented using 2 bits each and a carry 
bit) and gives a three-bit number representing the output. 

The training set consists of 32 fitness cases with 5 inputs and 3 outputs. 

The relationship between the success rate and the chromosome length and the 
population size is analyzed for this problem.

When the relationship between the success rate and the population size is 
analyzed the number of genes in each MEP chromosome is set to 20 genes. When 
the relationship between the success rate and the population size is 
analyzed a population consisting of 20 MEP chromosomes is used. Gates 10 and 
16 (see Table \ref{dc_tb1}) are used in this experiment (as indicated in \cite{miller5}). Other 
MEP parameters are given in Table \ref{dc_tb2}.

\begin{figure}[htbp]
\centerline{\includegraphics[width=6.18in,height=3.20in]{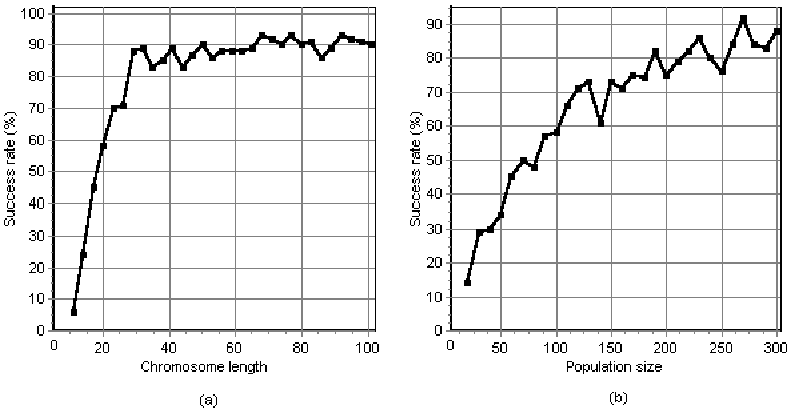}}
\caption{The relationship between the success rate of the MEP algorithm and (a) number of genes in a chromosome, (b) the number of individuals in population. Results are averaged over 100 runs.}
\label{Circuits4}
\end{figure}

Results are depicted in Figure \ref{Circuits4}.

Figure \ref{Circuits4} shows that MEP is able solve this problem very well. 
When the number of genes in a MEP chromosome is larger than 30 in more than 
80 cases (out of 100) MEP was able to find a perfect solution (see Figure 
\ref{Circuits4}(a)). After this value, the success rate does not increase significantly. A 
population with 270 individuals yields over 90 (out of 100) successful runs 
(see Figure \ref{Circuits4}(b)).

This problem is more difficult than the two-bit multiplier even if we used a 
smaller function set (functions 10 and 16) that the set used for the 
multiplier (function 6, 7 and 10).

\subsubsection{Two-bit adder}

The circuit implementing the \textit{N-Bit Adder }problem adds two numbers represented using $N$ 
bits each and gives a ($N$ + 1)-bit number representing the output.

The training set for this problem consists of 16 fitness cases with 4 inputs 
and 3 outputs.

For this problem the relationship between the success rate and the 
chromosome length and the population size is analyzed.

When the relationship between the success rate and the population size is 
analyzed the number of genes in each MEP chromosome is set to 12 genes. When 
the relationship between the success rate and the chromosome length is 
analyzed a population consisting of 100 MEP chromosomes is used. 

Gates 0 to 9 (see Table \ref{dc_tb1}) are used in this experiment. Other MEP parameters 
are given in Table \ref{dc_tb6}.

\begin{table}[htbp]
\caption{Parameters of the MEP algorithm for evolving digital circuits.}
\label{dc_tb6}
\begin{center}
\begin{tabular}
{|p{122pt}|p{160pt}|}
\hline
\textbf{Parameter}& 
\textbf{Value} \\
\hline
Crossover type& 
Uniform \\
\hline
Crossover probability& 
0.9 \\
\hline
Mutation & 
2 symbols / chromosome \\
\hline
Selection& 
Binary Tournament \\
\hline
\end{tabular}
\end{center}
\end{table}

For reducing the chromosome length and for preventing input redundancy we 
keep all terminals on the first positions of the MEP chromosomes.

\begin{figure}[htbp]
\centerline{\includegraphics[width=6.13in,height=3.21in]{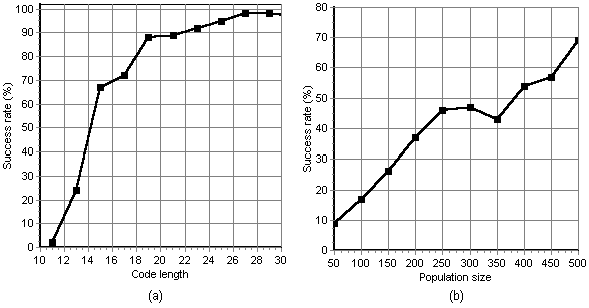}}
\caption{The relationship between the success rate of the MEP algorithm and (a) number of genes in chromosome, (b) the number of individuals in population. Results are averaged over 100 runs.}
\label{isca2}
\end{figure}

Results are depicted in Figure \ref{isca2}

Figure \ref{isca2} shows that MEP is able solve this problem very well. 
When the number of genes in a MEP chromosome is 27 in more than 98 cases 
(out of 100) MEP was able to find a perfect solution (see Figure \ref{isca2}(a)). 
After this value, the success rate does not increase significantly. A 
population with 500 individuals yields over 69 (out of 100) successful runs 
(see Figure \ref{isca2}(b)). The success rate for this problem may be increased by 
reducing the set of function symbols to an optimal set. 

From Figure \ref{isca2}(a) we may infer that larger MEP chromosomes are better than 
the shorter ones. The minimum number of gates for this circuit is 7. A MEP 
chromosome implementing the optimal digital circuit has 11 genes (the actual 
digital circuit + 4 genes storing the inputs). From Figure \ref{isca2}(a) we can see 
that, for a MEP chromosome with 11 genes, only 2 correct solutions have been 
evolved. As the chromosome length increases the number of correct solutions 
evolved by also increases. If the chromosome has more than 21 genes the 
success rate never decreases below than 89{\%}.

\subsubsection{Three-bit adder}\label{treiadd}

The training set for this problem consists of 64 fitness cases with 6 inputs 
and 4 outputs.

Due to the increased size of the training set we analyze in this section 
only the cumulative probability of success and the computation effort 
over 100 independent runs. 

Gates 0 to 9 (see Table \ref{dc_tb1}) are used in this experiment. Other MEP parameters 
are given in Table \ref{dc_tb7}.

\begin{table}[htbp]
\caption{Parameters of the MEP algorithm for solving the 3-Bit Adder problem.}
\label{dc_tb7}
\begin{center}
\begin{tabular}
{|p{122pt}|p{121pt}|}
\hline
\textbf{Parameter}& 
\textbf{Value} \\
\hline
Population size& 
2000 \\
\hline
Code Length& 
30 genes \\
\hline
Crossover type& 
Uniform \\
\hline
Crossover probability& 
0.9 \\
\hline
Mutation & 
2 symbols / chromosome \\
\hline
Selection& 
Binary Tournament \\
\hline
\end{tabular}
\end{center}
\end{table}

\begin{figure}[htbp]
\centerline{\includegraphics[width=3.58in,height=2.97in]{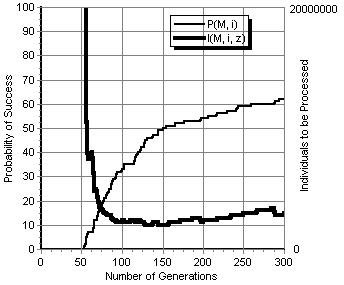}}
\caption{The cumulative probability of success and the number of individuals to be processed in order to obtained a solution with 99{\%} probability. Results are averaged over 100 runs.}
\label{isca3}
\end{figure}

Results are depicted in Figure \ref{isca3}.

Figure \ref{isca3} shows that MEP is able solve this problem very well. 
In 62 cases (out of 100) MEP was able to produce a perfect solution. The 
minimum number of individuals required to be processed in order to obtain a 
solution with a 99{\%} probability is 2030000. This number was obtained at 
generation 145. The shortest evolved circuit for this problem contains 12 
gates.

\subsubsection{Four-bit adder: preliminary results}

The training set for this problem consists of 256 fitness cases, each of 
them having 8 inputs and 5 outputs. Due to the increased complexity of this 
problem we performed only 30 independent runs using a population of 5000 
individuals having 60 genes each. The number of generations was set to 1000.

In 24 (out of 30) runs MEP was able to find a perfect solution. The shortest 
evolved digital circuit contains 19 gates. Further numerical experiments 
will be focused on evolving more efficient circuits for this problem.

\subsubsection{Two-bit multiplier}

The two-bit multiplier circuit implements the binary multiplication of 
two $N$-bit numbers to produce a possible 2 * $N$-bit number. 

The training set for this problem consists of 16 fitness cases, each of them 
having 4 inputs and 4 outputs.

Several experiments for evolving a circuit that implements the two-bit 
multiplier are performed. Since the problem has a reduced computational 
complexity we perform a detailed analysis by computing the relationship 
between the success rate and the code length and the population size.

The number of genes in each MEP chromosome is set to 14 genes when the 
relationship between the success rate and the population size is analyzed. 
When the relationship between the success rate and the chromosome length is 
analyzed a population consisting of 50 MEP chromosomes is used. Gates 0 to 9 
(see Table \ref{dc_tb1}) are used in this experiment. Other MEP parameters are given in 
Table \ref{dc_tb2}.

\begin{figure}[htbp]
\centerline{\includegraphics[width=6.13in,height=3.23in]{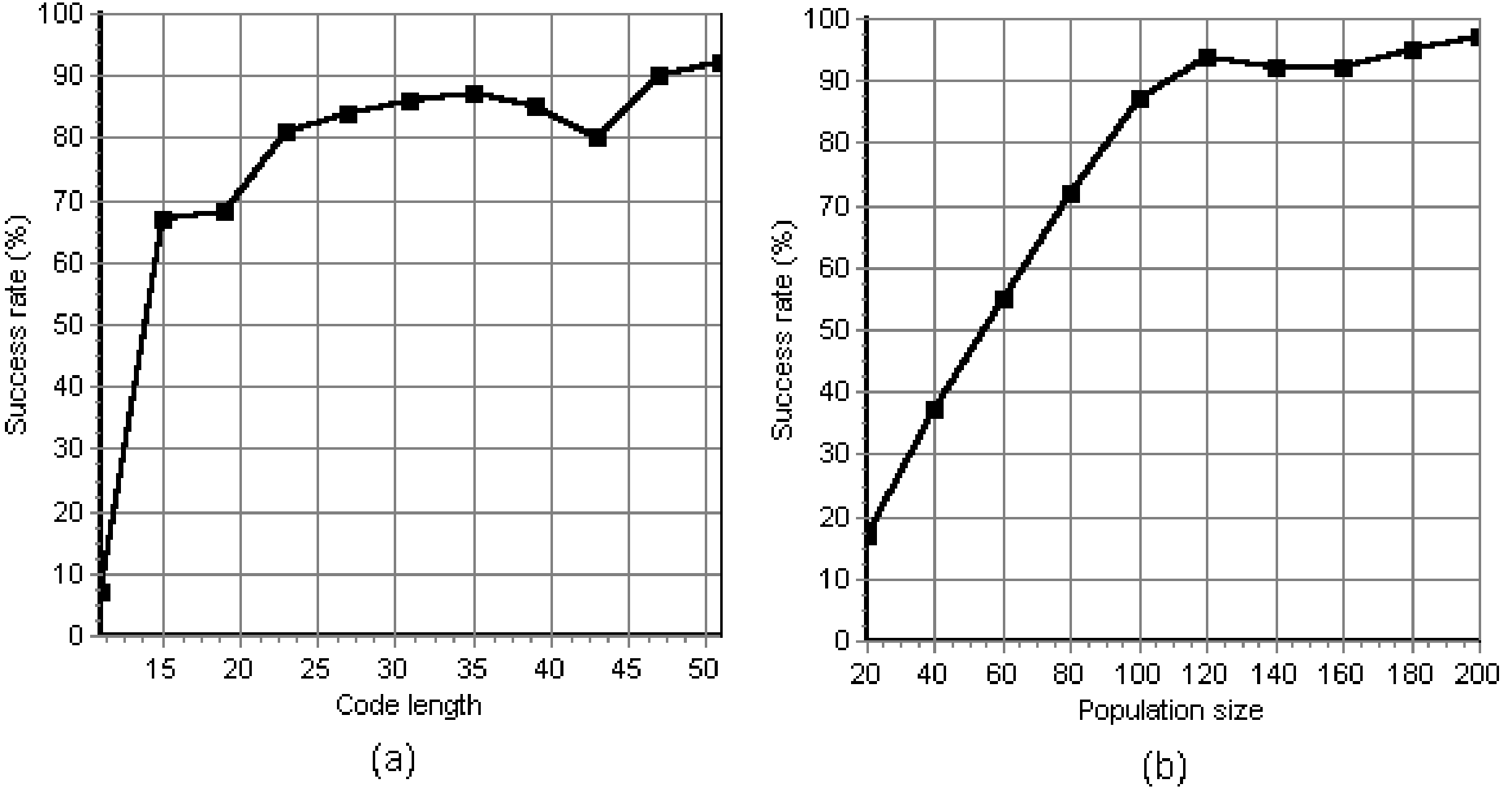}}
\caption{The relationship between the success rate of the MEP algorithm and (a) number of genes in a chromosome, (b) the number of individuals in population. Results are averaged over 100 runs.}
\label{isca4}
\end{figure}

Results are depicted in Figure \ref{isca4}.

Figure \ref{isca4} shows that MEP is able to find a correct digital 
circuit in many runs. A population consisting of 120 individuals with 14 
genes yields a success rate of 94{\%} (see Figure \ref{isca4}(b)) and a population 
with 50 individuals with 47 genes yields a success rate of 90{\%} (see 
Figure \ref{isca4}(a)). We can see that the success rate increase with more than 
60{\%} from MEP chromosomes with 11 genes to MEP chromosomes with 15 genes. 
The minimum number of gates required by this circuit is 7 and it has been 
obtained in most of the runs.

\subsubsection{3-Bit multiplier}

The training set for this problem consists of 64 fitness cases, each of them 
having 6 inputs and 6 outputs.

This problem turned out to be more difficult than the corresponding 3-Bit 
Adder (see section \ref{treiadd}). Using a population of 10000 individuals each 
having 100 genes we have obtained only 20 perfect solutions (out of 50 
independent runs). The shortest evolved digital circuit contains 35 gates. 
Further research will be focused on evolving more efficient circuits for 
this problem.

\section{Designing digital circuits for NP-Complete problems}\label{dc_np}

MEP is used for evolving digital circuits for a well-known 
NP-Complete \cite{garey1} problem: the knapsack (subset sum) problem. Numerical results are reported in \cite{oltean_knapsack}.

Since this problem is NP-Complete we cannot realistically expect to find a 
polynomial-time algorithm for it. Instead, we have to speed-up the existing 
techniques in order to reduce the time needed to obtain a solution. A 
possibility for speeding-up the algorithms for this problem is to implement 
them in assembly language. This could lead sometimes to improvements of over 
two orders of magnitude. Another possibility is to design and build a 
special hardware dedicated to that problem. This approach could lead to 
significantly improvements of the running time. Due to this reason we have 
chosen to design, by the means of evolution, digital circuits for several 
instances of the knapsack problem. 

\subsection{Evolving circuits for the knapsack problem}

The knapsack problem may also be used as benchmarking problem for the 
evolutionary techniques which design electronic circuits. The main advantage 
of the knapsack problem is its scalability: increasing the number of inputs 
leads to more and more complicated circuits. The results show that MEP 
performs very well for all the considered test problems.

The knapsack problem is a well-known NP-Complete problem \cite{garey1}. No 
polynomial-time algorithm is known for this problem.

Instead of designing a heuristic for this problem we will try to evolve 
digital circuits which will provide the answer for a given input.

In the experiments performed in this section the set $M$ consists of several 
integer numbers from the set of consecutive integers starting with 1. For 
instance if the base set is {\{}1, 2, 3, 4, 5, 6, 7{\}} then $M$ may be {\{}2, 
5, 6{\}}. We will try to evolve a digital circuit that is able t provide the 
correct answer for all subsets $M$ of the base set.

The input for this problem is a sequence of bits. A value of 1 in position 
$k$ means that the integer number $k$ belongs to the set $M$, otherwise the number 
$k$ does not belong to the set $M$. 

For instance consider the consecutive integer numbers starting with 1 and 
ending with 7. The string 0100110 encodes the set $M$ = {\{}2, 5, 6{\}}. The 
number 1, 3, 4 and 7 do not belong to $M$ since the corresponding positions are 
0. The possible subsets of $M$ instance have the sum 2, 5, 6, 7, 8, 11 or 13. 
In our approach, the target sum is fixed and we are asking if is there a 
subset of given sum.

The number of training instances for this problem depends on the number of 
consecutive integers used as base for $M$. If we use numbers 1, 2 and 3, we 
have 2$^{3}$ = 8 training instances. If we use number 1, 2, 3, 4, 5, 6 and 
7, we have 2$^{7}$ = 128 training instances. In this case, whichever subset 
$M$ of {\{}1,\ldots ,7{\}} will be presented to the evolved circuit we have to 
obtain a binary answer whether the target sum $k$ may or not be obtained from a 
subset of $M$.

\subsection{Numerical experiments}

In this section several numerical experiments for evolving digital circuits 
for the knapsack problem are performed. The general parameters of the MEP 
algorithm are given in Table \ref{dc_np1}. Since different instances of the problem 
being solved will have different degrees of difficulty we will use different 
population sizes, number of genes in a chromosome and number of generations 
for each instance. Particular parameters are given in Table \ref{dc_np2}.

\begin{table}[htbp]
\caption{General parameters of the MEP algorithm for evolving digital circuits.}
\label{dc_np1}
\begin{center}
\begin{tabular}
{|p{130pt}|p{230pt}|}
\hline
\textbf{Parameter}& 
\textbf{Value} \\
\hline
Crossover probability& 
0.9 \\
\hline
Crossover type& 
Uniform \\
\hline
Mutations& 
5 / chromosome \\
\hline
Function set& 
Gates 0 to 9  \\
\hline
Terminal set& 
Problem inputs \\
\hline
Selection& 
Binary Tournament \\
\hline
\end{tabular}
\end{center}
\end{table}

\begin{table}[htbp]
\caption{Particular parameters of the MEP algorithm for different instances of the knapsack problem. In the second column the base set of numbers is given for each instance. In the third column the target sum is given.}
\label{dc_np2}
\begin{center}
\begin{tabular}
{|p{10pt}|p{40pt}|p{30pt}|p{60pt}|p{50pt}|p{50pt}|p{60pt}|}
\hline
{\#}& 
Set of numbers& 
Sum& 
Number of fitness cases& 
Population size& 
Number of genes& 
Number of generations \\
\hline
1& 
{\{}1\ldots 4{\}}& 
5& 
16& 
20& 
10& 
51 \\
\hline
2& 
{\{}1\ldots 5{\}}& 
7& 
32& 
100& 
30& 
101 \\
\hline
3& 
{\{}1\ldots 6{\}}& 
10& 
64& 
500& 
50& 
101 \\
\hline
4& 
{\{}1\ldots 7{\}}& 
14& 
128& 
1000& 
100& 
201 \\
\hline
\end{tabular}
\end{center}
\end{table}

Experimental results are given in Table \ref{dc_np3}. We are interested in computing 
the number of successful runs and the number of gates in the shortest 
evolved circuit.

\begin{table}[htbp]
\caption{Results obtained by MEP for the considered test problems. 100 independent runs have been performed for all problems.}
\label{dc_np3}
\begin{center}
\begin{tabular}
{|p{10pt}|p{50pt}|p{40pt}|p{100pt}|p{100pt}|}
\hline
{\#}& 
Set of numbers& 
Sum& 
Successful runs& 
Number of gates in  \par the shortest circuit \\
\hline
1& 
{\{}1\ldots 4{\}}& 
5& 
39 out of 100& 
3 \\
\hline
2& 
{\{}1\ldots 5{\}}& 
7& 
31 out of 100& 
5 \\
\hline
3& 
{\{}1\ldots 6{\}}& 
10& 
10 out of 100& 
11 \\
\hline
4& 
{\{}1\ldots 7{\}}& 
14& 
7 out of 100& 
21 \\
\hline
\end{tabular}
\end{center}
\end{table}

Table \ref{dc_np3} shows that MEP successfully found at least a solution 
for the considered test problems. The difficulty of evolving a digital 
circuit for this problem increases with the number of inputs of the problem. 
Only 20 individuals are required to obtain 39 solutions (out of 100 runs) 
for the instance with 4 inputs. In return, 1000 individuals (50 times more) 
are required to obtain 10 perfect solutions (out of 100 independent runs) 
for the instance with 7 inputs. Also the size of the evolved circuits 
increases with the number of problem inputs. However, due to the reduced 
number of runs we cannot be sure that we have obtained the optimal circuits. 
Additional experiments are required in this respect.

Due to the NP-Completeness of the problem it is expected that the number of 
gates in the shortest circuit to increase exponentially with the number of 
inputs.

\section{Conclusions and Further Work}

In this section, Multi Expression Programming has been used for evolving digital circuits. It has been shown the way in which multiple digital circuits may be encoded in the same chromosome and the way in which MEP chromosomes are read only once for computing their quality. There was no human input about how the circuits should be designed, just a measurement of the degree to which a given circuit achieves the desired response.

Several numerical experiments for evolving digital circuits have been performed. The circuits evolved during the numerical experiments are for the 2 and 3-bit Multiplier, the 2, 3 and 4-bit Adder problems, even-parity, NP-complete problems and multiplexers. 

These problems are well-known benchmark instances used for assessing the performance of the algorithms evolving digital circuits.

Further numerical experiments with Multi Expression Programming will be focused on evolving digital circuits for other interesting problems.

\chapter{MEP for Evolving Algorithms and Game Strategies}\label{mep_heuristics}

\section{Discovering game strategies}\label{game_strategies}

In this section we investigate the application of MEP technique for 
discovering game strategies. This chapter is entirely original and it is based on the papers \cite{oltean_mep,oltean_tsp,oltean_nim}.

Koza \cite{koza1} suggested that GP can be applied to discover game strategy. The 
game-playing strategy may be viewed as a computer program that takes the 
information about the game as its input and produces a move as output. 

The available information may be an explicit history of previous moves or an 
implicit history of previous moves in the form of a current state of game 
(e.g. the position of each piece on the chess board) \cite{koza1}.

Tic-tac-toe (TTT, or naughts and crosses) is a game with simple rules, but 
complex enough to illustrate the ability of MEP to discover game strategy.

\subsection{TTT game description}

In Tic-Tac-Toe there are two players and a 3 $\times $ 3 grid. Initially the 
grid is empty. Each player moves in turn by placing a marker in an open 
square. By convention, the first player's marker is "X" and the second 
player's marker is "0".

The player that put three markers of his type ("X" for the first player 
and "0" for the second player) in a row is declared the winner. 

The game is over when one of the players wins or all squares are marked and 
no player wins. In the second case, the game ends with draw (none of the 
players win). Enumerating the game tree shows that the second player can 
obtain at least a draw.

A well-known evolutionary algorithm that evolves game strategy has been 
proposed in \cite{chellapilla1}. This algorithm will be reviewed in the next section.

\subsection{Chellapilla's approach of TTT}

In \cite{chellapilla1} Evolutionary Programming has been used in order to obtain a good 
strategy (that never loses) for the Tic-Tac-Toe game. A strategy is encoded 
in a neural network. A population of strategies encoded by neural networks 
is evolved. 

Each network receives a board pattern as input and yields a move as output. 
The aim is to store in a neural network the function that gives the quality 
of a configuration. When a configuration is presented to the network, the 
network output (supplies) the next move.

Each neural network has an input layer with 9 nodes, an output layer with 9 
nodes, and a hidden layer with a variable number of nodes.

Fogel's algorithm starts with a random population of 50 neural networks. For 
each network the number of nodes from the hidden layer is randomly chosen 
with a uniform distribution over integers 1...10. The initial weighted 
connection strengths and bias terms are randomly distributed according to a 
uniform distribution ranging over [-0.5, 0.5].

From each parent a single offspring is obtained by mutation. Mutation 
operator affects the hidden layer structure, weight connections and bias 
terms.

Each strategy encoded in a neural network was played 32 times against a 
heuristic rule base procedure.

The payoff function has several values corresponding to winning, loosing and 
draw.

The best individuals from a generation are retained to form the next 
generation.

The process is evolved for 800 generations. According to \cite{chellapilla1}, the best 
obtained neural network is able to play to win or draw with a perfect play 
strategy. 

\subsection{MEP approach of TTT}

In this section we illustrate the use of MEP to discover an unbeatable play 
strategy for Tic-Tac-Toe.

We are searching for a mathematical function $F$ that gives the quality of each 
game configuration. Using this function the best configurations that can be 
reached in one move from the current configuration, is selected. Therefore 
function $F$ supplies the move to be performed for each game configuration.

Function $F$ evaluating each game configuration is represented as a MEP 
chromosome. The best expression encoded by a chromosome is chosen to be the 
game strategy of that chromosome.

Without any loose of generality we may allow MEP strategy to be the first 
player in each game.

All expressions in the chromosome are considered in the fitness assignment 
process. Each expression is evaluated using an "all-possibilities" 
procedure. This procedure executes all moves that are possible for the 
second player. The fitness of an expression $E$ is the number of games that the 
strategy encoded by the expression $E$ loses. Obviously the fitness has to be 
minimized.

Let us denote by \\

$P$ = ($p_{0}$, $p_{1}$\ldots $p_{8})$\\

a game configuration. 

Each $p_{k}$ describes the states "X", "0" or an empty square. In our 
experiments the set {\{}5, -5, 2{\}} has been used for representing the 
symbols "X", "0" and the empty square. 

\begin{figure}[htbp]
\centerline{\includegraphics[width=0.99in,height=1.01in]{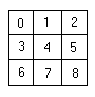}}
\caption{Game board linearized representation.}
\label{teza9}
\end{figure}

The game board has been linearized by scanning board squares from up to down 
and from left to right. Thus the squares in the first line have indices 0, 1 
and 2, etc. (see Figure \ref{teza9}).

Algorithm parameters are given in Table \ref{g1}.

\begin{table}[htbp]
\caption{Algorithm parameters for TTT game.}
\label{g1}
\begin{center}
\begin{tabular}
{|p{117pt}|p{117pt}|}
\hline
\textbf{Parameter}& 
\textbf{Value} \\
\hline
Population size& 
50 \\
\hline
Chromosome length& 
50 genes \\
\hline
Mutation probability& 
0.05 \\
\hline
Crossover type& 
Two-point-crossover \\
\hline
Selection& 
Binary tournament \\
\hline
Elitism size& 
1 \\
\hline
Terminal set& 
$T$ = {\{}$p_{0}$, $p_{1}$,\ldots , $p_{8}${\}} \\
\hline
Function set& 
$F$ = {\{}+, -, *, /{\}} \\
\hline
\end{tabular}
\end{center}
\end{table}

The MEP algorithm is able to evolve a \textit{perfect}, non-loosing, game strategy in 11 
generations. This process requires less than 10 seconds when an Intel 
Pentium 3 processor at 1GHz is used.

\begin{figure}[htbp]
\centerline{\includegraphics[width=3.34in,height=3.63in]{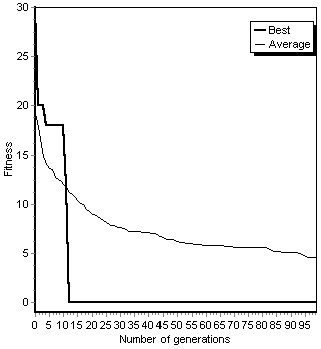}}
\caption{Fitness of the best individual in the best runs and the 
average fitness of the best individuals over all runs. The results are taken 
over 30 runs.}
\label{teza10}
\end{figure}

In Figure \ref{teza10}, the fitness of the best individual in the best run and average 
fitness of the best individuals over all runs are depicted.

Figure \ref{teza10} shows that an individual representing a non-loosing 
strategy appears in the population at generation 14.

Some functions evolved by the MEP algorithm are given below:\\

$F_{1}(P)$ = 
(($p_{4}-p_{5}$-($p_{6}+p_{5}))$*$p_{8}+p_{4}$*$p_{3})$*($p_{4}-p_{7})$, \\

$F_{2}(P)$ = 
$p_{2}$-($p_{8}$*$p_{7}-p_{4})-p_{7}$-($p_{2}$*$p_{5})$, \\

$F_{3}(P)$ = 
($p_{4}$*$p_{1}+p_{2})$*$p_{7}$-($p_{1}-p_{2}+p_{7}$*$p_{5})$-($p_{8}$-($p_{3}$*$p_{5}))$. \\

These functions do not force the win when it is possible, but they never 
lose. This is a consequence of fitness assignment process. However, the proposed 
technique can also generate a function that forces the win whenever it is 
possible. 

It is not easy to compare this result with the result obtained by 
Chellapilla and Fogel \cite{chellapilla1} as the experiment conditions were not the same. In 
\cite{chellapilla1} evolved strategies play against a heuristic procedure, but here MEP 
formulas play against an all-moves procedure. Population size was the same 
(50 individuals). Individual sizes are difficult to be compared. All MEP 
individual have the same size: 148 symbols. Neural network's sizes used in 
\cite{chellapilla1} are variable since the hidden layer contains a variable number of nodes. 
If the number of nodes in the hidden layer is 9, then the size of the neural 
network (biases + connection weights) is 9 * 9 + 9 * 9 * 9 + 9 * 9 = 324.

MEP approach seems to be faster as MEP was able to discover a non-losing 
strategy in no more than 17 generations. As noted in \cite{chellapilla1} neural network 
approach requires 800 generations.

\subsubsection{A TTT heuristic vs. MEP approach} 

A good heuristic for Tic-Tac-Toe is described in what follows: \\

S1. \textit{If a winning move is available make that move, else}

S2. \textit{If a winning move is available for the opponent, move to block it, else}

S3. \textit{If a move of the opponent that leads to two winning ways is available, block that move, else}

S4. \textit{If the board center is available, move in the board center},

S5. \textit{If one ore more corners of the table are available, move in one of them, else}

S6. \textit{Move randomly in an available square}.\\

This heuristic performs well on most of the game positions. However, by 
applying one of the formulas evolved by the MEP algorithm some benefits are 
obtained:

\begin{itemize}

\item{easy implementation in programming languages,}

\item{MEP evolved formula is a faster algorithm than the previously shown 
heuristic.}

\end{itemize}

\subsubsection{Applying MEP for generating complex game strategies}

Using the \textit{all-possibilities} technique (a backtracking procedure that 
plays all the moves for the second player) allows us to compute the absolute 
quality of a game position. 

For complex games a different fitness assignment technique is needed since 
the moves for the second player can not be simulated by an 
\textit{all-possibilities} procedure (as the number of moves that needs to be 
simulated is too large).

One fitness assignment possibility is to use a heuristic procedure that acts 
as the second player. But there are several difficulties related to this 
approach. If the heuristic is very good it is possible that none of evolved 
strategy could ever beat the heuristic. If the heuristic procedure plays as 
a novice then many evolved strategy could beat the heuristic from the 
earlier stages of the search process. In the last case fitness is not 
correctly assigned to population members and thus we can not perform a 
correct selection. 

A good heuristic must play on several levels of complexity. At the beginning 
of the search process the heuristic procedure must play at easier levels. As 
the search process advances, the level of difficulty of the heuristic 
procedure must increases.

However, for complex games such a procedure is difficult to implement.

Another possibility is to search for a game strategy using a coevolutionary 
algorithm \cite{chellapilla1}. This approach seems to offer the most spectacular results. In 
this case, MEP population must develop intelligent behavior based only on 
internal competition. 

\section{Evolving winning strategies for Nim-like games}\label{Nim_game}

In this section, we propose an evolutionary approach for computing the winning 
strategy for \textit{Nim}-like games. The proposed approach is is entirely original and it is reported in the paper \cite{oltean_nim}.

\subsection{Introduction}

\textbf{\textit{Nim}} is one of the older two-person games known today. 
Whereas the standard approaches for determining winning strategies for \textit{Nim} are 
based on the Grundy-Sprague theory \cite{berlekamp1,conway1}, 
this problem can be solved using other techniques. For instance, the first 
winning strategy for this game was proposed in 1901 by L.C. Bouton from the 
Harvard University. The Bouton's solution is based on computing the \textit{xor} sum of 
the numbers of objects in each heap. In other words Bouton computed a 
relation between the current state of the game and the player which has a 
winning strategy if it is his/her turn to move.

In this section, we propose an evolutionary approach for computing the winning 
strategy for \textit{Nim}-like games. The proposed approach is based on Multi Expression 
Programming. The idea is to find a mathematical relation (an expression) 
between the current game state and the winner of the game (assuming that 
both players do not make wrong moves). The searched expression should 
contain some mathematical operators (such as +, -, *, \textbf{\textit{div}}, 
\textbf{\textit{mod}}, \textbf{\textit{and}}, \textbf{\textit{or}}, 
\textbf{\textit{not}}, \textbf{\textit{xor}}) and some operands (encoding 
the current game state).

It is widely known \cite{berlekamp1,gardner1} that a winning 
strategy is based on separation of the game's states in two types of 
positions: $P$-positions (advantage to the previous player) and $N$-positions 
(advantage to the next player). Our aim is to find a formula that is able to 
detect whether a given game position belongs to $P$-positions or to 
$N$-positions. Our formula has to return 0 if the given position is a 
$P$-position and a nonzero value otherwise. That could be easily assimilated to 
a symbolic regression \cite{koza1} or a classification task. It is 
well-known that machine learning techniques (such as Neural Networks or 
Evolutionary Algorithms \cite{goldberg1} are very suitable for solving this 
kind of problems. However, the proposed approach is different from the 
classical approaches mainly because the $P$ and $N$-positions are usually 
difficult to be identified for a new game. Instead we propose an approach 
that checks $P$ and $N$-position during the traversing of the game tree.

This theory can be easily extended for other games that share several 
properties with the \textit{Nim} game (i.e. games for which the winning strategy is 
based on $P$ and $N$-positions).

The problem of finding $N$ and $P$-positions could be also viewed as a 
classification task with two classes. However, we do not use this approach 
because in this case it is required to know the class ($P$ or $N)$ for each game 
position.

The results presented in this section enter in the class of human-competitive 
results produced by an artificial machine. According to \cite{koza1,koza3} a result produced by an artificial machine is considered intelligent if it is equal or better than a result that was accepted as a 
new scientific result at the time when it was published in a peer-reviewed 
scientific. A list with other human-competitive results produced by the 
Genetic Programming can be found in \cite{koza3}.

\subsection{Basics on Nim game}

\textit{Nim} is one of the oldest and most engaging of all two-person mathematical games 
known today \cite{berlekamp1,conway1}. The name and the 
complete theory of the game were invented by the professor Charles Leonard 
Bouton from Harvard University about 100 years ago.

Players take turns removing objects (counters, pebbles, coins, pieces of 
paper) from heaps (piles, rows, boxes), but only from one heap at a time. In 
the normal convention the player who removes the last object wins.

The usual practice in impartial games is to call a hot position 
($N$-position - advantage to the next player, i.e. the one who is about to make 
a move) and a cold one ($P$-position - advantage to the previous player, i.e. 
the one who has just made a move).

In 1930, R. P. Sprague and P. M. Grundy developed a theory of impartial 
games in which \textbf{\textit{Nim}} played a most important role. According 
to the Sprague-Grundy theory every position in an impartial game can be 
assigned a Grundy number which makes it equivalent to a 
\textbf{\textit{Nim}} heap of that size. The Grundy number of a position is 
variously known as its \textit{Nim-heap} or \textit{nimber} for short \cite{berlekamp1,conway1}.

A P-position for the \textbf{\textit{Nim}} game is given by the equation:

$x_{1}$ \textbf{\textit{xor}} $x_{2}$ \textbf{\textit{xor}} \ldots 
\textbf{\textit{xor}} $x_{n}$ = 0,

\noindent
where $n$ is the number of heaps, $x_{i}$ is the number of objects in the 
$i^{th}$ heap and \textbf{\textit{xor}} acts as the \textbf{modulo} 2 
operator.

A variant of the \textbf{\textit{Nim}} game, also analyzed in this section, is 
the one in which players may remove no more than $k$ objects from a heap. In 
this variant a $P$-position is characterized by the equation:

($x_{1}$ \textbf{\textit{mod}} $k)$ \textbf{\textit{xor}} ($x_{2}$ 
\textbf{\textit{mod}} $k)$ \textbf{\textit{xor}} \ldots \textbf{\textit{xor}} 
($x_{n}$ \textbf{\textit{mod}} $k)$ = 0,

\noindent
where the equation parameters have been previously explained.

Due to the way of computing $P$-position we shall call this game 
\textbf{\textit{NimModK}}.

\subsection{Fitness assignment process}

The procedure used for computing the quality of a chromosome 
is described in this section.

Even if this problem could be easily handled as a classification problem 
(based on a set of fitness cases), we do not use this approach since for the 
new games it is difficult to find which the $P$-positions and $N$-positions are. 
Instead we employ an approach based on the traversing the game tree. Each 
nod in this tree is a game configuration (state).

There are three theorems that run the winning strategy for the 
\textbf{\textit{Nim}} game \cite{berlekamp1}:

\begin{itemize}

\item[{\it (i)}]{Any move applied to a $P$-position turns the game into a $N$-position.}

\item[{\it (ii)}]{There is at least one move that turns the game from a $N$-position into a 
$P$-position.}

\item[{\it (iii)}]{The final position (when the game is over) is a $P$-position.}

\end{itemize}

The value of the expression encoded into a MEP chromosome is computed for 
each game state. If the obtained value is 0, the corresponding game state is 
considered as being a $P$-position, otherwise the configuration is considered 
as being a $N$-position.

The fitness of a chromosome is equal to the number of violations of the 
above described rule that arises in a game tree. Thus, if the current 
formula (chromosome) indicates that the game state encoded into a node of 
the game tree is a $P$-position and (the same current formula indicates that) 
all the game states encoded in the offspring nodes are also $P$-positions means 
that we have a violation of the rule $b)$.

Since we do not want to have violations of the previously described rule, 
our chromosome must have the fitness equal to zero. This means that the 
fitness has to be minimized.

For a better understanding of the fitness assignment process we provide an 
example where we shall compute by hand the fitness of a chromosome. 

\begin{figure}[htbp]
\centerline{\includegraphics[width=2.52in,height=1.95in]{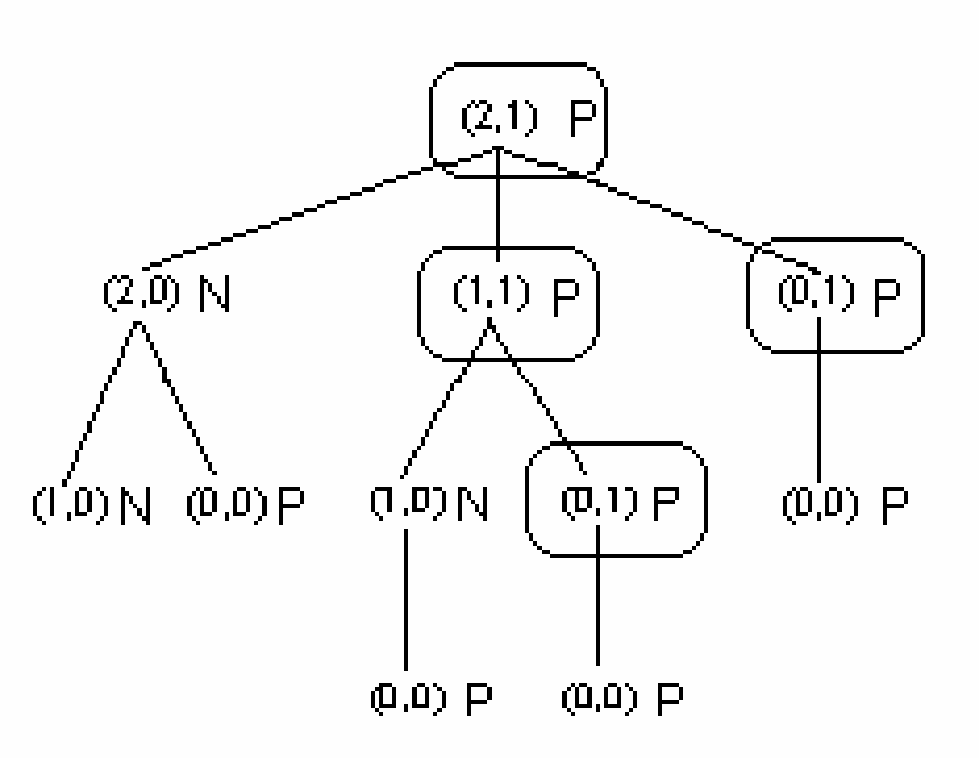}}
\caption{The game tree for a Nim game that starts with the configuration (2, 1). At the right side of each game configuration is printed the configuration' state ($P$-position or $N$-position) as computed by the formula $E=a_{1} - a_{1}$*$a_{2}$. The configurations that violate one of the three rules described above are encircled.}
\label{nim1}
\end{figure}

Consider the game state (2,1), and a MEP chromosome encoding the expression 
$E=a_{1} - a_{2}$*$a_{1}$. The game tree of the \textbf{\textit{Nim}} game 
is given in Figure \ref{nim1}.

Figure \ref{nim1} shows that the fitness of a MEP chromosome encoding 
the formula $E=a_{1} - a_{2}$*$a_{1}$ is four (there are four violations 
of the winning strategy rules).

\subsection{Numerical experiments}

Several numerical for evolving winning strategies for \textit{Nim}-like games are 
performed in this section.

The purpose of these experiments is to evolve a formula capable to 
distinguish between a $N$-position and a $P$-position for the 
\textbf{\textit{Nim}} game. We shall analyze the relationships between the 
success rate and the population size, the chromosome length and the number 
of generations used during the search process. 

In all the experiments it is considered the following configuration for the 
\textbf{\textit{Nim}} game: (4, 4, 4, 4). This configuration has been chosen 
in order to have a small computational time. However, this configuration has 
proved to be enough for evolving a winning strategy.

The total number of game configurations is 70 (which can be obtained either 
by counting nodes in the game tree or by using the formula of combinations 
with repetitions). Two permutations of the same configuration are not 
considered different.\\

\textbf{Remark}\\
The success rate is computed by using the formula:

\[
\mbox{Success}\,\mbox{rate} = \frac{\mbox{the number of successful 
}\,\mbox{runs}}{\mbox{the total number of runs}}.
\]

\textbf{Experiment 1}\\

In the first experiment the relationship between the population size and the 
success rate is analyzed. MEP algorithm parameters are given in Table \ref{nim_tb1}.

\begin{table}[htbp]
\caption{MEP algorithm parameters for Experiment 1.}
\label{nim_tb1}
\begin{center}
\begin{tabular}
{|p{125pt}|p{220pt}|}
\hline
\textbf{Parameter}& 
\textbf{Value} \\
\hline
Chromosome length& 
15 genes \\
\hline
Number of generations& 
100 \\
\hline
Crossover probability& 
0.9 \\
\hline
Mutations& 
2 mutations / chromosome \\
\hline
Selection strategy& 
binary tournament \\
\hline
Terminal set& 
$T_{Nim}$ = {\{}$n$, $a_{1}$, $a_{2}$, \ldots , $a_{n}${\}}. \\
\hline
Function set& 
$F $= {\{}+, -, *, \textbf{\textit{div}}, \textbf{\textit{mod}}, \textbf{\textit{and}}, \textbf{\textit{not}}, \textbf{\textit{xor}}, \textbf{\textit{or}}{\}} \\
\hline
\end{tabular}
\end{center}
\end{table}

\begin{figure}[htbp]
\centerline{\includegraphics[width=3.02in,height=2.49in]{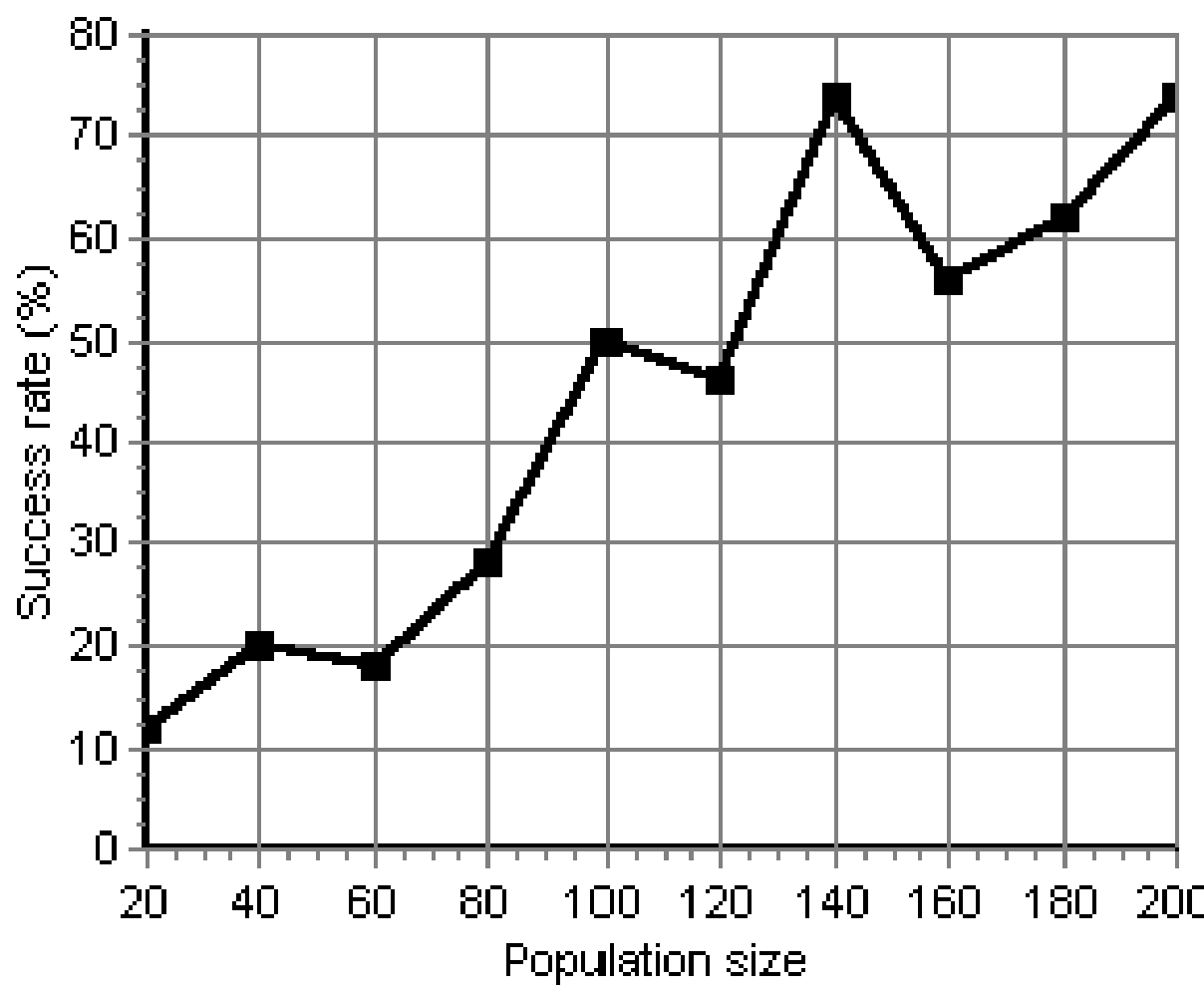}}
\caption{The relationship between the population size and the rate of success. The results are averaged over 50 runs. The population size varies between 20 and 200.}
\label{nim2}
\end{figure}

The results of this experiment are depicted in Figure \ref{nim2}.

Figure \ref{nim2} shows that the success rate increases as the 
population size increases. The highest value - 37 successful runs (out of 
50) - is obtained with a population containing 140 individuals. Even a 
population with 20 individuals is able to yield 6 successful runs (out of 
50).\\

\textbf{Experiment 2}\\

In the second experiment the relationship between the number of generations 
and the success rate is analyzed. MEP algorithm parameters are given in 
Table \ref{nim_tb2}.

\begin{table}[htbp]
\caption{MEP algorithm parameters for Experiment 2.}
\label{nim_tb2}
\begin{center}
\begin{tabular}
{|p{125pt}|p{220pt}|}
\hline
\textbf{Parameter}& 
\textbf{Value} \\
\hline
Chromosome length& 
15 genes \\
\hline
Population Size& 
100 individuals \\
\hline
Crossover probability& 
0.9 \\
\hline
Mutations& 
2 mutations / chromosome \\
\hline
Selection strategy& 
binary tournament \\
\hline
Terminal set& 
$T_{Nim}$ = {\{}$n$, $a_{1}$, $a_{2}$, \ldots , $a_{n}${\}}. \\
\hline
Function set& 
$F $= {\{}+, -, *, \textbf{div}, \textbf{mod}, \textbf{and}, \textbf{not}, \textbf{xor}, \textbf{or}{\}} \\
\hline
\end{tabular}
\end{center}
\end{table}

\begin{figure}[htbp]
\centerline{\includegraphics[width=3.01in,height=2.50in]{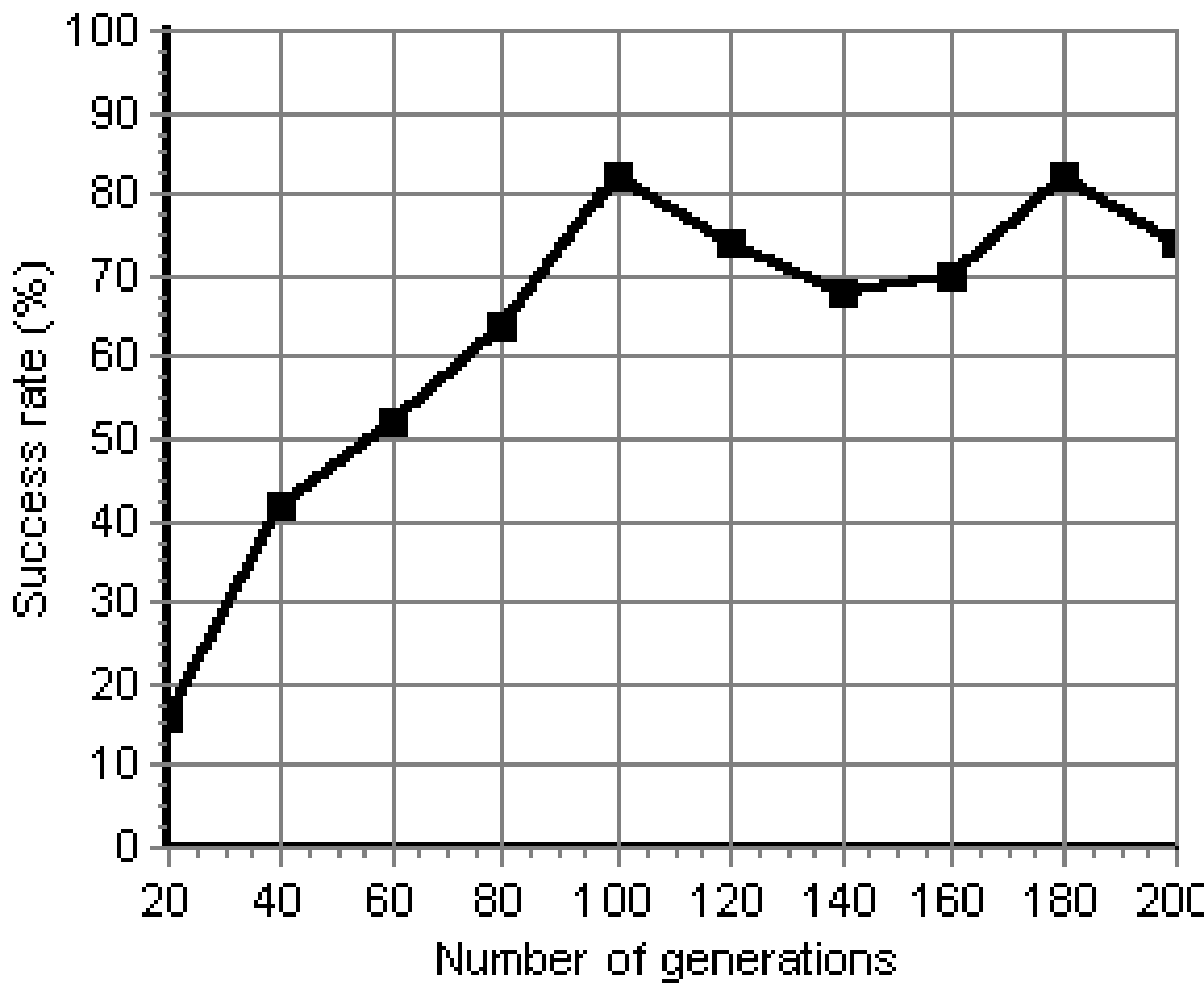}}
\caption{The relationship between the number of generations and the rate of success. The results are averaged over 50 runs. The number of generations varies between 20 and 200.}
\label{nim3}
\end{figure}

The results of this experiment are depicted in Figure \ref{nim3}.

Figure \ref{nim3} shows that MEP is able to find a winning strategy for 
the \textbf{\textit{Nim}} game in most of the runs. In 41 runs (out of 50) a 
perfect solutions was obtained after 100 generations. 9 successful runs were 
obtained when the algorithm is run for 20 generations.\\

\textbf{Experiment 3}\\

In the third experiment the relationship between the chromosome length and 
the success rate is analyzed. MEP algorithm parameters are given in Table \ref{nim_tb3}.

\begin{table}[htbp]
\caption{MEP algorithm parameters for Experiment 3.}
\label{nim_tb3}
\begin{center}
\begin{tabular}
{|p{140pt}|p{211pt}|}
\hline
\textbf{Parameter}& 
\textbf{Value} \\
\hline
Number Of Generations& 
50 \\
\hline
Population Size& 
100 individuals \\
\hline
Crossover probability& 
0.9 \\
\hline
Mutations& 
2 mutations / chromosome \\
\hline
Selection strategy& 
binary tournament \\
\hline
Terminal set& 
$T_{Nim}$ = {\{}$n$, $a_{1}$, $a_{2}$, \ldots , $a_{n}${\}}. \\
\hline
Function set& 
$F $= {\{}+, -, *, \textbf{div}, \textbf{mod}, \textbf{and}, \textbf{not}, \textbf{xor}, \textbf{or}{\}} \\
\hline
\end{tabular}
\end{center}
\end{table}

\begin{figure}[htbp]
\centerline{\includegraphics[width=2.97in,height=2.46in]{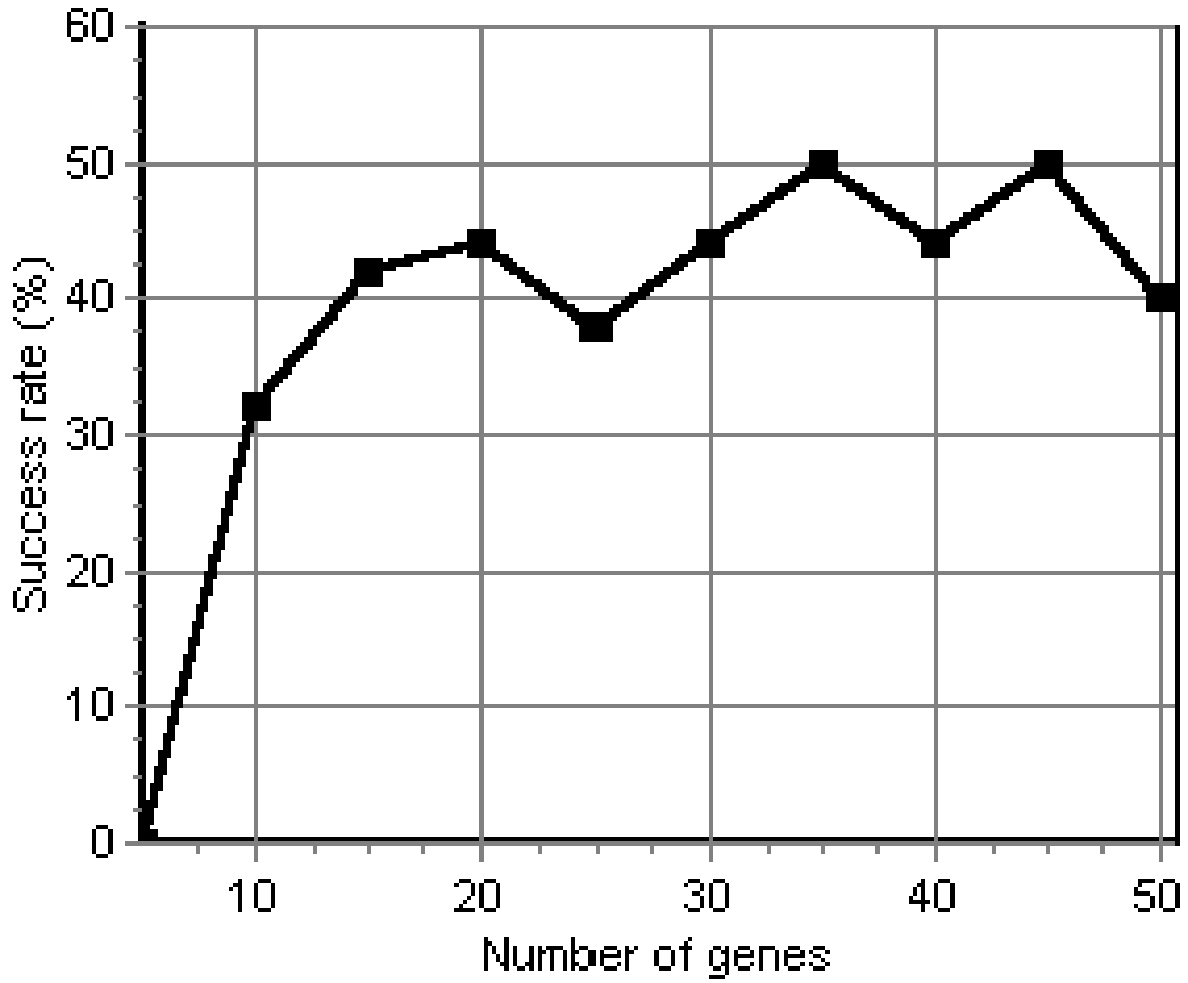}}
\caption{The relationship between the chromosome length and the success rate. The results are averaged over 50 runs. The chromosome length varies between 5 and 50.}
\label{nim4}
\end{figure}

The results of this experiment are depicted in Figure \ref{nim4}.

Figure \ref{nim4} shows that the optimal number of genes of a MEP 
chromosome is 35. With this value 25 runs (out of 50) were successful.

It is interesting to note that the formulas evolved by MEP are sometimes 
different from the classical $a_{1}$ \textbf{\textit{xor}} $a_{2}$ 
\textbf{\textit{xor}} $a_{3}$ \textbf{\textit{xor}} $a_{4}$. For instance a 
correct formula evolved by MEP is:\\

$F=a_{1}$ \textbf{\textit{xor}} $a_{2}$ \textbf{\textit{xor}} $a_{3} \quad -$ $a_{4}$.\\

This formula is also correct due to the properties of the 
\textbf{\textit{xor}} operator.\\

\textbf{Experiment 4}\\

In this experiment a formula for the \textbf{\textit{NimModK}} game is 
evolved. The initial configuration was the same (4, 4, 4, 4) and $k$ was set to 
2. The parameters for the MEP algorithm are given in Table \ref{nim_tb4}.

\begin{table}[htbp]
\caption{MEP algorithm parameters for Experiment 4.}
\label{nim_tb4}
\begin{center}
\begin{tabular}
{|p{125pt}|p{220pt}|}
\hline
\textbf{Parameter}& 
\textbf{Value} \\
\hline
Chromosome length& 
35 genes \\
\hline
Number of generations& 
100 \\
\hline
Population size& 
1000 individuals \\
\hline
Crossover probability& 
0.9 \\
\hline
Mutations& 
2 mutations / chromosome \\
\hline
Selection strategy& 
binary tournament \\
\hline
Terminal set& 
$T_{NimModK}$ = {\{}$n$, $k$, $a_{1}$, $a_{2}$, \ldots , $a_{n}${\}}. \\
\hline
Function set& 
$F $= {\{}+, -, *, \textbf{\textit{div}}, \textbf{\textit{mod}}, \textbf{\textit{and}}, \textbf{\textit{not}}, \textbf{\textit{xor}}, \textbf{\textit{or}}{\}} \\
\hline
\end{tabular}
\end{center}
\end{table}

This problem turned out to be more difficult than the previous one. In only 
3 runs (out of 50) MEP was able to find a perfect formula (i.e. a formula 
that has the fitness equal to 0). 

\section{Evolving heuristics for NP-Complete problems}\label{NP_Complete}

MEP technique is used for discovering TSP heuristics for 
graphs satisfying triangle inequality (TI graphs). This option was chosen 
due to the existence of a big number of real-world applications implying TI 
graphs (e.g. plains, trains and vehicles routes). MEP technique is used to 
learn a path function $f$ that is used for evaluating the reachable nodes. This 
function serves as a heuristic for detecting the optimum path.

This section is entire original and it is based on the paper \cite{oltean_tsp}.

\subsection{MEP for TSP}

In the proposed approach the TSP path starts with a randomly selected node 
of the graph. Each node reachable from the current node in one step is 
evaluated using the function (computer program) $f$ evolved by MEP algorithm. 
The best node is added to the already detected path. The algorithm stops 
when the path contains all graph nodes.

MEP learning process for TSP has a remarkable quality: the evolved (learned) 
heuristic works very well for data sets much larger than the training set. 
For MEP training stage graphs having 3 to 50 nodes are considered. Evolved 
MEP function was tested and performs well for graphs having maximum 1000 
nodes.

Evolved function $f$ is compared with some well known heuristics. Numerical 
experiments emphasize that (for considered examples) MEP function 
outperforms dedicated heuristics. 

\subsection{TSP problem with triangle inequality}

TSP problem for TI graphs (i.e. satisfying triangle inequality) is stated as 
follows.

Consider a set $C$ = {\{}$c_{0}$, $c_{1}$,\ldots , $c_{N\mbox{--}1}${\}} of 
cities, and a distance $d(c_{i}$, $c_{j}) \quad  \in $ Z$^{ + }$ for each pair 
$c_{i}$, $c_{j} \quad  \in  \quad C, d(c_{i}$, $c_{j})=d(c_{j}$, $c_{i})$, and for each 
three cities $c_{i}$, $c_{j}$, $c_{k} \quad  \in  \quad C$, $d(c_{i}$, $c_{j}) \quad  \le $ 
$d(c_{i}$, $c_{k})+d(c_{k}$, $c_{j})$. The tour $<$c$_{\pi (0)}$, c$_{\pi (1)}$, 
\ldots , c$_{\pi (N\mbox{--}1)}>$ of all cities in $C$ having minimum length is 
needed \cite{aarts1,garey1}

TSP problem with triangle inequality is an NP-complete problem \cite{garey1}. No 
polynomial time algorithm for solving TSP problem is known.

Several heuristics for solving TSP problem have been proposed. The most 
important are Nearest Neighbor  and the Minimum Spanning Tree 
\cite{cormen1,garey1}.

In this section we address the problem of discovering heuristics that can 
solve TSP rather than solving a particular instance of the problem.

MEP technique is used for evolving a path function $f$ that gives a way to 
choose graph vertices in order to obtain a Hamiltonian cycle. The fitness is 
assigned to a function $f$ in the current population by applying $f$ on several 
randomly chosen graphs (training set) and evaluating the results.

Evolved path function may be used for solving particular instances of TSP. 
For each problem the graph nodes are evaluated using the path function $f$ and 
are added one by one to the already build path. 

The algorithm for TSP using evolved path function $f$ may be described as 
follows:\\

$S_{1}$. Let $c_{\pi (0)}=c_{0}$ {\{}the path starts with the node 
c$_{0}${\}}

$S_{2}$. $k$ = 1; 

$S_{3}$. \textbf{while} $k < N$ -- 1 \textbf{do}

$S_{4}$. \hspace{1cm}Using function $f$ select $c_{\pi (k + 1) }$ -- the next node of the 
path 

$S_{5}$. \hspace{1cm}Add $c_{\pi (k + 1)}$ to the already built path. 

$S_{6}$. \hspace{1cm}$k=k$ + 1;

$S_{7}$. \textbf{endwhile}\\

$S_{4}$ is the key step of this algorithm. The procedure that selects the 
next node of the path in an optimal way uses the function $f$ evolved by the 
MEP technique as described in the next sections.

\subsection{Terminals and functions for evolving heuristic function \textit{f}}

Path function $f$ has to use (as input) some information about already build 
path and some information about unvisited nodes. We consider a special terminal set which is independent with respect to the number of graph nodes.

Let us denote by $y_{1}$ the last visited node (current node). We have to 
select the next node to be added to the path. In this respect all unvisited 
nodes are considered. Let us denote by $y_{2}$ the next node to be visited.

For evolving path function $f$ we consider a set $T$ of terminals involving the 
following elements:\\

\textit{d{\_}y}$_{1}$\textit{{\_}y}$_{2}$ -- distance between the graph nodes $y_{1}$ and $y_{2}$,

\textit{min{\_}g{\_}y}$_{1}$ (\textit{min{\_}g{\_}y}$_{2})$ -- the minimum distance from the nodes $y_{1}$ ($y_{2})$ to 
unvisited nodes, 

\textit{sum{\_}g{\_}y}$_{1}$ (\textit{sum{\_}g{\_}y}$_{2})$ -- the sum of all distances between nodes $y_{1}$ ($y_{2})$ 
and unvisited nodes, 

\textit{prod{\_}g{\_}y}$_{1}$ (\textit{prod{\_}g{\_}y}$_{2})$ -- the product of all distances between nodes $y_{1}$ 
($y_{2})$ and unvisited nodes,

\textit{max{\_}g{\_}y}$_{1}$ (\textit{max{\_}g{\_}y}$_{2})$ -- the maximum distance from the nodes $y_{1}$ ($y_{2})$ to 
unvisited nodes,

\textit{length} -- the length of the already built path.\\

The set $T$ of terminals (function variables) is thus:\\

$T$ = {\{}\textit{d{\_}y}$_{1}$\textit{{\_}y}$_{2}$, \textit{min{\_}g{\_}y}$_{1}$, \textit{min{\_}g{\_}y}$_{2}$, \textit{max{\_}g{\_}y}$_{1}$, \textit{max{\_}g{\_}y}$_{2}$, \textit{sum{\_}g{\_}y}$_{1}$, 
\textit{sum{\_}g{\_}y}$_{2}$, \textit{prod{\_}g{\_}y}$_{1}$, \textit{prod{\_}g{\_}y}$_{2}$, \textit{length}{\}}.\\

Let us remark that members of $T$ are not actual terminals (in the standard 
acceptation). For this reason we may call members of $T$ as \textit{instantiated} (or 
\textit{intermediate}) \textit{nonterminals}.

Set $T$ of terminals is chosen in such way to be independent of the number of 
graph nodes. This choice confers flexibility and robustness to the evolved 
heuristic.

For evolving a MEP function for TSP problem we may consider the following 
set of function symbols: $F$ = {\{}+, -, /, *, \textit{cos}, \textit{sin},\textit{ min},\textit{ max}{\}}.

The node $y_{2}$ that generates the lowest output of evolved function $f$ is 
chosen to be the next node of the path. Ties are solved arbitrarily. For 
instance we may consider the node with the lowest index is selected.

\textbf{Example}\\

Consider the MEP linear structure:\\

1: $d${\_}$y_{1}${\_}$y_{2}$

2: \textit{min{\_}g{\_}y}$_{1}$

3: + 1, 2

4: \textit{sum{\_}g{\_}y}$_{2}$

5: * 2, 4\\

This MEP individual encodes the path functions $f_{1}$, $f_{2}$, $f_{3}$, 
$f_{4}$, $f_{5}$ given by:\\

$f_{1}=d${\_}$y_{1}${\_}$y_{2 }$,

$f_{2}$ = \textit{min{\_}g{\_}y}$_{1 }$,

$f_{3}=d${\_}$y_{1}${\_}$y_{2}$ + \textit{min{\_}g{\_}y}$_{1 }$,

$f_{4}$ = \textit{sum{\_}g{\_}y}$_{2 }$,

$f_{5}$ = \textit{min{\_}g{\_}y}$_{1 }$* \textit{sum{\_}g{\_}y}$_{2}$.

\subsection{Fitness assignment}

In order to obtain a good heuristic we have to train the path function $f$ 
using several graphs. The training graphs are randomly generated at the 
beginning of the search process and remain unchanged during the search 
process. To avoid overfitting (see \cite{prechelt1}), another set of randomly generated 
graphs (validation set) is considered. After each generation the quality of 
the best-so-far individual is calculated using the validation set in order 
to check its generalization ability during training. At the end of the 
search process, the function with the highest quality is supplied as the 
program output.

The fitness (quality) of a detected path function $f$ is defined as the sum of 
the TSP path length of graphs in the training set. Thus the fitness is to be 
minimized.

\subsection{A numerical experiment}

In this experiment we evolve a heuristic for solving TSP problem.

Let us denote by $G_{k}$ the set of class of TI graphs having maximum $k$ nodes.

MEP algorithm considers the class $G_{50}$ (i.e. graphs having 3 to 50 nodes) 
for training and the class $G_{100}$ for validation. Evolved path function 
was tested for graphs in the class $G_{1000}$ (i.e. graphs having maxim 1000 
nodes). MEP algorithm parameters are given in Table \ref{tsp_tb1}.

\begin{table}[htbp]
\label{tsp_tb1}
\caption{MEP algorithm parameters for evolving a heuristic for TSP with triangle inequality}
\begin{center}
\begin{tabular}
{|p{220pt}|p{130pt}|}
\hline
Parameter&
Value\\
\hline
Population size& 
300 \\
\hline
Number of generations& 
100 \\
\hline
Chromosome length& 
40 genes \\
\hline
Mutation probability& 
0.1 \\
\hline
Crossover type& 
One-Crossover-Point \\
\hline
Crossover probability& 
0.9 \\
\hline
Training set size& 
30 \\
\hline
Maximum number of nodes in training set& 
50 \\
\hline
Validation set size& 
20 \\
\hline
Maximum number of nodes in validation set& 
100 \\
\hline
\end{tabular}
\end{center}
\end{table}

\begin{figure}[htbp]
\centerline{\includegraphics[width=3.58in,height=3.07in]{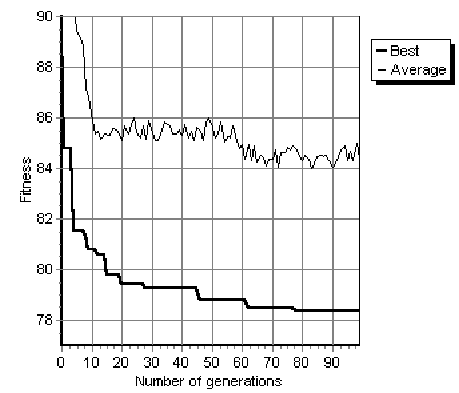}}
\caption{The fitness evolution of the best individual in the best 
run and the average fitness of the best individuals over 30 runs.}
\label{tsp1}
\end{figure}

The evolution of the best individual fitness and the average fitness of the 
best individuals over 30 runs are depicted in Table \ref{tsp1}.

A path function evolved by the MEP algorithm is:\\

$f$ = (\textit{sum{\_}g}($y2))$ * (\textit{d{\_}y}$_{1}$\textit{{\_}y}$_{2}$ - (\textit{max}(\textit{d{\_}y}$_{1}$\textit{{\_}y}$_{2}$, \textit{max{\_}g}($y_{1}))$) + 
\textit{d{\_}y}$_{1}$\textit{{\_}y}$_{2})$.\\

Heuristic function $f$ that is evolved by MEP technique is directly used for 
building the optimum path. The corresponding learning process has a 
remarkable quality: the evolved (learned) heuristic works very well on data 
sets significantly larger than the training set. In our example the training 
set $G_{50}$ is significantly smaller than the set $G_{1000}$ used for 
testing.

\subsection{Assessing the performance of the evolved MEP heuristic}

In this section the performance of evolved MEP heuristic, NN and MST are 
compared. In the first experiment we compare the considered algorithms on 
some randomly generated graphs. In the second experiment the heuristics are 
compared against several difficult problems in TSPLIB \cite{reinelt1}.\\

\textbf{Experiment 1}\\

In this experiment we provide a direct comparison of the evolved MEP 
heuristic, NN and MST. The considered heuristics are tested for randomly 
generated graphs satisfying triangle inequality.

Evolved heuristic was tested for different graphs from the classes 
$G_{200}$, $G_{500}$ and $G_{1000}$. For each graph class 1000 graphs 
satisfying triangle inequality have been randomly generated. These graphs 
have been considered for experiments with evolved MEP heuristic, NN and MST.

Performance of evolved MEP heuristic, NN and MST are depicted in Table \ref{tsp_tb2}.

\begin{table}[htbp]
\caption{Evolved MEP heuristic vs. NN, MST. For each graph class we present the number of graphs for which evolved MEP heuristic generates a cycle shorter than the cycle obtained by the algorithm MST and NN.}
\label{tsp_tb2}
\begin{center}
\begin{tabular}
{|p{87pt}|p{45pt}p{42pt}|}
\hline
Graphs types& 
MST& 
NN \\
\hline
$G_{200}$& 
953& 
800 \\
\hline
$G_{500}$& 
974& 
906 \\
\hline
$G_{1000}$& 
990& 
948 \\
\hline
\end{tabular}
\end{center}
\end{table}

Results obtained emphasizes that evolved MEP heuristic outperforms NN and 
MST algorithms on random graphs. \\

\textbf{Experiment 2}\\

To obtain a stronger evidence of the results above we test the performance 
of the considered heuristics against some difficult problems in TSPLIB. The 
results are presented in Table \ref{tsp_tb3}.

\begin{table}[htbp]
\caption{The performance of evolved MEP heuristic, NN and MST on 
some problems in TSPLIB. \textit{Length} is the length of the TSP path obtained with one of 
the considered heuristics. Error is calculated as (\textit{Length} - 
\textit{Shortest{\_}Length})/ \textit{Shortest{\_}Length} * 100. Each node of the graph has been considered as the first node of the path}
\label{tsp_tb3}
\begin{center}
\begin{tabular}
{|p{45pt}|p{45pt}|p{43pt}|p{45pt}|p{50pt}|p{45pt}|p{40pt}|}
\hline
\textbf{Problem}& 
\multicolumn{2}{p{94pt}}{\textbf{MEP}} & 
\multicolumn{2}{p{103pt}}{\textbf{NN}} & 
\multicolumn{2}{p{98pt}}{\textbf{MST}}  \\
& 
Length& 
Error ({\%})& 
Length& 
Error ({\%})& 
Length& 
Error ({\%}) \\
\hline
a280 & 
2858.86 & 
10.85& 
3084.22& 
19.58976& 
3475.23& 
34.75 \\
\hline
att48 & 
37188.2 & 
10.93& 
39236.9& 
17.04227& 
43955.8& 
31.11 \\
\hline
berlin52 & 
7672.1 & 
1.72& 
8182.19& 
8.488332& 
10403.9& 
37.94 \\
\hline
bier127 & 
134945 & 
14.08& 
127954& 
8.177068& 
152747& 
29.13 \\
\hline
ch130 & 
6558.03 & 
7.33& 
7198.74& 
17.81899& 
8276.51& 
35.45 \\
\hline
ch150 & 
7104.03 & 
8.82& 
7078.44& 
8.431985& 
9142.99& 
40.05 \\
\hline
d198 & 
17780.7 & 
12.67& 
17575.1& 
11.37579& 
17957.6& 
13.79 \\
\hline
d493 & 
43071.3 & 
23.05& 
41167& 
17.61328& 
41846.6& 
19.55 \\
\hline
d657 & 
56965.6 & 
16.46& 
60398.7& 
23.48442& 
63044.2& 
28.89 \\
\hline
eil101 & 
685.013 & 
8.9& 
753.044& 
19.72083& 
846.116& 
34.51 \\
\hline
eil51 & 
441.969 & 
3.74& 
505.298& 
18.61455& 
605.049& 
42.03 \\
\hline
eil76 & 
564.179 & 
4.86& 
612.656& 
13.87658& 
739.229& 
37.4 \\
\hline
fl417& 
13933.8 & 
17.47& 
13828.2& 
16.58545& 
16113.2& 
35.85 \\
\hline
gil262& 
2659.17& 
11.82& 
2799.49& 
17.72456& 
3340.84& 
40.48 \\
\hline
kroA150 & 
28376.3 & 
6.98& 
31482& 
18.6925& 
38754.8& 
46.11 \\
\hline
kroA200 & 
32040.3 & 
9.09& 
34547.7& 
17.63722& 
40204.1& 
36.89 \\
\hline
kroB100 & 
24801 & 
12.01& 
25883& 
16.90077& 
28803.5& 
30.09 \\
\hline
kroB200 & 
33267.4 & 
13.01& 
35592.4& 
20.91042& 
40619.9& 
37.98 \\
\hline
lin105 & 
15133.2 & 
5.24& 
16939.4& 
17.80652& 
18855.6& 
31.13 \\
\hline
lin318 & 
46203.4 & 
9.93& 
49215.6& 
17.09915& 
60964.8& 
45.05 \\
\hline
pcb442 & 
56948.3 & 
12.15& 
57856.3& 
13.9397& 
73580.1& 
44.9 \\
\hline
pr226 & 
84937.8 & 
5.68& 
92905.1& 
15.59818& 
111998& 
39.35 \\
\hline
pr264& 
55827.1& 
13.61& 
54124.5& 
10.15468& 
65486.5& 
33.27 \\
\hline
rat195 & 
2473.49 & 
6.47& 
2560.62& 
10.22901& 
2979.64& 
28.26 \\
\hline
rat575& 
7573.6& 
11.82& 
7914.2& 
16.84925& 
9423.4& 
39.13 \\
\hline
rat783& 
9982.96& 
13.36& 
10836.6& 
23.05928& 
11990.5& 
36.16 \\
\hline
rd400& 
16973.3& 
11.07& 
18303.3& 
19.77816& 
20962& 
37.17 \\
\hline
ts225 & 
136069 & 
7.44& 
140485& 
10.92994& 
187246& 
47.85 \\
\hline
u574& 
43095.6& 
16.77& 
44605.1& 
20.86465& 
50066& 
35.66 \\
\hline
u724& 
46545.7& 
11.06& 
50731.4& 
21.04844& 
60098.9& 
43.39 \\
\hline
\end{tabular}
\end{center}
\end{table}

From Table \ref{tsp_tb3} we can see that evolved MEP heuristic performs better than NN 
and MST on most of the considered problems. Only for five problems (bier127, 
ch150, d198, d493, fl417) NN performs better than evolved MEP heuristic. MST 
does not perform better than evolved MEP heuristic for no problem. The 
highest error obtained by the evolved MEP heuristic is 23.05 (the problem 
d493) while the highest error obtained by NN is 23.45 (the problem rd400). 
The lowest error obtained with MEP is 1.72 (problem berlin52) while the 
lowest error obtained by NN is 8.17 (problem bier127). The mean of errors 
for all considered problems is 10.61 (for evolved MEP heuristic) 16.33 (for 
NN heuristic) and 35.77 (for MST heuristic).

\section{Conclusions and further work}

Three approaches have been proposed in this section:

\begin{itemize}

\item{an evolutionary approach for the \textbf{\textit{Nim}} game. The underlying evolutionary technique is \textit{Multi Expression Programming} - a very fast and efficient \textit{Genetic Programming} variant. Numerical experiments have shown that MEP is able to discover a winning strategy in most of the runs.

The proposed method can be easily applied for games whose winning strategy 
is based on $P$ and $N$-positions. The idea is to read the game tree and to count 
the number of configurations that violates the rules of the winning 
strategy.}

\item{an evolutionary approach for the \textbf{\textit{Tic-Tac-Toe}} game. Evolved strategy is very fast. About 400.000 games/second can be played without loss.}
\item{an evolutionary approach for evolving heuristics for the TSP. Numerical experiments have shown that the evolved heuristic performs better than the NN and MST heuristics.}

\end{itemize}

\subsection{Applying MEP for generating complex game strategies}
Using the \textit{all-possibilities} technique (a backtracking procedure that plays all the moves for the second player) allows us to compute the absolute quality of a game position. 
For complex games a different fitness assignment technique is needed since the moves for the second player can not be simulated by an \textit{all-possibilities} procedure (as the number of moves that needs to be simulated is too large).
One fitness assignment possibility is to use a heuristic procedure that acts as the second player. But there are several difficulties related to this approach. If the heuristic is very good it is possible that none of evolved strategy could ever beat the heuristic. If the heuristic procedure plays as a novice then many evolved strategy could beat the heuristic from the earlier stages of the search process. In the last case fitness is not correctly assigned to population members and thus we can not perform a correct selection. 
A good heuristic must play on several levels of complexity. At the beginning of the search process the heuristic procedure must play at easier levels. As the search process advances, the level of difficulty of the heuristic procedure must increases.
However, for complex games such a procedure is difficult to implement.
Another possibility is to search for a game strategy using a coevolutionary algorithm. This approach seems to offer the most spectacular results. In this case, MEP population must develop intelligent behavior based only on internal competition.

\chapter{Infix Form Genetic Programming}\label{ifgp_chap}

A new GP variant called Infix Form Genetic Programming (IFGP) is described. IFGP individuals are arrays of integer values encoding mathematical expressions in infix form. IFGP is used for solving real-world classification problems.

The chapter is entirely original and it is based on the paper \cite{oltean_ifgp}.

\section{Introduction}

Classification is the task of assigning inputs to a number of discrete 
categories or classes \cite{haykin1}. Examples include classifying a handwritten 
letter as one from A-Z, classifying a speech pattern to the corresponding 
word, etc.

Machine learning techniques have been extensively used for solving 
classification problems. In particular Artificial Neural Networks (ANNs) \cite{haykin1,prechelt1} 
have been originally designed for classifying a set of points in two 
distinct classes. Genetic Programming (GP) techniques \cite{koza1} have also been used for 
classification purposes. For instance, LGP \cite{brameier1} has been used for solving several classification 
problems in PROBEN1. The conclusion was that LGP is able to solve the 
classification problems with the same error rate as a neural network.

Infix Form Genetic Programming (IFGP), chromosomes are strings encoding complex 
mathematical expressions using infix form. An interesting 
feature of IFGP is its ability of storing multiple solutions of a problem in 
a chromosome. 

In what follows IFGP is described and used for solving several real-world classification 
problems taken from PROBEN1 \cite{prechelt1}. 

\section{Prerequisite}

We denote by $F$ the set of function symbols (or operators) that may appear in 
a mathematical expression. $F$ usually contains the binary operators {\{}+, $ - 
$, *, /{\}}. \textit{Number{\_}of{\_}Operators} denotes the number of elements in $F$. A correct 
mathematical expression also contains some terminal symbols. The set of 
terminal symbols is denoted by $T$. The number of terminal symbols is denoted 
by \textit{Number{\_}of{\_}Variables}.

The symbols that may appear in a mathematical expression encoded by the IFGP are from the set $T \quad  \cup $ 
$F \quad  \cup $ {\{}'(', ')'{\}}. The total number of symbols that may appear in a 
valid mathematical expression is denoted by \textit{Number{\_}of{\_}Symbols}.

By $C_{i}$ we denote the value on the $i^{th}$ gene in a IFGP chromosome and 
by $G_{i}$ the symbol in the $i^{th}$ position in the mathematical expression 
encoded into an IFGP chromosome.

\section{Individual representation}

In this section we describe how IFGP individuals are represented and how 
they are decoded in order to obtain a valid mathematical expression.

Each IFGP individual is a fixed size string of genes. Each gene is an 
integer number in the interval [0 .. \textit{Number{\_}Of{\_}Symbols }- 1]. An IFGP individual can be 
transformed into a functional mathematical expression by replacing each gene 
with an effective symbol (a variable, an operator or a parenthesis).\\

\textbf{Example}\\

If we use the set of functions symbols $F$ = {\{}+, *, -, /{\}}, and the 
set of terminals $T$ = {\{}$a$, $b${\}}, the following chromosome

C = 7, 3, 2, 2, 5

\noindent
is a valid chromosome in IFGP system.

\section{Decoding IFGP individuals}\label{ifgp}

We will begin to decode this chromosome into a valid mathematical 
expression. In the first position (in a valid mathematical expression) we 
may have either a variable, or an open parenthesis. That means that we have 
\textit{Number{\_}Of{\_}Variables} + 1 possibilities to choose a correct symbol on the first position. We put 
these possibilities in order: the first possibility is to choose the 
variable $x_{1}$, the second possibility is to choose the variable $x_{2}$ 
\ldots the last possibility is to choose the closed parenthesis ')'. The 
actual value is given by the value of the first gene of the chromosome. 
Because the number stored in a chromosome gene may be larger than the number 
of possible correct symbols for the first position we take only the value of 
the first gene \textit{modulo} number of possibilities for the first gene.

Generally, when we compute the symbol stored in the $i^{th}$ position in 
expression we have to compute first how many symbols may be placed in that 
position. The number of possible symbols that may be placed in the current 
position depends on the symbol placed in the previous position. Thus:

\begin{itemize}
\item[{\it (i)}]{if the previous position contains a variable ($x_{i})$, then for the current 
position we may have either an operator or a closed parenthesis. The closed 
parenthesis is considered only if the number of open parentheses so far is 
larger than the number of closed parentheses so far.}
\item[{\it (ii)}]{if the previous position contains an operator, then for the current position 
we may have either a variable or an open parenthesis. }
\item[{\it (iii)}]{if the previous position contains an open parenthesis, then for the current 
position we may have either a variable or another open parenthesis.}
\item[{\it (iv)}]{if the previous position contains a closed parenthesis, then for the current 
position we may have either an operator or another closed parenthesis. The 
closed parenthesis is considered only if the number of open parentheses so 
far is larger than the number of closed parentheses.}
\end{itemize}

Once we have computed the number of possibilities for the current position 
it is easy to determine the symbol that will be placed in that position: 
first we take the value of the corresponding gene modulo the number of 
possibilities for that position. Let $p$ be that value \\

($p=C_{i}$ \textit{mod} \textit{Number{\_}Of{\_}Possibilities}). \\

The $p^{th}$ symbol from the permitted symbols for the current is placed 
in the current position in the mathematical expression. (Symbols that may 
appear into a mathematical expression are ordered arbitrarily. For instance 
we may use the following order: $x_{1}$, $x_{2}$, \ldots , +, -, *, /, 
'(', ')'. )

All chromosome genes are translated but the last one. The last gene is used 
by the correction mechanism (see below).

The obtained expression usually is syntactically correct. However, in some 
situations the obtained expression needs to be repaired. There are two cases 
when the expression needs to be corrected:

The last symbol is an operator (+, -, *, /) or an open parenthesis. In 
that case a terminal symbol (a variable) is added to the end of the 
expression. The added symbol is given by the last gene of the chromosome.

The number of open parentheses is greater than the number of closed 
parentheses. In that case several closed parentheses are automatically added 
to the end in order to obtain a syntactically correct expression. \\

\textbf{Remark}\\
If the correction mechanism is not used, the last gene of the chromosome will not be used.\\

\textbf{Example}\\

Consider the chromosome \\

$C$ = 7, 3, 2, 0, 5, 2 \\

and the set of terminal and function symbols previously defined \\

$T$ = {\{}$a$, $b${\}}, \\

$F$ = {\{}+, -, *, /{\}}.\\

For the first position we have 3 possible symbols ($a$, $b$ and `(`). Thus, the 
symbol in the position $C_{0}$ \textbf{\textit{mod}} 3 = 1 in the array of 
possible symbols is placed in the current position in expression. The chosen 
symbol is $b$, because the index of the first symbol is considered to be 0.

For the second position we have 4 possibilities (+, -, *, /). The 
possibility of placing a closed parenthesis is ignored since the difference 
between the number of open parentheses and the number of closed parentheses 
is zero. Thus, the symbol '/' is placed in position 2.

For the third position we have 3 possibilities ($a$, $b$ and '('). The symbol 
placed on that position is an open parenthesis '('. 

In the fourth position we have 3 possibilities again ($a$, $b$ and '('). The 
symbol placed on that position is the variable $a$.

For the last position we have 5 possibilities (+, -, *, /) and the 
closed parenthesis ')'. We choose the symbol on the position 5 \textbf{mod} 
5 = 0 in the array of possible symbols. Thus the symbol `+' is placed in 
that position.

The obtained expression is $E=b$ / ($a$+.

It can be seen that the expression $E$ it is not syntactically correct. For 
repairing it we add a terminal symbol to the end and then we add a closed 
parenthesis. Now we have obtained a correct expression:

\begin{center}
$E=b$ / ($a+a)$.
\end{center}

\begin{figure}[htbp]
\centerline{\includegraphics[width=1.49in,height=1.59in]{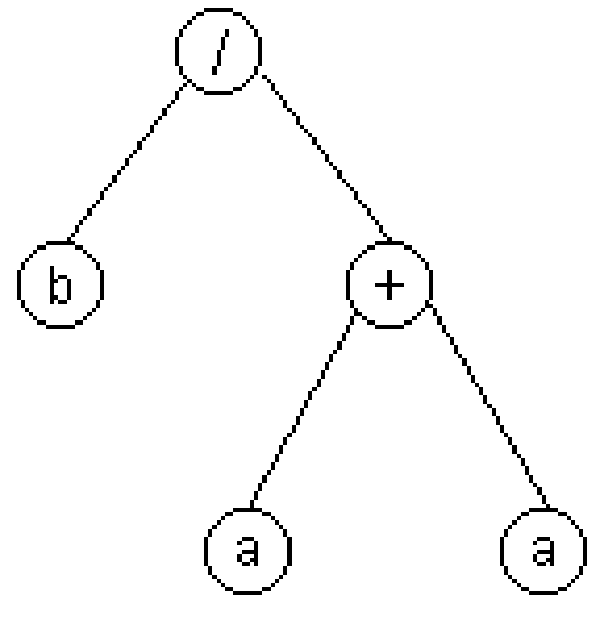}}
\caption{The expression tree of $E=b$ / ($a+a)$.}
\label{ida1}
\end{figure}

The expression tree of $E$ is depicted in Figure \ref{ida1}.

\section{Using constants within the IFGP model}

An important issue when designing a new GP technique is the way in which the 
constants are embaded into the proposed model.

Fixed or ephemeral random constants have been extensively tested within GP 
systems \cite{koza1}. The interval over which the constants are initially generated 
is usually problem-dependent. For a good functionality of the program a 
priori knowledge about the required constants is usually needed. 

By contrast, IFGP involves a problem-independent system of constants. 

It is known that each real number may be written as a sum of powers of 2. 

Within the IFGP model each constant is a power of 2. The total number of 
constants is also problem-independent. For instance, if we want to solve 
problems using double precision constants, we need 127 constants:\\

2$^{ - 63}$, 2$^{ - 62}$,\ldots , 2$^{ - 1}$, 2$^{0}$, 2$^{1}$,\ldots , 2$^{63}$. \\

Particular instances of the classification problems may require fewer 
constants. As can be seen in section \ref{ifgp} a good solution for some 
classification problems can be obtained without using constants. However, if 
we do not know what kind of constants are required by the problems being 
solved it is better the use the double precision system of constants (as 
described above).

Within the IFGP chromosome each constant is represented by its exponent. For 
instance the constant 2$^{17}$ is stored as the number 17, the constant 2$^{ 
- 4}$ is stored as the number  -4 and the constant 1 = 2$^{0}$ is stored as 
the number 0. These constants will act as terminal symbols. Thus the 
extended set of terminal symbols is \\

$T$ = {\{}$x_{1}$, $x_{2}$,\ldots , -63, -62, .., -1, 0, 1, \ldots , 62, 63{\}}.

\section{Fitness assignment process}

In this section we describe how IFGP may be efficiently used for solving 
classification problems.

A GP chromosome usually stores a single solution of a problem and the 
fitness is normally computed using a set of fitness cases. 

Instead of encoding a single solution, an IFGP individual is allowed to 
store multiple solutions of a problem. The fitness of each solution is 
computed in a conventional manner and the solution having the best fitness 
is chosen to represent the chromosome. 

In the IFGP representation each sub-tree (sub-expression) is considered as a 
potential solution of a problem. \\

\textbf{Example} \\

The previously obtained expression (see section \ref{ifgp}) contains 4 distinct solutions 
(sub-expressions):
\\

$E_{1}=a$,

$E_{2}=b$,

$E_{3}=a+a$,

$E_{4}=b$ / ($a+a)$.
\\

Now we will explain how the fitness of a (sub)expression is computed.

Each class has associated a numerical value: the first class has the value 
0, the second class has the value 1 and the $m^{th}$ class has associated the 
numerical value $m - $1. Any other system of distinct numbers may be used. We 
denote by $o_{k}$ the number associated to the $k^{th}$ class. 

The value $v_{j}(E_{i})$ of each expression $E_{i}$ (in an IFGP) chromosome 
for each row (example) $j$ in the training set is computed. Then, each row in 
the training set will be classified to the nearest class (the class $k$ for 
which the difference $\vert v_{j}(E_{i})-o_{k}\vert $ is minimal). The 
fitness of a (sub)expression is equal to the number of incorrectly 
classified examples in the training set. The fitness of an IFGP chromosome 
will be equal to the fitness of the best expression encoded in that 
chromosome.\\

\textbf{Remarks}\\

\begin{itemize}
\item[{\it (i)}] Since the set of numbers associated with the problem classes was 
arbitrarily chosen it is expected that different systems of number to 
generate different solutions.

\item[{\it(ii)}] When solving symbolic regression or classification problems IFGP 
chromosomes need to be traversed twice for computing the fitness. That means 
that the complexity of the IFGP decoding process it is not higher than the 
complexity of other methods that store a single solution in a chromosome.

\end{itemize}

\section{Search operators}

Search operators used within the IFGP model are recombination and mutation. 
These operators are similar to the genetic operators used in conjunction 
with binary encoding \cite{dumitrescu1}.

\subsection{Crossover}

By recombination two parents exchange genetic 
material in order to obtain two offspring. In our numerical experiments only two-point 
recombination is used. 

\subsection{Mutation}

Mutation operator is applied with a fixed mutation probability ($p_{m})$. By mutation a randomly generated value over 
the interval [0, \textit{Number{\_}of{\_}Symbols}-1] is assigned to the target gene.

\section{Handling exceptions within IFGP}

Exceptions are special situations that interrupt the normal flow of 
expression evaluation (program execution). An example of exception is 
\textit{division by zero} which is raised when the divisor is equal to zero.

GP techniques usually use a \textit{protected exception} handling mechanism \cite{koza1}. For instance if a 
division by zero exception is encountered, a predefined value (for instance 
1 or the numerator) is returned. This kind of handling mechanism is specific 
for Linear GP \cite{brameier1}, standard GP \cite{koza1} and GE \cite{oneill1}.

IFGP uses a new and specific mechanism for handling exceptions. When an 
exception is encountered (which is always generated by a gene containing a 
function symbol), the entire (sub) tree which has generated the exception is 
mutated (changed) into a terminal symbol. Exception handling is performed 
during the fitness assignment process.

\section{IFGP algorithm}

A steady-state \cite{syswerda1} variant of IFGP is employed in this section. The algorithm 
starts with a randomly chosen population of individuals. The following stpng are repeated until a termination condition is reached. Two parents are chosen at each step using binary tournament selection \cite{dumitrescu1}. The selected 
individuals are recombined with a fixed crossover probability $p_{c}$. By 
recombining two parents, two offspring are obtained. The offspring are 
mutated and the best of them replaces the worst individual in the current 
population (only if the offspring is better than the worst individual in 
population). 

The algorithm returns as its answer the best expression evolved 
for a fixed number of generations.

\section{Solving classification problems using IFGP}

IFGP technique is applied for solving difficult learning problems. Real-world data sets are considered for the training process.

\subsection{Data sets}

Numerical experiments performed in this section are based on several benchmark 
problems taken from PROBEN1 \cite{prechelt1}. These datasets were created based on the 
datasets from the UCI Machine Learning Repository \cite{uci}.

Used problems are briefly described in what follows.\\

\textbf{Cancer}\\

Diagnosis of breast cancer. Try to classify a tumor as either benignant or 
malignant based on cell descriptions gathered by microscopic examination.\\

\textbf{Diabetes}\\

Diagnosis diabetes of Pima Indians. Based on personal data and the results of 
medical examinations try to decide whether a Pima Indian individual is 
diabetes positive or not.\\

\textbf{Heartc}\\

Predicts heart disease. Decides whether at least one of four major vessels 
is reduced in diameter by more than 50{\%}. The binary decision is made 
based on personal data such as age, sex, smoking habits, subjective patient 
pain descriptions and results of various medical examinations such as blood 
pressure and electro cardiogram results.

This data set was originally created by Robert Detrano from V.A. Medical 
Center Long Beach and Cleveland Clinic Foundation.\\

\textbf{Horse}\\

Predicts the fate of a horse that has colic. The results of a veterinary 
examination of a horse having colic are used to predict whether the horse 
will survive will die or should be euthanized.

The number of inputs, of classes and of available examples, for each test 
problem, are summarized in Table \ref{ifgp_tb1}.\\

\begin{table}[htbp]
\caption{Summarized attributes of several classification problems from PROBEN1.}
\label{ifgp_tb1}
\begin{center}
\begin{tabular}
{|p{47pt}|p{92pt}|p{92pt}|p{99pt}|}
\hline
\textbf{Problem}& 
\textbf{Number of inputs}& 
\textbf{Number of classes }& 
\textbf{Number of examples} \\
\hline
\textbf{cancer}& 
9& 
2& 
699 \\
\hline
\textbf{diabetes}& 
8& 
2& 
768 \\
\hline
\textbf{heartc}& 
35& 
2& 
303 \\
\hline
\textbf{horse}& 
58& 
3& 
364 \\
\hline
\end{tabular}
\end{center}
\end{table}

\subsection{Numerical experiments}\label{ifgp_experiments}

The results of several numerical experiments with ANNs, LGP and IFGP are 
presented in this section.

Each data set is divided in three sub-sets (training set -50{\%}, validation 
set - 25 {\%}, and test set - 25{\%}) (see \cite{prechelt1}).

The test set performance is computed for that chromosome which had minim 
validation error during the search process. This method, called \textit{early stopping}, is a good 
way to avoid overfitting \cite{prechelt1} of the population individuals to the particular 
training examples used. In that case the generalization performance will be 
reduced.

In \cite{brameier1} Linear GP was used to solve several classification problems from 
PROBEN1. The parameters used by Linear GP are given in Table \ref{ifgp_tb2}.

\begin{table}[htbp]
\caption{Linear GP parameters used for solving classification tasks from PROBEN1}
\label{ifgp_tb2}
\begin{center}
\begin{tabular}
{|p{196pt}|p{155pt}|}
\hline
\textbf{Parameter}& 
\textbf{Value} \\
\hline
Population size& 
5000 \\
\hline
Number of demes& 
10 \\
\hline
Migration rate& 
0.05 \\
\hline
Classification error weight in fitness& 
1.0 \\
\hline
Maximum number of generations& 
250 \\
\hline
Crossover probability& 
0.9 \\
\hline
Mutation probability& 
0.9 \\
\hline
Maximum mutation step size for constants& 
$\pm 5$ \\
\hline
Maximum program size& 
256 instructions \\
\hline
Initial maximum program size& 
25 instructions \\
\hline
Function set& 
{\{}+, -, *, /, \textit{sin}, \textit{exp}, if $>$, if $ \le ${\}}\\
\hline
Terminal set& 
{\{}0,..,256{\}} $ \cup $ {\{}input variables{\}} \\
\hline
\end{tabular}
\end{center}
\end{table}

IFGP algorithm parameters are given in Table \ref{ifgp_tb3}.

\begin{table}[htbp]
\caption{IFGP algorithm parameters for solving classification problems from PROBEN1.}
\label{ifgp_tb3}
\begin{center}
\begin{tabular}
{|p{230pt}|p{175pt}|}
\hline
\textbf{Parameter}& 
\textbf{Value} \\
\hline
Population Size& 
250 \\
\hline
Chromosome length& 
30 \\
\hline
Number of generations& 
250 \\
\hline
Crossover probability & 
0.9 \\
\hline
Crossover type& 
Two-point Crossover \\
\hline
Mutation& 
2 mutations per chromosome \\
\hline
Number of Constants& 
41 {\{}2$^{ -20}$, \ldots , 2$^{0}$, 2$^{20}${\}} \\
\hline
Function set& 
{\{}+, -, *, /, \textit{sin}, \textit{exp}{\}} \\
\hline
Terminal set& 
{\{}input variables{\}} $ \cup $ The set of constants. \\
\hline
\end{tabular}
\end{center}
\end{table}

The results of the numerical experiments are presented in Table \ref{ifgp_tb4}.

\begin{table}[htbp]
\caption{Classification error rates of IFGP, LGP and ANN for some 
date sets from PROBEN1. LGP results are taken from \cite{brameier1}. ANNs results are 
taken from \cite{prechelt1}. The cases where IFGP is better than LGP have been written on 
a grey background. The cases where IFGP is better than ANNs have been bolded 
and italicized. Results are averaged over 30 runs}
\label{ifgp_tb4}
\begin{center}
\begin{tabular}
{|p{47pt}|p{30pt}|p{35pt}|p{30pt}|p{30pt}|p{35pt}|p{30pt}|p{35pt}|p{30pt}|}
\hline
\textbf{Problem}& 
\multicolumn{3}{p{105pt}}{\textbf{IFGP--test set}} & 
\multicolumn{3}{p{105pt}}{\textbf{LGP--test set}} & 
\multicolumn{2}{p{70pt}}{\textbf{NN--test set}}  \\
\hline
 & 
\textbf{best}& 
\textbf{mean}& 
\textbf{stddev}& 
\textbf{best}& 
\textbf{mean}& 
\textbf{stddev}& 
\textbf{mean}& 
\textbf{stddev} \\
\hline
cancer1& 
1.14& 
2.45& 
0.69& 
0.57& 
2.18& 
0.59& 
1.38& 
0.49 \\
\hline
cancer2& 
4.59& 
6.16& 
0.45& 
4.02& 
5.72& 
0.66& 
4.77& 
0.94 \\
\hline
cancer3& 
3.44& 
4.92& 
1.23& 
3.45& 
4.93& 
0.65& 
3.70& 
0.52 \\
\hline
diabetes1& 
22.39& 
25.64& 
1.61& 
21.35& 
23.96& 
1.42& 
24.10& 
1.91 \\
\hline
diabetes2& 
25.52& 
28.92& 
1.71& 
25.00& 
27.85& 
1.49& 
26.42& 
2.26 \\
\hline
diabetes3& 
21.35& 
25.31& 
2.20& 
19.27& 
23.09& 
1.27& 
22.59& 
2.23 \\
\hline
heart1& 
16.00& 
23.06& 
3.72& 
18.67& 
21.12& 
2.02& 
20.82& 
1.47 \\
\hline
heart2& 
1.33& 
\textbf{\textit{4.40}} & 
2.35& 
1.33& 
7.31& 
3.31& 
5.13& 
1.63 \\
\hline
heart3& 
12.00& 
\textbf{\textit{13.64}} & 
2.34& 
10.67& 
13.98& 
2.03& 
15.40& 
3.20 \\
\hline
horse1& 
23.07& 
31.11& 
2.68& 
23.08& 
30.55& 
2.24& 
29.19& 
2.62 \\
\hline
horse2& 
30.76& 
\textbf{\textit{35.05}} & 
2.33& 
31.87& 
36.12& 
1.95& 
35.86& 
2.46 \\
\hline
horse3& 
30.76& 
35.01& 
2.82& 
31.87& 
35.44& 
1.77& 
34.16& 
2.32 \\
\hline
\end{tabular}
\end{center}
\end{table}

Table \ref{ifgp_tb4} shows that IFGP is able to obtain similar performances as 
those obtained by LGP even if the population size and the chromosome length 
used by IFGP are smaller than those used by LGP. When compared to ANNs we 
can see that IFGP is better only in 3 cases (out of 12).

We are also interested in analysing the relationship between the 
classification error and the number of constants used by the IFGP 
chromosomes. For this purpose we will use a small population made up of only 
50 individuals. Note that this is two magnitude orders smaller than those 
used by LGP. Other IFGP parameters are given in Table \ref{ifgp_tb3}. Experimental 
results are given in Table \ref{ifgp_tb5}.

\begin{table}[htbp]
\caption{Classification error rates of IFGP (on the test set) using 
different number of constants. The cases where IFGP is better than LGP have 
been written on a grey background. The cases where IFGP is better than ANNs 
have been bolded and italicized. Results are averaged over 30 runs.}
\label{ifgp_tb5}
\begin{center}
\begin{tabular}
{|p{43pt}|p{28pt}|p{32pt}|p{27pt}|p{28pt}|p{30pt}|p{30pt}|p{28pt}|p{30pt}|p{26pt}|}
\hline
\textbf{Problem}& 
\multicolumn{3}{p{100pt}}{\textbf{IFGP--41 constants}} & 
\multicolumn{3}{p{99pt}}{\textbf{IFGP--0 constants }} & 
\multicolumn{3}{p{91pt}}{\textbf{IFGP--81 constants}}  \\
\hline
 & 
\textbf{best}& 
\textbf{mean}& 
\textbf{stddev}& 
\textbf{best}& 
\textbf{mean}& 
\textbf{stddev}& 
\textbf{best}& 
\textbf{mean}& 
\textbf{stddev} \\
\hline
cancer1& 
1.14& 
2.45& 
0.60& 
1.14& 
3.18& 
1.06& 
1.14& 
2.41& 
0.77 \\
\hline
cancer2& 
4.02& 
6.16& 
0.91& 
4.02& 
6.14& 
0.81& 
4.59& 
6.24& 
0.99 \\
\hline
cancer3& 
3.44& 
5.17& 
1.10& 
2.87& 
5.07& 
1.50& 
2.87& 
5.15& 
1.07 \\
\hline
diabetes1& 
22.39& 
25.74& 
2.19& 
21.87& 
26.04& 
1.76& 
21.87& 
25.34& 
2.08 \\
\hline
diabetes2& 
26.04& 
29.91& 
1.65& 
25.00& 
29.21& 
2.21& 
25.52& 
29.82& 
1.30 \\
\hline
diabetes3& 
21.87& 
25.34& 
1.76& 
19.79& 
24.79& 
1.91& 
22.91& 
25.88& 
3.60 \\
\hline
heart1& 
17.33& 
24.44& 
3.76& 
18.67& 
23.28& 
3.33& 
18.66& 
25.28& 
3.64 \\
\hline
heart2& 
1.33& 
6.97& 
4.07& 
1.33& 
\textbf{\textit{4.97}} & 
3.16& 
1.33& 
7.06& 
4.60 \\
\hline
heart3& 
12.00& 
\textbf{\textit{14.00}} & 
2.51& 
10.67& 
15.42& 
3.40& 
9.33& 
\textbf{\textit{15.11}} & 
4.25 \\
\hline
horse1& 
26.37& 
32.16& 
3.12& 
25.27& 
30.69& 
2.49& 
27.47& 
31.57& 
1.91 \\
\hline
horse2& 
30.76& 
\textbf{\textit{35.64}}& 
2.37& 
30.76& 
\textbf{\textit{35.49}} & 
2.81& 
31.86& 
\textbf{\textit{35.71}} & 
2.23 \\
\hline
horse3& 
28.57& 
\textbf{\textit{34.13}}& 
3.14& 
28.57& 
35.67& 
3.90& 
28.57& 
34.90& 
3.53 \\
\hline
\end{tabular}
\end{center}
\end{table}

Table \ref{ifgp_tb5} shows that the best results are obtained when the 
constants are not used in our IFGP system. For 8 (out of 12) cases the best 
result obtained by IFGP outperform the best result obtained by LGP. That 
does not mean that the constants are useless in our model and for the 
considered test problems. An explanation for this behaviour can be found if 
we take a look at the parameters used by IFGP. The population size (of only 
50 individuals) and the chromosome length (of only 30 genes) could not be 
enough to obtain a perfect convergence knowing that some problems have many 
parameters (input variables). For instance the \textit{horse} problem has 58 attributes and a chromosome of 
only 30 genes could not be enough to evolve a complex expression that 
contains sufficient problem's variables and some of the considered 
constants. It is expected that longer chromosomes will increase the 
performances of the IFGP technique.

\section{Conclusion and further work}

An evolutionary technique, Infix Form Genetic Programming (IFGP) has been 
described in this chapter. The IFGP technique has been used for solving several 
classification problems. Numerical experiments show that the error rates 
obtained by using IFGP are similar and sometimes even better than those obtained by Linear Genetic 
Programming. 

Further numerical experiments will try to analyse the relationship between 
the parameters of the IFGP algorithm and the classification error for the 
considered test problems. 

\chapter{Multi Solution Linear Genetic Programming}\label{mslgp_chap}

A new Linear Genetic Programming \cite{brameier1} variant called Multi-Solution Linear Genetic Programming (MS-LGP) is described. Each MS-LGP chromosome encodes multiple solutions of the problem being solved. The best of these solutions is used for fitness assignment purposes.

This chapter is entirely original and it is based on the papers \cite{oltean_mslgp,oltean_improving}.

\section{MS-LGP representation and fitness assignment process}

MS-LGP enrichs LGP structure in two ways:

\begin{itemize}
\item {Each destination variable is allowed to represent the output of the 
program. In the standard LGP only one variable is chosen to provide the 
output.} 
\item{The program output is checked after each instruction in chromosome. Note that within the standard LGP the output is checked after the execution of all instructions in a chromosome.}
\end{itemize}

After each instruction, the value stored in the destination variable is 
considered as a potential solution of the problem. The best value stored in 
one of the destination variables is considered for fitness assignment 
purposes. \\

\textbf{Example}\\

Consider the chromosome $C$ given below:\\

\textsf{\textbf{void}}\textsf{ LGP(}\textsf{\textbf{double}}\textsf{ 
}\textsf{\textit{r}}\textsf{[8])}

\textsf{{\{}}

\hspace{1cm}$r$[5] = $r$[3] * $r$[2];

\hspace{1cm}$r$[3] = $r$[1] + 6;

\hspace{1cm}$r$[0] = $r$[4] * $r$[7];

\hspace{1cm}$r$[6] = $r$[4] -- $r$[1];

\hspace{1cm}$r$[1] = $r$[6] * 7;

\hspace{1cm}$r$[0] = $r$[0] + $r$[4];

\hspace{1cm}$r$[2] = $r$[3] / $r$[4];

{\}}\\

Instead of encoding the output of the problem in a single variable (as in 
SS-LGP) we allow that each of the destination variables ($r$[5], $r$[3], $r$[0], 
r[6], $r$[1] or $r$[2]) to store the program output. The best output stored in 
these variables will provide the fitness of the chromosome.

For instance, if we want to solve symbolic regression problems, the fitness 
of each destination variable $r$[$i$] may be computed using the formula:

\[
f(r[i]) = \sum\limits_{k = 1}^n {\left| {o_{k,i} - w_k } \right|} ,
\]

\noindent
where $o_{k,i}$ is the result obtained in variable $r$[$i$] for the fitness case 
$k$, $w_{k}$ is the targeted result for the fitness case $k$ and $n$ is the number of 
fitness cases. For this problem the fitness needs to be minimized.

The fitness of an individual is set to be equal to the lowest fitness of the 
destination variables encoded in the chromosome:

\[
f(C) = \mathop {\min }\limits_i f(r[i]).
\]

Thus, we have a Multi-Solution program at two levels: 

\begin{itemize}

\item{First level is given 
by the possibility that each variable to represent the output of the program.}

\item{Second level is given by the possibility of checking for the output 
at each instruction in the chromosome.}

\end{itemize}

Our choice was mainly motivated by the No Free Lunch Theorems for Search 
\cite{wolpert1,wolpert2}. There is neither practical nor theoretical evidence that one of the 
variables employed by the LGP is better than the others. More than that, 
Wolpert and McReady \cite{wolpert1} proved that we cannot use the search algorithm's 
behavior so far for a particular test function to predict its future 
behavior on that function.

The Multi-Solution ability has been tested within other evolutionary model 
such as Multi Expression Programming \cite{oltean_mep} or Infix Form Genetic 
Programming \cite{oltean_ifgp}. For these methods it has been shown \cite{oltean_mep} that encoding 
multiple solutions in a single chromosome leads to significant improvements.

\section{Numerical experiments}

In this section several experiments with SS-LGP and MS-LGP are performed. 
For this purpose we use several well-known symbolic regression problems. The 
problems used for assessing the performance of the compared algorithms are:\\

$f_{1}(x)=x^{4}+x^{3}+x^{2}+x$.\\

$f_{2}(x)=x^{6}$ -- 2$x^{4}+x^{2}$.\\

$f_{3}(x)$ = \textit{sin}($x^{4}+x^{2})$.\\

$f_{4}(x)$ = \textit{sin}($x^{4})$ + \textit{sin}($x^{2})$.\\

For each function 20 fitness cases have been randomly generated with a 
uniform distribution over the [0, 1] interval.

The general parameters of the LGP algorithms are given in Table \ref{ehard_tb1}. The same 
settings are used for Multi Solution LGP and for Single-Solution LGP.

\begin{table}[htbp]
\caption{The parameters of the LGP algorithm for symbolic regression problems.}
\label{ehard_tb1}
\begin{center}
\begin{tabular}
{|p{131pt}|p{225pt}|}
\hline
\textbf{Parameter}& 
\textbf{Value} \\
\hline
Number of generations& 
51 \\
\hline
Crossover probability& 
0.9 \\
\hline
Crossover type& 
Uniform \\
\hline
Mutations& 
2 / chromosome \\
\hline
Function set& 
$F$ = {\{}+, -, *, /, \textit{sin}{\}} \\
\hline
Terminal set& 
Problem inputs + 4 supplementary registers  \\
\hline
Constants& 
Not used \\
\hline
Selection& 
Binary Tournament \\
\hline
Algorithm& 
Steady State \\
\hline
\end{tabular}
\end{center}
\end{table}

For all problems the relationship between the success rate and the 
chromosome length and the population size is analyzed. The success rate is 
computed as the number of successful runs over the total number of runs. \\

\textbf{Experiment 1}\\

\begin{figure}[htbp]
\centerline{\includegraphics[width=4.98in,height=5.73in]{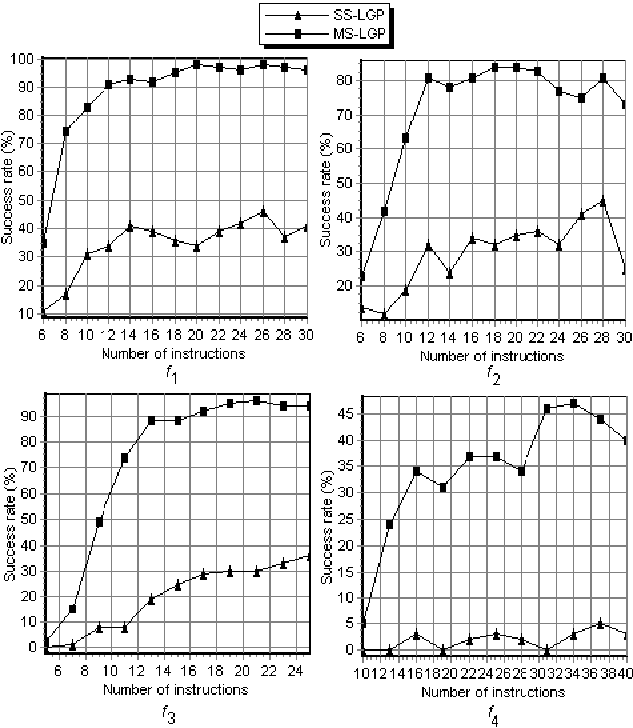}}
\caption{The relationship between the success rate and the number of instructions in a chromosome. Results are averaged over 100 runs.}
\label{ehard1}
\end{figure}

In this experiment the relationship between the success rate and the 
chromosome length is analyzed. For this experiment the population size was 
set to 50 individuals. Other parameters of the LGP algorithms are given in 
Table \ref{ehard_tb1}. Results are depicted in Figure \ref{ehard1}.

Figure \ref{ehard1} shows that Multi-Solution LGP significantly outperforms 
Single-Solution LGP for all the considered test problems and for all the 
considered parameter setting. More than that, large chromosomes are better 
for MS-LGP than short chromosomes. This is due to the multi-solution 
ability: increasing the chromosome length leads to more solutions encoded in 
the same individual.

The easiest problem is $f_{1}$. MS-LGP success rate for this problem is over 
90{\%} when the number of instructions in a chromosome is larger than 12. 
The most difficult problem is $f_{4}$. For this problem and with the 
parameters given in Table \ref{ehard_tb1}, the success rate of the MS-LGP algorithm never 
increases over 47{\%}. However, these results are very good compared to 
those obtained by SS-LGP (the success rate never increases over 5{\%}).\\

\textbf{Experiment 2}\\

\begin{figure}[htbp]
\centerline{\includegraphics[width=4.98in,height=5.77in]{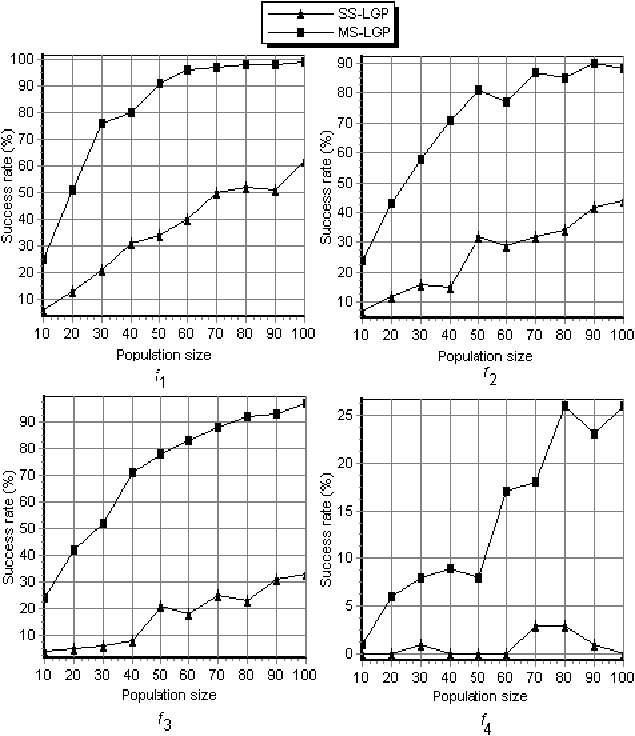}}
\caption{The relationship between the population size and the success rate. Population size varies between 10 and 100. Results are averaged over 100 runs.}
\label{ehard2}
\end{figure}

In this experiment the relationship between the success rate and the 
population size is analyzed. For this experiment the number of instructions 
in a LGP chromosome was set to 12. Other parameters for the LGP algorithms 
are given in Table \ref{ehard_tb1}. Results are depicted in Figure \ref{ehard2}.

Figure \ref{ehard2} also shows that Multi-Solution LGP performs better 
than Single-Solution LGP. Problem $f_{1}$ is the easiest one and problem 
$f_{4}$ is the most difficult one.

\section{Conclusions and further work}

The ability of encoding multiple solutions in a single chromosome has been analyzed in this chapter for Linear Genetic Programming. It has been shown how to efficiently decode the considered chromosomes by traversing them only once. 

Numerical experiments have shown that Multi-Solution LGP significantly improves the evolutionary search for all the considered test problems. There are several reasons for which Multi Solution Programming performs better than Single Solution Programming:

\begin{itemize}

\item{MS-LGP chromosomes act like variable-length chromosomes even if they are stored as fixed-length chromosomes. The variable-length chromosomes are better than fixed-length chromosomes because they can easily store expressions of various complexities,}

\item{MS-LGP algorithms perform more function evaluations than their SS-LGP counterparts. However the complexity of decoding individuals is the same for both MS-LGP and SS-LGP techniques.}

\end{itemize}

The multi-solution ability will be investigated within other evolutionary models. 

\chapter{Evolving Evolutionary Algorithms}\label{eea_chap}

Two new models for evolving Evolutionary Algorithms are described in this chapter. The models are based on Multi Expression Programming and Linear Genetic Programming. Several Evolutionary Algorithms for function optimization and Traveling Salesman Problem are evolved using the proposed models.

This chapter is entirely original and it is based on the papers \cite{oltean_lgp_eea,oltean_mep_eea}.

\section{Introduction}

Evolutionary Algorithms (EAs) \cite{holland1,goldberg1} are nonconventional tools for solving difficult real-world problems. They were developed under the pressure generated by the inability of classical (mathematical) methods to solve some complex real-world problems. Many of these unsolved problems are (or could be turned into) optimization problems. Solving an optimization problem means finding of solutions that maximize or minimize a criteria function \cite{dumitrescu1,holland1,goldberg1}.

Many EAs were proposed for dealing with optimization problems. Many solution representations and search operators were proposed and tested within a wide range of evolutionary models. There are several natural questions that are to be answered in all of these evolutionary models: 

What is the optimal population size?

What is the optimal individual representation?

What are the optimal probabilities for applying specific genetic operators?

What is the optimal number of generations before halting the evolution?

A breakthrough arose in 1995 when Wolpert and McReady unveiled their work on the No Free Lunch (NFL) theorems \cite{wolpert1,wolpert2}. The NFL theorems state that all of the black-box algorithms perform equally well over the entire set of optimization problems. A black-box algorithm does not take into account any information about the problem or the particular instance being solved. 

The magnitudes of the NFL results stroke all of the efforts for developing a universal black-box optimization algorithm able to solve best all the optimization problems.

In their attempt to solve problems, men delegated computers to develop algorithms able to perform certain tasks. The most prominent effort in this direction is Genetic Programming (GP) \cite{koza1}. Instead of evolving solutions for a particular problem instance, GP is mainly intended for discovering computer programs able to solve particular classes of problems. (This statement is only partially true, since the discovery of computer programs may be also viewed as a technique for solving a particular problem instance. The following could be an example of a problem: "Find a computer program that calculates the sum of the elements of an array of integers.")

There are many such approaches so far in the GP literature \cite{koza1,koza2,koza3}. The evolving of deterministic computer programs able to solve specific problems requires a lot of effort.

Instead of evolving deterministic computer programs we evolve a full-featured evolutionary algorithm (i.e. the output of the main program will be an EA able to perform a given task). Proposed approach works with EAs at two levels: 

\begin{itemize}

\item{The first (macro) level consists of a steady-state EA \cite{syswerda1} which uses a fixed population size, a fixed mutation probability, a fixed crossover probability etc.} 

\item{The second (micro) level consists of the solution encoded in a chromosome from the GA on the first level.}

\end{itemize}

We propose  two evolutionary models similar to Multi Expression Programming (MEP) \cite{oltean_mep} and Linear Genetic Programming \cite{brameier1}. These models are very suitable for evolving computer programs that may be easily translated into an imperative language (like \textit{C} or \textit{Pascal}).

\section {Evolving evolutionary algorithms using Multi Expression Programming}

In this section, Multi Expression Programming is used for evolving Evolutionary Algorithms.

\subsection{Evolutionary model}

In order to use MEP for evolving EAs we have to define a set of terminal symbols and a set of function symbols. When we define these sets we have to keep in mind that the value stored by a terminal symbol is independent of other symbols in the chromosome and a function symbol changes the solution stored in another gene. 

An EA usually involves 4 types of genetic operators:

\begin{itemize}
\item {\it Initialize} - randomly initializes a solution,
\item {\it Select} - selects the best solution among several already existing solutions
\item {\it Crossover} - recombines two already existing solutions,
\item {\it Mutate} - varies an already existing solution.
\end{itemize}

These operators act as symbols that may appear into an MEP chromosome. The only operator that generates a solution independent of the already existing solutions is the {\it Initialize} operator. This operator will constitute the terminal set. The other operators will be considered function symbols. Thus, we have:\\

$T$ = \{{\it Initialize}\}, \\

$F$ = \{{\it Select}, {\it Crossover}, {\it Mutate}\}.\\

A MEP chromosome $C$, storing an evolutionary algorithm is:
\\
\\
{\tt
1: {\it Initialize}\hspace{1.88cm}\{Randomly generates a solution.\}\\
2: {\it Initialize}\hspace{1.88cm}\{Randomly generates another solution.\}\\
3: {\it Mutate} 1\hspace{1,70cm}\{Mutates the solution stored on position 1\}\\
4: {\it Select} 1, 3\hspace{1,41cm}\{Selects the best solution from those\}\\
\hspace*{3,82cm}\{stored on positions 1 and 3\}\\
5: {\it Crossover} 2, 4\hspace{0,75cm}\{Recombines the solutions on positions 2 and 4\}\\ 
6: {\it Mutate} 4\hspace{1,79cm}\{Mutates the solution stored on position 4\}\\
7: {\it Mutate} 5\hspace{1,79cm}\{Mutates the solution stored on position 5\}\\
8: {\it Crossover} 2, 6\hspace{0,79cm}\{Recombines the solutions on positions 2 and 6\}\\
}

This MEP chromosome encodes multiple evolutionary algorithms. Each EA is obtained by reading the chromosome bottom up, starting with the current gene and following the links provided by the function pointers. Thus we deal with EAs at two different levels: a micro level representing the evolutionary algorithm encoded in a MEP chromosome and a macro level GA, which evolves MEP individuals. The number of genetic operators (initializations, crossovers, mutations, selections) is not fixed and it may vary between 1 and the MEP chromosome length. These values are automatically discovered by the evolution. The macro level GA execution is bound by the known rules for GAs (see \cite{goldberg1}).

For instance, the chromosome defined above encodes 8 EAs. They are given in Table \ref{mep_eas}.

\begin{table}[htbp]
\caption{Evolutionary Algorithms encoded in the MEP chromosome $C$.}
\label{mep_eas}
\begin{center}
\begin{tabular}
{|p{120pt}|p{120pt}|}
\hline
EA$_{1}$& 
EA$_{2}$ \\
\hline
$i_{1}$=\textit{Initialize} \par & 
$i_{1}$=\textit{Initialize} \\
\hline
EA$_{3}$& 
EA$_{4}$ \\
\hline
$i_{1}$=\textit{Initialize} \par $i_{2}$=\textit{Mutate} ($i_{1})$& 
$i_{1}$=\textit{Initialize} \par $i_{2}$=\textit{Mutate} ($i_{1})$ \par $i_{3}$=\textit{Select} ($i_{1}$, $i_{2})$ \\
\hline
EA$_{5}$& 
EA$_{6}$ \\
\hline
$i_{1}$=\textit{Initialize} \par $i_{2}$=\textit{Initialize} \par $i_{3}$=\textit{Mutate} ($i_{1})$ \par $i_{4}$=\textit{Select} ($i_{1}$, $i_{3})$ \par $i_{5}$=\textit{Crossover} ($i_{2}$, $i_{4})$& 
$i_{1}$=\textit{Initialize} \par $i_{2}$=\textit{Mutate} ($i_{1})$ \par $i_{3}$=\textit{Select} ($i_{1}$, $i_{2})$ \par $i_{4}$=\textit{Mutate} ($i_{3})$ \\
\hline
EA$_{7}$& 
EA$_{8}$ \\
\hline
$i_{1}$=\textit{Initialize} \par $i_{2}$=\textit{Initialize} \par $i_{3}$=\textit{Mutate} ($i_{1})$ \par $i_{4}$=\textit{Select} ($i_{1}$, $i_{3})$ \par $i_{5}$=\textit{Crossover} ($i_{2}$, $i_{4})$ \par $i_{6}$ =\textit{Mutate} ($i_{5})$& 
$i_{1}$=\textit{Initialize} \par $i_{2}$=\textit{Initialize} \par $i_{3}$=\textit{Mutate} ($i_{1})$ \par $i_{4}$=\textit{Select} ($i_{1}$, $i_{3})$ \par $i_{5}$=\textit{Mutate} ($i_{4})$ \par $i_{6}$=\textit{Crossover} ($i_{2}$, $i_{5})$ \\
\hline
\end{tabular}
\end{center}
\end{table}

\textbf{Remarks}\\

\begin{itemize}
\item[$(i)$]In our model the {\it Crossover} operator always generates a single offspring from two parents. The crossover operators generating two offspring may also be designed to fit our evolutionary model.
\item[$(ii)$]The {\it Select} operator acts as a binary tournament selection. The best out of two individuals is always accepted as the selection result.
\item[$(iii)$]	The {\it Initialize, Crossover and Mutate} operators are problem dependent. 
\end{itemize}

\subsection {Fitness assignment}

We have to compute the quality of each EA encoded in the chromosome in order to establish the fitness of a MEP individual. For this purpose each EA encoded in a MEP chromosome is run on the particular problem being solved.

Roughly speaking the fitness of a MEP individual is equal to the fitness of the best solution generated by one of the evolutionary algorithms encoded in that MEP chromosome. But, since the EAs encoded in a MEP chromosome use pseudo-random numbers it is likely that successive runs of the same EA generate completely different solutions. This stability problem is handled in the following manner: each EA encoded in a MEP chromosome is executed (run) more times and the fitness of a MEP chromosome is the average of the fitness of the best EA encoded in that chromosome over all runs. In all of the experiments performed in this section each EA encoded into a MEP chromosome was run 200 times.

\subsection{Numerical experiments}

In this section, we evolve an EA for function optimization. For training purposes we use the Griewangk's function \cite{yao1}.

Griewangk's test function is defined by the equation \ref{Griewangk_eq}.

\begin{equation}
\label{Griewangk_eq}
f(x)=\frac{1}{4000}\sum_{i=1}^{n}x_{i}^{2}-\prod_{i=1}^{n}cos\left(\frac{x_i}{\sqrt{i}}\right) + 1.
\end{equation}

The domain of definition is $[-500, 500]^n$. We use $n = 5$ in this study. The optimal solution is $x_0$ = (0,\dots,0) and $f(x_0)=0$. Griewangk's test function has many widespread local minima which are regularly distributed.

An important issue concerns the representation of the solutions evolved by the EAs encoded in an MEP chromosome and the specific genetic operators used for this purpose. The solutions evolved by the EAs encoded in MEP chromosomes are represented by using real values \cite{goldberg1} (i.e. a chromosome of the second level EA is an array of real values). By initialization, a random point within the definition domain is generated. The convex crossover with $\alpha = \frac12$ and the Gaussian mutation with $\sigma = 0.5$ are used.\\

\textbf{Experiment 1}\\

In this experiment we are interested in seeing the way in which the quality of the best evolved EA improves as the search process advances. MEP algorithm parameters are given in Table \ref{mep_ea1}. 

\begin{figure}[ht]
\centerline{\includegraphics[height=4.97cm]{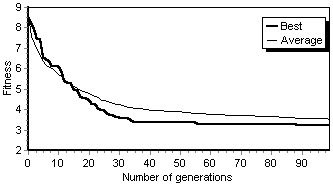}}
\caption{The fitness of the best individual in the best run and the average (over 10 runs) of the fitness of the best individual over all runs.}
\label{Figure1}
\end{figure}

The results of this experiment are depicted in Figure \ref{Figure1}.

\begin{table}[htbp]
\caption{The parameters of the MEP algorithm for Experiment 1.}
\label{mep_ea1}
\begin{center}
\begin{tabular}
{|p{140pt}|p{171pt}|}
\hline
\textbf{Parameter}& 
\textbf{Value} \\
\hline
Population size& 
100 \\
\hline
Code Length& 
3000 genes \\
\hline
Number of generations& 
100 \\
\hline
Crossover probability& 
0.7 \\
\hline
Crossover type& 
Uniform Crossover \\
\hline
Mutation & 
5 mutations per chromosome \\
\hline
Terminal set& 
$F$ = \{\textit{Initialize}\} \\
\hline
Function set& 
$F$ = {\{}\textit{Select}, \textit{Crossover}, \textit{Mutate}{\}} \\
\hline
\end{tabular}
\end{center}
\end{table}

Figure \ref{Figure1} clearly shows the effectiveness of our approach. The MEP technique is able to evolve an EA for solving optimization problems. The quality of the best evolved EA is 8.5 at generation 0. That means that the fitness of the best solution obtained by the best evolved EA is 8.5 (averaged over 200 runs). This is a good result, knowing that the worst solution over the definition domain is about 313. After 100 generations the quality of the best evolved EA is 3.36.\\

\textbf{Experiment 2}\\

We are also interested in seeing how the structure of the best evolved EA changed during the search process.

\begin{figure}[ht]
\centerline{\includegraphics[height=5.42cm]{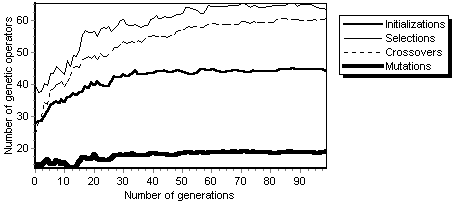}}
\caption{The fitness of the best individual in the best run and the average (over 10 runs) of the fitness of the best individual over all runs.}
\label{Figure2}
\end{figure}

The evolution of the number of the genetic operators used by the best evolved EA is depicted in Figure \ref{Figure2}.

Figure \ref{Figure2} shows that the number of the genetic operators used by the best EA increases as the search process advances. For instance the averaged number of {\it Initializations} in the best EA from generation 0 is 27, while the averaged number of {\it Initializations} in the best evolved EA (after 100 generations) is 43. The averaged number of {\it Mutations} is small (less than 18) when compared to the number of occurrences of other genetic operators.

\section{Evolving evolutionary algorithms with Linear Genetic Programming}

In order to use LGP for evolving EAs we have to modify the structure of an 
LGP chromosome and to define a set of function symbols. This model was proposed in \cite{oltean_lgp_eea}.

\subsection{Individual representation for evolving EAs}

Instead of working with registers, our LGP program will modify an array of 
individuals (the population). In what follows we denote by \textit{Pop} the array of 
individuals (the population) which will be modified by an LGP program. 

The set of function symbols will consist of genetic operators that may 
appear into an evolutionary algorithm. Usually, there are 3 types of genetic 
operators that may appear into an EA. These genetic operators are:

\textit{Select} - that selects the best solution among several already existing solutions,

\textit{Crossover} - that recombine two existing solutions,

\textit{Mutate} - that varies an existing solution.

These operators will act like possible function symbols that may appear into 
a LGP chromosome. Thus, each simple \textbf{\textit{C}} instruction that has 
appeared into a standard LGP chromosome will be replaced by a more complex 
instruction containing genetic operators. More specific, in the modified LGP 
chromosomes we may have three major types of instructions. These 
instructions are:

\textit{Pop}[$k$] = \textit{Select} (\textit{Pop}[$i$], \textit{Pop}[$j$]); \textsf{// Select the best individual from those stored in }

\textsf{// }\textsf{\textit{Pop}}\textsf{[}\textsf{\textit{i}}\textsf{] and 
}\textsf{\textit{Pop}}\textsf{[}\textsf{\textit{j}}\textsf{] and keep the 
result in position }\textsf{\textit{k}}\textsf{.}

\textit{Pop}[$k$] = \textit{Crossover} (\textit{Pop}[$i$], \textit{Pop}[$j$]); \textsf{// Crossover the individuals stored in }\textsf{\textit{Pop}}\textsf{[}\textsf{\textit{i}}\textsf{] and 
}\textsf{\textit{Pop}}\textsf{[}\textsf{\textit{j}}\textsf{] }

\textsf{// and keep the result in position }\textsf{\textit{k}}\textsf{.}

\textit{Pop}[$k$] = \textit{Mutate} (\textit{Pop}[$i$]); \textsf{// Mutate the individual stored in }

\textsf{// position }\textsf{\textit{i}}\textsf{ and keep the result in 
position }\textsf{\textit{k}}\textsf{.}\\

An LGP chromosome $C$, storing an evolutionary algorithm is the following.\\

\textsf{\textbf{void}}\textsf{ 
LGP{\_}Program(}\textsf{\textbf{Chromosome}}\textsf{ Pop[8]) // a population 
with 8 individuals}

\textsf{{\{}}

\textsf{...}

\textsf{Pop[0] = Mutate(Pop[5]);}

\textsf{Pop[7] = Select(Pop[3], Pop[6]);}

\textsf{Pop[4] = Mutate(Pop[2]);}

\textsf{Pop[2] = Crossover(Pop[0], Pop[2]);}

\textsf{Pop[6] = Mutate(Pop[1]);}

\textsf{Pop[2] = Select(Pop[4], Pop[3]);}

\textsf{Pop[1] = Mutate(Pop[6]);}

\textsf{Pop[3] = Crossover(Pop[5], Pop[1]);}

\textsf{... }

\textsf{{\}}}\\

These statements will be considered as genetic operations that are executed 
during an EA generation. Since our purpose is to evolve a generational EA we 
have to add a wrapper loop around the genetic operations that are executed 
during an EA generation. More than that, each EA starts with a random 
population of individuals. Thus, the LGP program must contain some 
instructions that initialize the initial population. 

The obtained LGP chromosome is given below:\\

\textsf{\textbf{void}}\textsf{ 
LGP{\_}Program(}\textsf{\textbf{Chromosome}}\textsf{ Pop[8]) // a population 
consisting of 8 individuals}

\textsf{{\{}}

\textsf{Randomly{\_}initialize{\_}the{\_}population();}

\textsf{\textbf{for}}\textsf{ (}\textsf{\textbf{int}}\textsf{ k = 0; k < 
MaxGenerations; k++){\{} // repeat for a fixed }

\textsf{// number of generations}

\textsf{Pop[0] = Mutate(Pop[5]);}

\textsf{Pop[7] = Select(Pop[3], Pop[6]);}

\textsf{Pop[4] = Mutate(Pop[2]);}

\textsf{Pop[2] = Crossover(Pop[0], Pop[2]);}

\textsf{Pop[6] = Mutate(Pop[1]);}

\textsf{Pop[2] = Select(Pop[4], Pop[3]);}

\textsf{Pop[1] = Mutate(Pop[6]);}

\textsf{Pop[3] = Crossover(Pop[5], Pop[1]);}

\textsf{{\}}}

\textsf{{\}}}\\

\textbf{Remark}\\

The initialization function and the \textbf{for} 
cycle will not be affected by the genetic operators. These parts are kept 
unchanged during the search process.

\subsection{Fitness assignment}

We deal with EAs at two different levels: a micro level representing the 
evolutionary algorithm encoded into a LGP chromosome and a macro level GA, 
which evolves LGP individuals. Macro level GA execution is bounded by known 
rules for GAs \cite{brameier1}. 

For computing the fitness of a LGP individual we have to compute the quality 
of the EA encoded in that chromosome. For this purpose the EA encoded into a 
LGP chromosome is run on the particular problem being solved.

Roughly speaking the fitness of a LGP individual is equal to the fitness of 
the best solution generated by the evolutionary algorithm encoded into that 
LGP chromosome. But, since the EA encoded into a LGP chromosome use 
pseudo-random numbers it is very possible as successive runs of the same EA 
to generate completely different solutions. This stability problem is 
handled in a standard manner: the EA encoded into a LGP chromosome is 
executed (run) more times (in fact 200 runs are executed in all the 
experiments performed for evolving EAs for function optimization and 15 runs 
for evolving EAs for the TSP) and the fitness of a LGP chromosome is the 
average of the fitness of the EA encoded in that chromosome over all runs.\\

\textbf{Remark}\\

In the standard LGP one of the registers is chosen 
as the program output. This register is not changed during the search 
process. In our approach the register storing the best value (best fitness) 
is chosen to represent the chromosome. Thus, each LGP chromosome stores 
multiple solutions of a problem in the same manner as Multi Expression 
Programming \cite{oltean_mep}.

\subsection{The model used for evolving EAs}

For evolving EAs we use the steady state algorithm \cite{syswerda1}. For increasing the generalization ability (e.g. the ability of 
the evolved EA to yield good results on new test problems), the problem set 
has been divided into three sets, suggestively called training set, 
validation set and test set (see \cite{prechelt1}). In our experiments the training set 
consists of a difficult test problem. Validation is performed using another 
difficult test problem. The test set consists of other well-known 
benchmarking problems.

A method called \textit{early stopping} is used to avoid overfitting of the population individuals 
to the particular training examples used \cite{prechelt1}. This method consists of 
computing the test set performance for that chromosome which had the minimum 
validation error during the search process. Using the early stopping 
technique will increase the generalization performance \cite{prechelt1}.

The test set consists of several well-known benchmarking problems \cite{reinelt1,yao1} 
used for assessing the performances of the evolutionary algorithms.

\subsection{Evolving EAs for function optimization}

\subsubsection{Test functions}

Ten test problems $f_{1}-f_{10}$ (given in Table \ref{lgp_ea1}) are used to asses the 
performance of the evolved EA. Functions $f_{1}-f_{6}$ are unimodal test 
function. Functions $f_{7}-f_{10}$ are highly multimodal (the number of 
local minima increases exponentially with the problem dimension \cite{yao1}).

\begin{table}[htbp]
\caption{Test functions used in our experimental study. The parameter $n$ is the space dimension ($n$ = 5 in our numerical experiments) and $f_{min}$ is the minimum value of the function.}
\label{lgp_ea1}
\begin{center}
\begin{tabular}
{|p{234pt}|p{60pt}|p{60pt}|}
\hline
Test function& 
Domain& 
$f_{min}$ \\
\hline
$f_1 (x) = \sum\limits_{i = 1}^n {(i \cdot x_i^2 )} .$& 
[-5, 5]$^{ n}$& 
0 \\
\hline
$f_2 (x) = 10 \cdot n + \sum\limits_{i = 1}^n {(x_i^2 - 10 \cdot \cos (2 \cdot \pi \cdot x_i ))} $& 
[-5, 5]$^{ n}$& 
0 \\
\hline
$f_3 (x) = - a \cdot e^{ - b\sqrt {\frac{\sum\limits_{i = 1}^n {x_i^2 } }{n}} } - e^{\frac{\sum {\cos (c \cdot x_i )} }{n}} + a + e.$& 
[-32, 32]$^{ n}$ \par $a$ = 20, $b$ = 0.2, $c$ = 2\textit{$\pi $}.& 
0 \\
\hline
$f_4 (x) = \frac{1}{4000} \cdot \sum\limits_{i = 1}^n {x_i^2 - \prod\limits_{i = 1}^n {\cos (\frac{x_i }{\sqrt i }) + 1} } .$& 
[-500, 500]$^{ n}$& 
0 \\
\hline
$f_5 (x) = \sum\limits_{i = 1}^n {( - x_i \cdot \sin (\sqrt {\left| {x_i } \right|} ))} $& 
[-500, 500]$^{ n}$& 
-$n \quad  \bullet $ 418.9829 \\
\hline
$f_6 (x) = \sum\limits_{i = 1}^n {x_i^2 } .$& 
[-100, 100]$^{ n}$& 
0 \\
\hline
$f_7 (x) = \sum\limits_{i = 1}^n {\vert x_i \vert + \prod\limits_{i = 1}^n {\vert x_i \vert } } .$& 
[-10, 10]$^{ n}$& 
0 \\
\hline
$f_8 (x) = \sum\limits_{i = 1}^n {\left( {\sum\limits_{j = 1}^i {x_j^2 } } \right)} .$& 
[-100, 100$^{ }$]$^{ n}$& 
0 \\
\hline
$f_9 (x) = \max _i \{x_i ,1 \le i \le n\}.$& 
[-100, 100]$^{ n}$& 
0 \\
\hline
$f_{10} (x) = \sum\limits_{i = 1}^{n - 1} {100 \cdot (x_{i + 1} - x_i^2 )^2 + (1 - x_i )^2} .$& 
[-30, 30]$^{ n}$& 
0 \\
\hline
\end{tabular}
\end{center}
\end{table}

\subsubsection{Experimental results}\label{ea1}

In this section we evolve an EA for function optimization and then we asses 
the performance of the evolved EA. A comparison with a standard GA is 
performed later in this section.

For evolving an EA we use $f_{2}$ as the training problem and the function 
$f_{3}$ as the validation problem.

An important issue regards the solutions evolved by the EAs encoded into a 
LGP chromosome and the specific genetic operators used for this purpose. 
Solutions evolved by the EA encoded into LGP chromosomes are represented 
using real values \cite{goldberg1}. By initialization, a point within the definition 
domain is randomly generated. Convex crossover with $\alpha $ = 
$\raise.5ex\hbox{$\scriptstyle 1$}\kern-.1em/ 
\kern-.15em\lower.25ex\hbox{$\scriptstyle 2$} $ and Gaussian mutation with 
$\sigma $ = 0.5 are used (for more information on real encoding and specific 
operators see \cite{dumitrescu1}).\\

\textbf{Experiment 1}\\

In this experiment, an EA for function optimization is evolved. 

There is a wide range of EAs that can be evolved using the technique 
described above. Since, the evolved EA has to be compared with another 
algorithm (such as a standard GA or an ES), the parameters of the evolved EA 
have to be similar to the parameters of the algorithm used for comparison.

For instance, a standard GA uses a primary population of $N$ individuals and an 
additional population (the new population) that stores the offspring 
obtained by crossover and mutation. Thus, the memory requirements for a 
standard GA is O(2*$N)$. In each generation there will be 2 * $N$ Selections, $N$ 
Crossovers and $N$ Mutations (we assume here that only one offspring is 
obtained by crossover of two parents). Thus, the number of genetic operators 
(\textit{Crossovers}, \textit{Mutations} and \textit{Selections}) in a standard GA is 4 * $N$. We do not take into account the 
complexity of the genetic operators, since in most of the cases this 
complexity is different from operator to operator. The standard GA algorithm 
is given below:\\

\begin{center}
\textbf{Standard GA algorithm}
\end{center}

\textsf{S}$_{1}$\textsf{. Randomly create the initial population 
}\textsf{\textit{P}}\textsf{(0)}

\textsf{S}$_{2}$\textsf{. }\textsf{\textbf{for}}\textsf{ 
}\textsf{\textit{t}}\textsf{ = 1 }\textsf{\textbf{to}}\textsf{ 
}\textsf{\textit{Max}}\textsf{{\_}}\textsf{\textit{Generations}}\textsf{ 
}\textsf{\textbf{do}}

\textsf{S}$_{3}$\textsf{. 
}\textsf{\textit{P'}}\textsf{(}\textsf{\textit{t}}\textsf{) = $\phi $;}

\textsf{S}$_{4}$\textsf{. }\textsf{\textbf{for}}\textsf{ 
}\textsf{\textit{k}}\textsf{ = 1 }\textsf{\textbf{to}}\textsf{ $\vert 
$}\textsf{\textit{P}}\textsf{(}\textsf{\textit{t}}\textsf{)$\vert $ 
}\textsf{\textbf{do}}

\textsf{S}$_{5}$\textsf{. }\textsf{\textit{p}}$_{1}$\textsf{ = 
}\textsf{\textit{Select}}\textsf{(}\textsf{\textit{P}}\textsf{(}\textsf{\textit{t}}\textsf{)); 
// select an individual from the mating pool}

\textsf{S}$_{6}$\textsf{. }\textsf{\textit{p}}$_{2}$\textsf{ = 
}\textsf{\textit{Select}}\textsf{(}\textsf{\textit{P}}\textsf{(}\textsf{\textit{t}}\textsf{)); 
// select the second individual }

\textsf{S}$_{7}$\textsf{. }\textsf{\textit{Crossover}}\textsf{ 
(}\textsf{\textit{p}}$_{1}$\textsf{, }\textsf{\textit{p}}$_{2}$\textsf{, 
}\textsf{\textit{offsp}}\textsf{); // crossover the parents p}$_{1}$\textsf{ 
and p}$_{2}$

\textsf{// the offspring }\textsf{\textit{offspr}}\textsf{ is obtained}

\textsf{S}$_{8}$\textsf{. }\textsf{\textit{Mutation}}\textsf{ 
(}\textsf{\textit{offspr}}\textsf{); // mutate the offspring 
}\textsf{\textit{offspr}}

\textsf{S}$_{9}$\textsf{. Add }\textsf{\textit{offspf}}\textsf{ to 
}\textsf{\textit{P'}}\textsf{(}\textsf{\textit{t}}\textsf{);}

\textsf{S}$_{10}$\textsf{. }\textsf{\textbf{endfor}}

\textsf{S11. }\textsf{\textit{P}}\textsf{(}\textsf{\textit{t}}\textsf{+1) = 
}\textsf{\textit{P}}\textsf{'(}\textsf{\textit{t}}\textsf{);}

\textsf{S}$_{12}$\textsf{. }\textsf{\textbf{endfor}}\\

Rewritten as an LGP program, the Standard GA is given below. The individuals 
of the standard (main) population are indexed from 0 to \textit{PopSize} -- 1 and the 
individuals of the new population are indexed from \textit{PopSize} up to 2 * \textit{PopSize} -- 1.\\

\textsf{\textbf{void}}\textsf{ 
}\textsf{\textit{LGP{\_}Program}}\textsf{(}\textsf{\textbf{Chromosome}}\textsf{ 
}\textsf{\textit{Pop}}\textsf{[2 * }\textsf{\textit{PopSize}}\textsf{]) }

\textsf{//an array containing of 2 * }\textsf{\textit{PopSize}}\textsf{ 
individuals}

\textsf{{\{}}

\textsf{Randomly{\_}initialize{\_}the{\_}population();}

\textsf{\textbf{for}}\textsf{ (}\textsf{\textbf{int}}\textsf{ k = 0; k < 
}\textsf{\textit{MaxGenerations}}\textsf{; k++){\{} // repeat for a fixed }

\textsf{// number of generations}

\textsf{// create the new population}

\textsf{p1 = 
}\textsf{\textit{Select}}\textsf{(}\textsf{\textit{Pop}}\textsf{[3], 
}\textsf{\textit{Pop}}\textsf{[6]);}

\textsf{p2 = 
}\textsf{\textit{Select}}\textsf{(}\textsf{\textit{Pop}}\textsf{[7], 
}\textsf{\textit{Pop}}\textsf{[7]);}

\textsf{o = }\textsf{\textit{Crossover}}\textsf{(p1, p2);}

\textsf{\textit{Pop}}\textsf{[}\textsf{\textit{PopSize}}\textsf{] = 
}\textsf{\textit{Mutate}}\textsf{(o);}

\textsf{p1 = 
}\textsf{\textit{Select}}\textsf{(}\textsf{\textit{Pop}}\textsf{[3], 
}\textsf{\textit{Pop}}\textsf{[6]);}

\textsf{p2 = 
}\textsf{\textit{Select}}\textsf{(}\textsf{\textit{Pop}}\textsf{[7], 
}\textsf{\textit{Pop}}\textsf{[7]);}

\textsf{o = }\textsf{\textit{Crossover}}\textsf{(p1, p2);}

\textsf{\textit{Pop}}\textsf{[}\textsf{\textit{PopSize}}\textsf{ + 1] = 
}\textsf{\textit{Mutate}}\textsf{(o);}

\textsf{p1 = 
}\textsf{\textit{Select}}\textsf{(}\textsf{\textit{Pop}}\textsf{[3], 
}\textsf{\textit{Pop}}\textsf{[6]);}

\textsf{p2 = 
}\textsf{\textit{Select}}\textsf{(}\textsf{\textit{Pop}}\textsf{[7], 
}\textsf{\textit{Pop}}\textsf{[7]);}

\textsf{o = }\textsf{\textit{Crossover}}\textsf{(p1, p2);}

\textsf{\textit{Pop}}\textsf{[}\textsf{\textit{PopSize}}\textsf{ + 2] = 
}\textsf{\textit{Mutate}}\textsf{(o);}

\textsf{...}

\textsf{p1 = 
}\textsf{\textit{Select}}\textsf{(}\textsf{\textit{Pop}}\textsf{[3], 
}\textsf{\textit{Pop}}\textsf{[6]);}

\textsf{p2 = 
}\textsf{\textit{Select}}\textsf{(}\textsf{\textit{Pop}}\textsf{[7], 
}\textsf{\textit{Pop}}\textsf{[7]);}

\textsf{o = }\textsf{\textit{Crossover}}\textsf{(p1, p2);}

\textsf{\textit{Pop}}\textsf{[2 * }\textsf{\textit{PopSize}}\textsf{ - 1] = 
}\textsf{\textit{Mutate}}\textsf{(o);}

\textsf{// pop(}\textsf{\textit{t}}\textsf{ + 1) = new{\_}pop 
(}\textsf{\textit{t}}\textsf{)}

\textsf{// copy the individuals from new{\_}pop to the next population}

\textsf{\textit{Pop}}\textsf{[0] = 
}\textsf{\textit{Pop}}\textsf{[}\textsf{\textit{PopSize}}\textsf{];}

\textsf{\textit{Pop}}\textsf{[1] = 
}\textsf{\textit{Pop}}\textsf{[}\textsf{\textit{PopSize}}\textsf{];}

\textsf{\textit{Pop}}\textsf{[2] = 
}\textsf{\textit{Pop}}\textsf{[}\textsf{\textit{PopSize}}\textsf{];}

\textsf{...}

\textsf{\textit{Pop}}\textsf{[}\textsf{\textit{PopSize}}\textsf{ - 1] = 
Pop[2 * }\textsf{\textit{PopSize}}\textsf{ - 1];}

\textsf{{\}}}

\textsf{{\}}}\\

The parameters of the standard GA are given in Table \ref{lgp_ea2}.

\begin{table}[htbp]
\caption{The parameters of a standard GA for Experiment 1.}
\label{lgp_ea2}
\begin{center}
\begin{tabular}
{|p{140pt}|p{175pt}|}
\hline
\textbf{Parameter}& 
\textbf{Value} \\
\hline
Population size& 
20 (+ 20 individuals in the new pop) \\
\hline
Individual encoding& 
Real \\
\hline
Number of generations& 
100 \\
\hline
Crossover probability& 
1 \\
\hline
Crossover type& 
Convex Crossover with $\alpha $ = 0.5 \\
\hline
Mutation & 
Gaussian mutation with $\sigma $ = 0.01  \\
\hline
Selection& 
Binary Tournament \\
\hline
\end{tabular}
\end{center}
\end{table}

We will evolve an EA that uses the same memory requirements and the same 
number of genetic operations as the standard GA described above. \\

\textbf{Remark}\\

Our comparison is based on the memory requirements 
(i.e. the population size) and the number of genetic operators used during 
the search process. A better comparison could be made if we take into 
account only the number of function evaluations performed during the search. 
Unfortunately, this comparison cannot be realized in our model since we 
cannot control the number of function evaluations (this number is decided by 
the evolution). The total number of genetic operators (crossovers + 
mutations + selections) is the only parameter that can be controlled in our 
model.

The parameters of the LGP algorithm are given in Table \ref{lgp_ea3}.

\begin{table}[htbp]
\caption{The parameters of the LGP algorithm used for Experiment 1.}
\label{lgp_ea3}
\begin{center}
\begin{tabular}
{|p{140pt}|p{171pt}|}
\hline
\textbf{Parameter}& 
\textbf{Value} \\
\hline
Population size& 
500 \\
\hline
Code Length& 
80 instructions \\
\hline
Number of generations& 
100 \\
\hline
Crossover probability& 
0.7 \\
\hline
Crossover type& 
Uniform Crossover \\
\hline
Mutation & 
5 mutations per chromosome \\
\hline
Function set& 
$F$ = {\{}\textit{Select}, \textit{Crossover}, \textit{Mutate}{\}} \\
\hline
\end{tabular}
\end{center}
\end{table}

The parameters of the evolved EA are given in Table \ref{lgp_ea4}.

\begin{table}[htbp]
\caption{The parameters of the evolved EA for function optimization.}
\label{lgp_ea4}
\begin{center}
\begin{tabular}
{|p{140pt}|p{189pt}|}
\hline
\textbf{Parameter}& 
\textbf{Value} \\
\hline
Population size& 
40 \\
\hline
Number of generations& 
100 \\
\hline
Crossover probability& 
1 \\
\hline
Crossover type& 
Convex Crossover with $\alpha $ = 0.5 \\
\hline
Mutation & 
Gaussian mutation with $\sigma $ = 0.01  \\
\hline
Selection& 
Binary Tournament \\
\hline
\end{tabular}
\end{center}
\end{table}

\begin{figure}[htbp]
\centerline{\includegraphics[width=3.52in,height=3.16in]{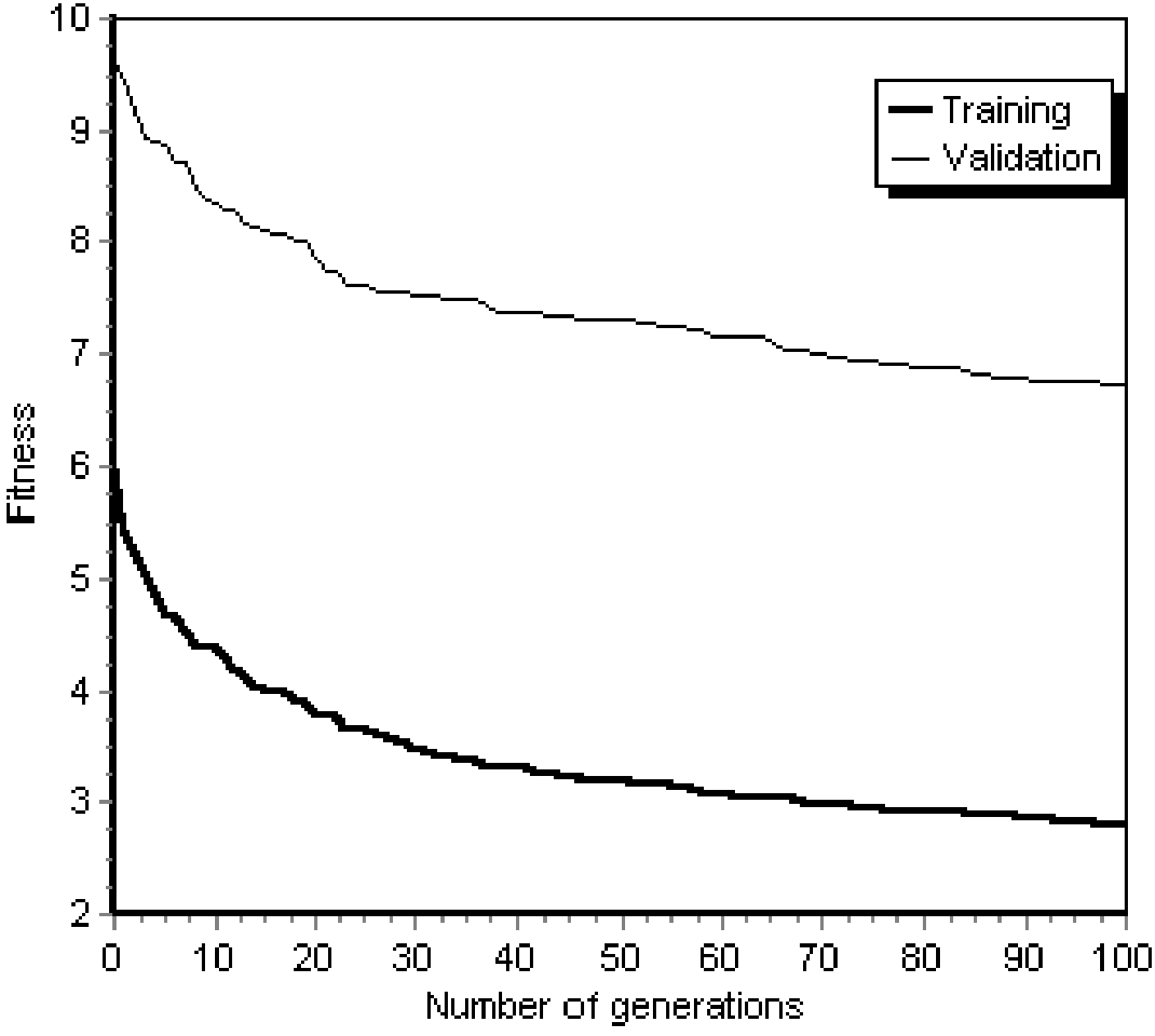}}
\caption{The relationship between the fitness of the best individual in each generation and the number of generations. Results are averaged over 10 runs.}
\label{lgpeas1}
\end{figure}

The results of this experiment are depicted in Figure \ref{lgpeas1}.

Figure \ref{lgpeas1} shows the effectiveness of our approach. LGP 
technique is able to evolve an EA for solving optimization problems. The 
quality of the evolved EA improves as the search process advances.\\

\textbf{Experiment 2}\\

In this experiment we compare the evolved EA to the standard Genetic 
Algorithm described in Experiment 1. The parameters used by the evolved EA 
are given in Table \ref{lgp_ea4} and the parameters used by the standard GA are given in 
Table \ref{lgp_ea2}. The results of the comparison are given in Table \ref{lgp_ea5}.

\begin{table}[htbp]
\begin{center}
\caption{The results of applying the Evolved EA and the Standard GA 
for the considered test functions. StdDev stands for the standard deviation. 
The results are averaged over 30 runs.}
\label{lgp_ea5}
\begin{tabular}
{|p{60pt}|p{60pt}|p{60pt}|p{60pt}|p{60pt}|}
\hline
\raisebox{-1.50ex}[0cm][0cm]{\textbf{Test function}}& 
\multicolumn{2}{|p{127pt}|}{\textbf{Evolved EA} \par \textbf{40 individuals}} & 
\multicolumn{2}{|p{141pt}|}{\textbf{Standard GA} \par \textbf{20 individuals in the standard population + 20 individuals in the new population}}  \\
\cline{2-5} 
 & 
Mean& 
StdDev& 
Mean& 
StdDev \\
\hline
$f_{1}$& 
0.6152& 
0.8406& 
3.1636& 
3.7997 \\
\hline
$f_{2}$& 
2.6016& 
1.7073& 
5.8268& 
3.9453 \\
\hline
$f_{3}$& 
7.5945& 
2.5006& 
10.8979& 
2.7603 \\
\hline
$f_{4}$& 
2.4639& 
1.6389& 
6.0176& 
4.4822 \\
\hline
$f_{5}$& 
-552.0043& 
218.8526& 
-288.3484& 
200.5584 \\
\hline
$f_{6}$& 
273.7000& 
235.7794& 
817.1237& 
699.2686 \\
\hline
$f_{7}$& 
2.0521& 
1.1694& 
4.8836& 
2.4269 \\
\hline
$f_{8}$& 
340.2770& 
348.3748& 
639.2252& 
486.7850 \\
\hline
$f_{9}$& 
10.3317& 
4.2009& 
20.6574& 
8.8268 \\
\hline
$f_{10}$& 
10123.3083& 
18645.7247& 
208900.5717& 
444827.6967 \\
\hline
\end{tabular}
\end{center}
\end{table}

From Table \ref{lgp_ea5} it can be seen that the Evolved EA significantly outperforms 
the standard GA on all of the considered test problems. 

To avoid any suspicion regarding the Evolved EA we will compare it with a GA 
that uses the same standard population size as the Evolved EA. Thus, the 
standard GA will use a double population (a standard population and a new 
population) vis-\`{a}-vis the population employed by the Evolved EA. Note 
that this will provide a significant advantage of the standard GA over the 
Evolved EA. However, we use this larger population because, in this case, 
both algorithms (the Standard GA and the Evolved EA) have one important 
parameter in common: they perform the same number of initializations. 
Results are presented in Table \ref{lgp_ea7}.

\begin{table}[htbp]
\caption{The results of applying the Evolved EA and the Standard GA for the considered test functions. StdDev stands for the standard deviation. Results are averaged over 30 runs.}
\label{lgp_ea7}
\begin{center}
\begin{tabular}
{|p{60pt}|p{60pt}|p{60pt}|p{60pt}|p{60pt}|}
\hline
\raisebox{-1.50ex}[0cm][0cm]{\textbf{Test function }}& 
\multicolumn{2}{|p{110pt}|}{\textbf{Evolved EA} \par \textbf{40 individuals}} & 
\multicolumn{2}{|p{110pt}|}{\textbf{Standard GA} \par \textbf{40 individuals in the standard population + 40 individuals in the new population}}  \\
\cline{2-5} 
 & 
Mean& 
StdDev& 
Mean& 
StdDev \\
\hline
$f_{1}$& 
0.6152& 
0.8406& 
0.2978& 
0.5221 \\
\hline
$f_{2}$& 
2.6016& 
1.7073& 
3.4031& 
2.7188 \\
\hline
$f_{3}$& 
7.5945& 
2.5006& 
6.2529& 
2.8255 \\
\hline
$f_{4}$& 
2.4639& 
1.6389& 
2.4669& 
1.5651 \\
\hline
$f_{5}$& 
-552.0043& 
218.8526& 
-287.0752& 
156.5294 \\
\hline
$f_{6}$& 
273.7000& 
235.7794& 
263.6049& 
239.6022 \\
\hline
$f_{7}$& 
2.0521& 
1.1694& 
2.0366& 
1.5072 \\
\hline
$f_{8}$& 
340.2770& 
348.3748& 
285.9284& 
254.9170 \\
\hline
$f_{9}$& 
10.3317& 
4.2009& 
10.3776& 
5.9560 \\
\hline
$f_{10}$& 
10123.3083& 
18645.7247& 
9102.8337& 
23981.1050 \\
\hline
\end{tabular}
\end{center}
\end{table}

Form Table \ref{lgp_ea7} it can be seen that the Evolved EA is better than the standard 
GA in 4 cases (out of 10). However, in this case the standard GA has a 
considerable advantage over the Evolved EA.\\

\textbf{Experiment 3}\\

We are also interested to analyze the relationship between the number of 
generations of the evolved EA and the quality of the solutions obtained by 
applying the evolved EA for the considered test functions. The parameters of 
the Evolved EA (EEA) are given in Table \ref{lgp_ea4} and the parameters of the Standard 
GA (SGA) are given in Table \ref{lgp_ea2}.

\begin{figure}[htbp]
\centerline{\includegraphics[width=6.71in,height=7.49in]{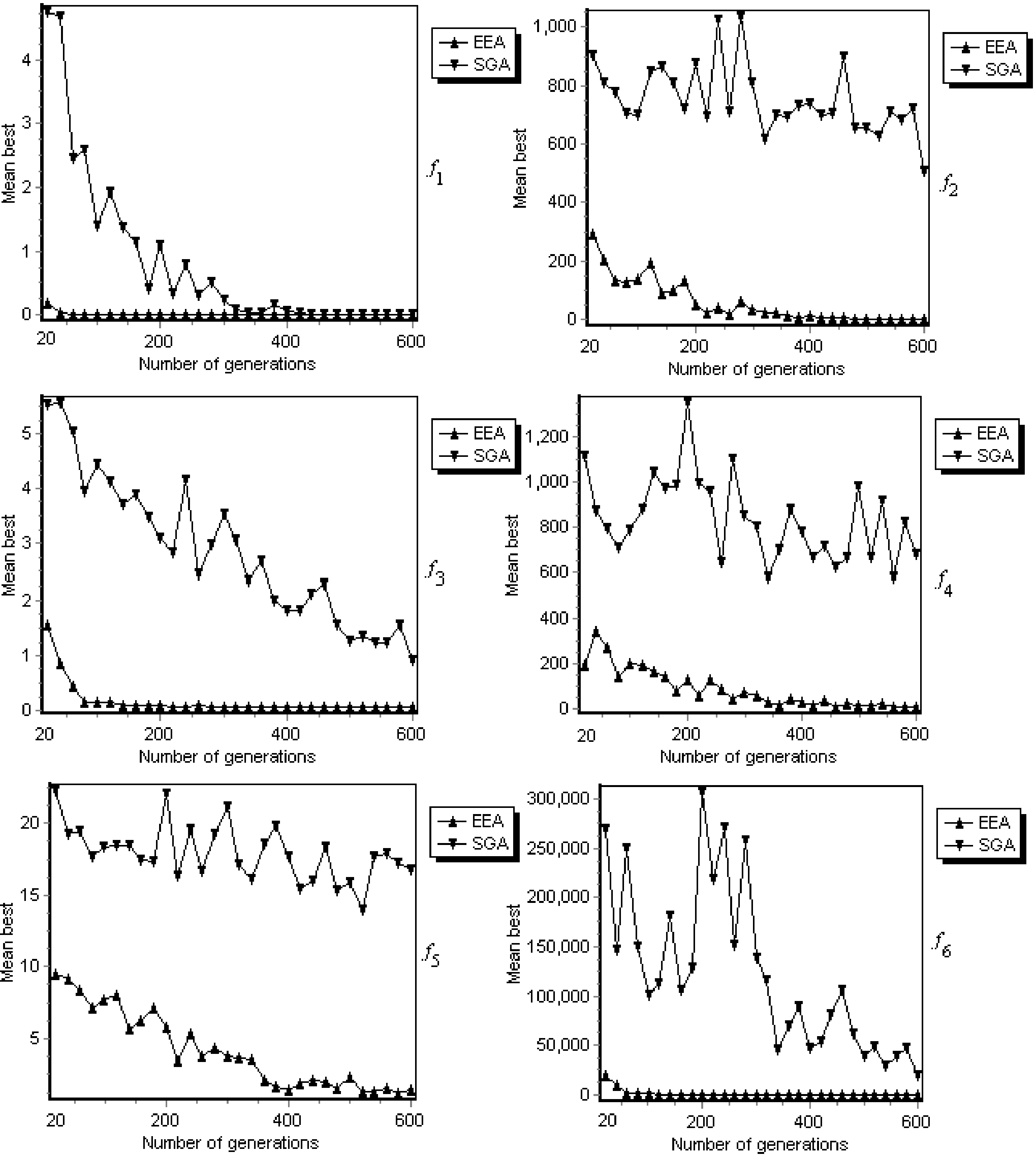}}
\caption{The relationship between the number of generations and the quality of the solutions obtained by the Evolved EA (EEA) and by the Standard GA (SGA) for the unimodal test functions $f_{1}-f_{6}$. The number of generations varies between 20 and 600. Results are averaged over 30 runs.}
\label{LGPEAs2}
\end{figure}

\begin{figure}[htbp]
\centerline{\includegraphics[width=6.77in,height=5.01in]{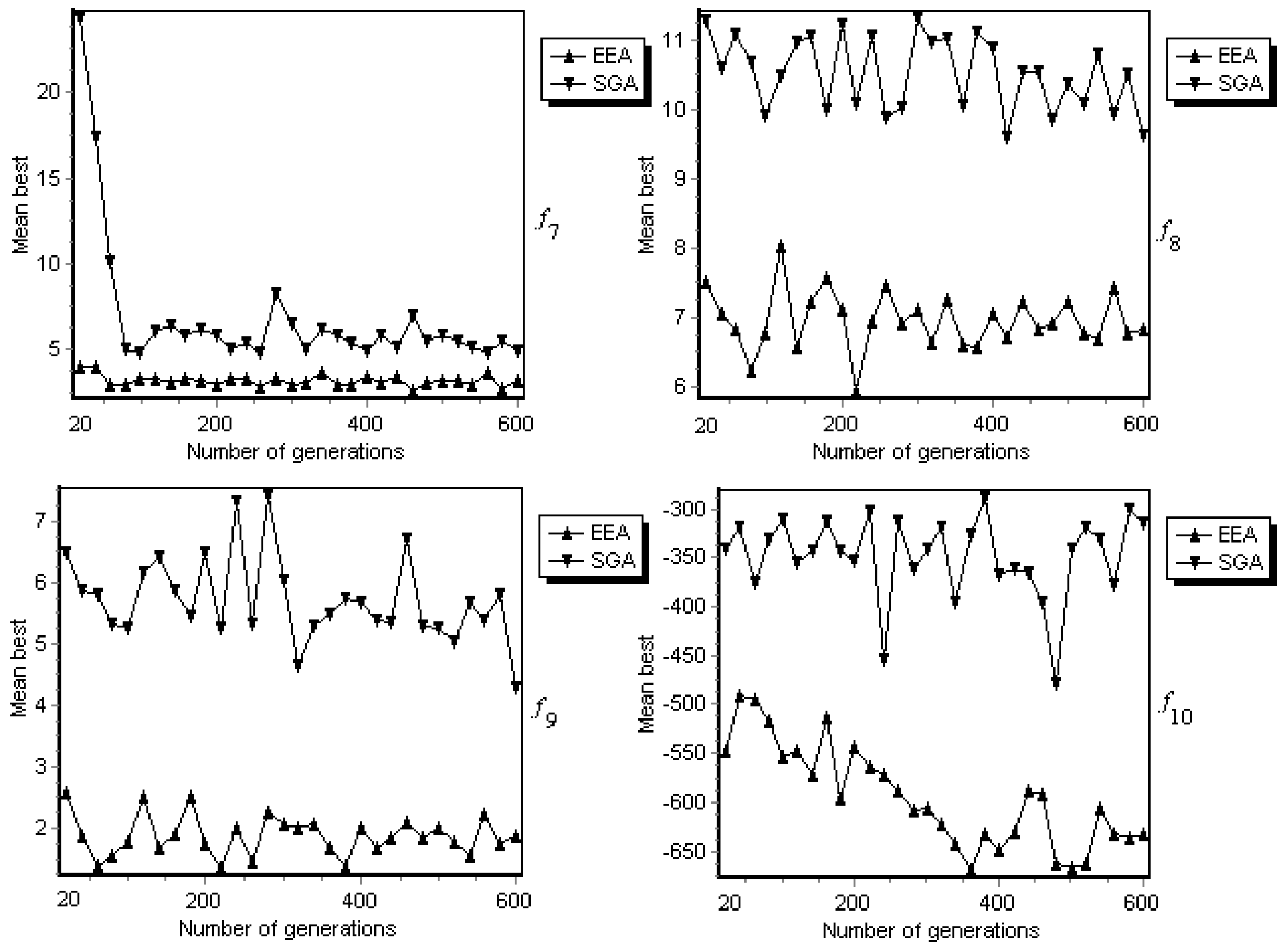}}
\caption{The relationship between the number of generations and the quality of the solutions obtained by the Evolved EA (EEA) and by the Standard GA (SGA) for the multimodal test functions $f_{7}-f_{10}$. The number of generations varies between 20 and 600. Results are averaged over 30 runs.}
\label{LGPEAs3}
\end{figure}

The results of this experiment are depicted in Figure \ref{LGPEAs2} (for the unimodal 
test functions) and in Figure \ref{LGPEAs3} (for the multimodal test functions).

Figures \ref{LGPEAs2} and \ref{LGPEAs3} show that the Evolved EA is scalable in 
comparison with the number of generations. It also can be sent that the 
Evolved EA outperforms the standard GA for all the considered number of 
generations. For the unimodal test functions ($f_{1}-f_{6})$ we can see a 
continuous improvement tendency during the search process. For the 
multimodal test functions we can see a similar behavior only for the test 
function $f_{10}$.

\subsection{Evolving EAs for TSP}

In this section, an Evolutionary Algorithm for solving the Traveling 
Salesman Problem \cite{garey1,merz3,krasnogor1} is evolved. First of all, an EA is evolved and its performance is assessed by 
running it on several well-known instances in TSPLIB \cite{reinelt1}.\\

This section is entirely original and it is based on the papers \cite{oltean_tsp}.

\textbf{Experiment 5}\\

In this experiment, an EA for the TSP problem is evolved.

A TSP path will be represented as a permutation of cities \cite{merz1} and it is 
initialized by using the Nearest Neighbor heuristic \cite{cormen1,garey1}. Genetic 
operators used by the Evolved EA are DPX as crossover \cite{merz2} and 2-Exchange \cite{krasnogor1} as 
mutation. These operators are briefly described in what follows.

DPX recombination operator copies into offspring all the common edges of the 
parents. Then it completes the offspring to achieve a valid tour with links 
that do not belong to the parents, in such way that the distance between 
parents in the newly created offspring is preserved. This completion may be 
done by using nearest neighbor information \cite{merz2}.

Mutation is done by applying 2-Exchange operator. The 2-\textit{Exchange} operator breaks the 
tour by 2 edges and then rebuilds the path by adding 2 new edges (see \cite{krasnogor1}).

The parameters used by the LGP algorithm are given in Table \ref{lgp_ea8}.

\begin{table}[htbp]
\caption{The parameters of the LGP algorithm used for Experiment 5.}
\label{lgp_ea8}
\begin{center}
\begin{tabular}
{|p{140pt}|p{171pt}|}
\hline
\textbf{Parameter}& 
\textbf{Value} \\
\hline
Population size& 
500 \\
\hline
Code Length& 
80 instructions \\
\hline
Number of generations& 
50 \\
\hline
Crossover probability& 
0.7 \\
\hline
Crossover type& 
Uniform Crossover \\
\hline
Mutation & 
5 mutations per chromosome \\
\hline
Function set& 
$F$ = {\{}\textit{Select}, \textit{Crossover}, \textit{Mutate}{\}} \\
\hline
\end{tabular}
\end{center}
\end{table}

The parameters of the Evolved EA are given in Table \ref{lgp_ea9}.

\begin{table}[htbp]
\caption{The parameters of the evolved EA for TSP.}
\label{lgp_ea9}
\begin{center}
\begin{tabular}
{|p{140pt}|p{189pt}|}
\hline
\textbf{Parameter}& 
\textbf{Value} \\
\hline
Population size& 
40 \\
\hline
Number of generations& 
100 \\
\hline
Crossover probability& 
1 \\
\hline
Crossover type& 
DPX \\
\hline
Mutation & 
2-Exchange \\
\hline
Selection& 
Binary Tournament \\
\hline
\end{tabular}
\end{center}
\end{table}

For training and testing stages of our algorithm we use several problems 
from the TSPLIB \cite{prechelt1}. The \textit{att48} problem (containing 48 nodes) is used for 
training purposes and the \textit{berlin52} problem (containing 52 nodes) is used for 
validation purposes. Other 25 well-known TSP instances are used as the test set.

Five runs for evolving EAs were performed. A run took about one day on a 
PIII -600 MHz computer. In each run an EA yielding very good performance has 
been evolved. One of these EAs has been tested against other 25 difficult 
instances from TSPLIB. The results of the Evolved EA along with the results 
obtained with the standard GA described in section \ref{ea1} are given in Table 
\ref{lgp_ea10}. The standard GA uses a standard population of 40 individuals and an 
additional population (the new population) with 40 individuals.

\begin{table}[htbp]
\caption{The results of the standard GA and Evolved EA for 27 
instances from TSPLIB. \textit{Mean} stands for the mean over all runs and \textit{StdDev} stands for 
the standard deviation. The difference $\Delta $ is in percent and it is 
computed considering the values of the Evolved EA as a baseline. Results are 
averaged over 30 runs.}
\label{lgp_ea10}
\begin{center}
\begin{tabular}
{|p{60pt}|p{60pt}|p{60pt}|p{60pt}|p{60pt}|p{40pt}|}
\hline
\raisebox{-1.50ex}[0cm][0cm]{Problem}& 
\multicolumn{2}{|p{130pt}|}{\textbf{Standard GA} \par \textbf{40 individuals in the standard population + 40 individuals in the new population}} & 
\multicolumn{2}{|p{130pt}|}{\textbf{Evolved EA} \par \textbf{40 individuals}} & 
\raisebox{-1.50ex}[0cm][0cm]{$\Delta $} \\
\cline{2-5} 
 & 
\textit{Mean}& 
\textit{StdDev}& 
\textit{Mean}& 
\textit{StdDev}& 
  \\
\hline
a280& 
3291.18& 
39.25& 
3066.72& 
51.67& 
7.31 \\
\hline
att48& 
39512.65& 
883.47& 
36464.63& 
780.86& 
8.35 \\
\hline
berlin52& 
8872.41& 
145.00& 
8054.99& 
128.36& 
10.14 \\
\hline
bier127& 
129859.76& 
854.03& 
128603.46& 
1058.51& 
0.97 \\
\hline
ch130& 
7531.64& 
116.92& 
6818.78& 
142.05& 
10.45 \\
\hline
ch150& 
8087.08& 
181.14& 
7019.56& 
140.01& 
15.20 \\
\hline
d198& 
18592.67& 
291.75& 
17171.83& 
254.37& 
8.27 \\
\hline
d493& 
42846.82& 
686.30& 
40184.86& 
544.35& 
6.62 \\
\hline
d657& 
62348.86& 
364.76& 
58421.90& 
740.38& 
6.72 \\
\hline
eil101& 
806.03& 
17.61& 
734.62& 
8.60& 
9.72 \\
\hline
eil51& 
510.81& 
4.41& 
470.64& 
12.19& 
8.53 \\
\hline
eil76& 
677.55& 
26.34& 
599.71& 
11.46& 
12.97 \\
\hline
fl417& 
15287.26& 
159.86& 
14444.20& 
268.73& 
5.83 \\
\hline
gil262& 
2952.11& 
67.68& 
2746.59& 
53.71& 
7.48 \\
\hline
kroA100& 
25938.18& 
650.96& 
23916.58& 
529.01& 
8.45 \\
\hline
kroA150& 
33510.69& 
445.14& 
30650.92& 
558.66& 
9.33 \\
\hline
kroA200& 
35896.96& 
295.57& 
34150.88& 
814.83& 
5.11 \\
\hline
kroB100& 
27259.50& 
1295.30& 
23912.50& 
346.77& 
13.99 \\
\hline
kroB150& 
32602.75& 
590.64& 
29811.95& 
519.03& 
9.36 \\
\hline
kroC100& 
25990.92& 
453.61& 
22263.00& 
585.83& 
16.74 \\
\hline
kroD100& 
26454.58& 
864.43& 
24454.33& 
383.18& 
8.17 \\
\hline
kroE100& 
27126.75& 
667.92& 
24295.64& 
517.73& 
11.65 \\
\hline
lin105& 
19998.93& 
339.39& 
16573.09& 
528.26& 
20.67 \\
\hline
lin318& 
53525.55& 
976.88& 
49778.67& 
768.35& 
7.52 \\
\hline
p654& 
45830.71& 
384.92& 
41697.25& 
1356.95& 
9.91 \\
\hline
pcb442& 
60528.60& 
294.78& 
59188.30& 
677.04& 
2.26 \\
\hline
pr107& 
48438.22& 
476.81& 
46158.41& 
268.34& 
4.93 \\
\hline
\end{tabular}
\end{center}
\end{table}

Table \ref{lgp_ea10} shows that the Evolved EA performs better than the 
standard GA for all the considered test problems. The difference $\Delta $ 
ranges from 0.97 {\%} (for the problem \textit{bier127}) up to 20.67 {\%} (for the problem 
\textit{lin105}).

One can see that the standard GA performs very poor compared to other 
implementations found in literature \cite{merz2,krasnogor1}. This is due to the weak 
(non-elitist) evolutionary scheme employed in this experiment. The 
performance of the GA can be improved by preserving the best individual 
found so far. However, this is beyond the purpose of this research. Our main 
aim was to evolve an Evolutionary Algorithm and then to compare it with some 
similar (in terms of number of genetic operations performed) EA structures.\\

\textbf{Experiment 6}\\

In this experiment we use the EA (evolved for function optimization) to 
solve TSP instances. This transmutation is always possible since the evolved 
EA does not store any information about the problem being solved. Results 
are given in Table \ref{lgp_ea11}.

\begin{table}[htbp]
\caption{The results of the Evolved EA for function optimization 
(see Appendix 1) and Evolved EA for TSP (see Appendix 2) for 27 instances 
from TSPLIB. \textit{Mean} stands for the mean over all runs and \textit{StdDev} stands for the standard 
deviation. The difference $\Delta $ is in percent and it is computed 
considering the values of the Evolved EA for TSP as a baseline. Results are 
averaged over 30 runs.}
\label{lgp_ea11}
\begin{center}
\begin{tabular}
{|p{60pt}|p{60pt}|p{60pt}|p{60pt}|p{60pt}|p{40pt}|}
\hline
\raisebox{-1.50ex}[0cm][0cm]{Problem}& 
\multicolumn{2}{|p{130pt}|}{\textbf{Evolved EA for function optimization} \par \textbf{40 individuals}} & 
\multicolumn{2}{|p{130pt}|}{\textbf{Evolved EA for TSP} \par \textbf{40 individuals}} & 
\raisebox{-1.50ex}[0cm][0cm]{$\Delta $} \\
\cline{2-5} 
 & 
\textit{Mean}& 
\textit{StdDev}& 
\textit{Mean}& 
\textit{StdDev}& 
  \\
\hline
a280& 
3156.39& 
8.54& 
3066.72& 
51.67& 
2.92 \\
\hline
att48& 
39286.76& 
192.38& 
36464.63& 
780.86& 
7.73 \\
\hline
berlin52& 
8389.56& 
316.86& 
8054.99& 
128.36& 
4.15 \\
\hline
bier127& 
131069.06& 
1959.86& 
128603.46& 
1058.51& 
1.91 \\
\hline
ch130& 
7221.11& 
30.51& 
6818.78& 
142.05& 
5.90 \\
\hline
ch150& 
7094.49& 
13.57& 
7019.56& 
140.01& 
1.06 \\
\hline
d198& 
18001.13& 
81.42& 
17171.83& 
254.37& 
4.82 \\
\hline
d493& 
41837.06& 
492.34& 
40184.86& 
544.35& 
4.11 \\
\hline
d657& 
60844.49& 
336.61& 
58421.90& 
740.38& 
4.14 \\
\hline
eil101& 
742.77& 
5.92& 
734.62& 
8.60& 
1.10 \\
\hline
eil51& 
506.59& 
2.95& 
470.64& 
12.19& 
7.63 \\
\hline
eil76& 
615.68& 
6.69& 
599.71& 
11.46& 
2.66 \\
\hline
fl417& 
15284.92& 
64.54& 
14444.20& 
268.73& 
5.82 \\
\hline
gil262& 
2923.88& 
27.38& 
2746.59& 
53.71& 
6.45 \\
\hline
kroA100& 
25273.13& 
550.62& 
23916.58& 
529.01& 
5.67 \\
\hline
kroA150& 
31938.69& 
384.85& 
30650.92& 
558.66& 
4.20 \\
\hline
kroA200& 
35015.95& 
377.26& 
34150.88& 
814.83& 
2.53 \\
\hline
kroB100& 
25919.43& 
90.71& 
23912.50& 
346.77& 
8.39 \\
\hline
kroB150& 
31822.17& 
403.77& 
29811.95& 
519.03& 
6.74 \\
\hline
kroC100& 
23588.58& 
107.12& 
22263.00& 
585.83& 
5.95 \\
\hline
kroD100& 
25028.15& 
171.43& 
24454.33& 
383.18& 
2.34 \\
\hline
kroE100& 
25061.34& 
200.77& 
24295.64& 
517.73& 
3.15 \\
\hline
lin105& 
16977.1& 
49.02& 
16573.09& 
528.26& 
2.43 \\
\hline
lin318& 
50008.23& 
413.72& 
49778.67& 
768.35& 
0.46 \\
\hline
p654& 
43689.11& 
228.15& 
41697.25& 
1356.95& 
4.77 \\
\hline
pcb442& 
59825.62& 
292.70& 
59188.30& 
677.04& 
1.07 \\
\hline
pr107& 
47718.90& 
113.56& 
46158.41& 
268.34& 
3.38 \\
\hline
\end{tabular}
\end{center}
\end{table}

From Tables \ref{lgp_ea10} and \ref{lgp_ea11} it can be seen that the EA evolved for function 
optimization performs better than the standard GA but worse than the EA 
evolved for TSP. These results suggest that the structure of an evolutionary 
algorithm might depend on the problem being solved. This observation is in 
full concordance with the NFL theorems which tell us that we cannot obtain 
"the best" EA unless we embed some information about the problem being solved.

\section{Conclusions}

In this chapter, LGP and MEP have been used for evolving 
Evolutionary Algorithms. A detailed description of the proposed approaches has 
been given allowing researchers to apply the method for evolving 
Evolutionary Algorithms that could be used for solving problems in their 
fields of interest. 

The proposed model has been used for evolving Evolutionary Algorithms for function optimization and the Traveling Salesman Problem. Numerical experiments emphasize the robustness and the 
efficacy of this approach. The evolved Evolutionary Algorithms perform similar and sometimes even better than some standard approaches in the literature.

\chapter{Searching for a Practical Evidence of the No Free Lunch Theorems}\label{nfl_chap}

A framework for constructing test functions that match a given algorithm is developed in this chapter. More specific, given two algorithms $A$ and $B$, the question is which the functions for which $A$ performs better than $B$ (and 
vice-versa) are. For obtaining such functions we will use an evolutionary 
approach: the functions matched to a given algorithm are evolved by using the 
Genetic Programming (GP) \cite{koza1} technique.

This chapter is entirely original and it is based on the paper \cite{oltean_nfl,oltean_nfl_search}.

\section{Introduction}

Since the advent of No Free Lunch (NFL) theorems in 1995 \cite{wolpert1}, the 
trends of Evolutionary Computation (EC) \cite{goldberg1} have not changed at all, 
although these breakthrough theories should have produced dramatic changes. 
Most researchers chose to ignore NFL theorems: they developed new 
algorithms that work better than the old ones on some particular test 
problems. The researchers have eventually added: "The algorithm $X$ performs 
better than another algorithm \textit{on the considered test functions}". That is somehow useless since the proposed 
algorithms cannot be the best on all the considered test functions. 
Moreover, most of the functions employed for testing algorithms are 
artificially constructed. 

Consider for instance, the field of evolutionary single-criteria 
optimization where most of the algorithms were tested and compared on some 
artificially constructed test functions (most of them being known as De'Jong 
test problems) \cite{goldberg1,yao1}. These test problems were used for comparison 
purposes before the birth of the NFL theorems and they are used even today 
(8 years later after the birth of the NFL theorems). Evolutionary 
multi-criteria optimization was treated in a similar manner: most of the 
recent algorithms in this field were tested on several artificially 
constructed test functions proposed by K. Deb in \cite{deb1}.

Roughly speaking, the NFL theorems state that all the black-box optimization 
algorithms perform equally well over the entire set of optimization 
problems. Thus, if an algorithm $A$ is better than another algorithm $B$ on some 
classes of functions, the algorithm $B$ is better than $A$ on the rest of the 
functions.

As a consequence of the NFL theories, even a computer program (implementing 
an Evolutionary Algorithm (EA)) containing programming errors can perform 
better than some other highly tuned algorithms for some test functions.

Random search (RS) being a black box search / optimization algorithm 
should perform better than all of the other algorithms for some classes of 
test functions. Even if this statement is true, there is no result reported in 
the specialized literature of a test function for which RS performs better 
than all the other algorithms (taking into account the NFL restriction 
concerning the number of distinct solutions visited during the search). 
However, a function which is hard for all Evolutionary Algorithms is 
presented in \cite{droste1}.

Three questions (on how we match problems to algorithms) are of high 
interest:

For a given class of problems, which is (are) the algorithm(s) that performs 
(perform) better than all other algorithms?

For a given algorithm which is (are) the class(es) of problems for which the 
algorithm performs best? 

Given two algorithms $A$ and $B$, which is (are) the class (es) of problems for 
which $A$ performs better than $B$?

Answering these questions is not an easy task. All these problems are still 
open questions and they probably lie in the class of the NP-Complete 
problems. If this assumption is true it means that we do not know if we are 
able to construct a polynomial algorithm that takes a function as input and 
outputs the best optimization algorithm for that function (and vice versa). 
Fortunately, we can try to develop a heuristic algorithm able to handle this 
problem.

\section{Basics on No Free Lunch Theorems}

The results provided by the No Free Lunch Theorems are divided in two main classes: No Free Lunch Theorems for Oprimization and No Free Lunch Theorems for Search.

Roughly speaking, the NFL theorems for Optimization \cite{wolpert2} state that all the black-box optimization 
algorithms perform equally well over the entire set of optimization 
problems. 

Thus, if an algorithm $A$ is better than another algorithm $B$ on some 
classes of functions, the algorithm $B$ is better than $A$ on the rest of the 
functions.

The NFL theorems for Search \cite{wolpert1} state that we cannot use the information about the algorithm' behaviour so far to predict it's future behaviour.

\section{A NFL-style algorithm}\label{nfl2}

Firstly, we define a black-box optimization algorithm as indicated by 
Wolpert and McReady in \cite{wolpert1}.

The evolutionary model (the NFL-style algorithm) employed in this study 
uses a population consisting of a single individual. This considerably simplify the description and the implementation of a NFL-style algorithm. No archive for storing the best solutions found so far (see for 
instance \textit{Pareto Archived Evolution Strategy} \cite{knowles1}) is maintained. However, we implicitly maintain an archive containing all the distinct solutions explored until the current state. We 
do so because only the number of distinct solutions is counted in the NFL 
theories. This kind of archive is also employed by Tabu Search \cite{glover1,glover2}.

The variables and the parameters used by a NFL algorithm are given in Table 
\ref{NFL1}.

\begin{table}[htbp]
\caption{The variables used by the NFL algorithm.}
\label{NFL1}
\begin{center}
\begin{tabular}
{|p{69pt}|p{200pt}|}
\hline
\textbf{Variable}& 
\textbf{Meaning} \\
\hline
\textit{Archive}& 
the archive storing all distinct solutions visited by algorithm \\
\hline
\textit{curr{\_}sol}& 
the current solution (point in the search space) \\
\hline
\textit{new{\_}sol}& 
a new solution (obtained either by mutation or by initialization) \\
\hline
\textit{MAX{\_}STpng}& 
the number of generations (the number of distinct points in the search space visited by the algorithm). \\
\hline
$t$& 
the number of distinct solutions explored so far \\
\hline
\end{tabular}
\end{center}
\end{table}

The NFL-style algorithm is outlined below:

\begin{center}
\textbf{NFL Algorithm}
\end{center}

\textsf{\textit{S}}$_{1}$\textsf{. Archive = $\emptyset $;}

\textsf{\textit{S}}$_{2}$\textsf{. Randomly initializes the current solution 
(curr{\_}sol)}

\textsf{// add the current solution to the archive}

\textsf{\textit{S}}$_{3}$\textsf{. Archive = Archive + {\{}curr{\_}sol{\}}; 
}

\textsf{\textit{S}}$_{4}$\textsf{. t = 1;}

\textsf{\textit{S}}$_{5}$\textsf{. }\textsf{\textbf{while}}\textsf{ t $<$ 
MAX{\_}STpng }\textsf{\textbf{do}}\textsf{ }

\textsf{\textit{S}}$_{6}$\textsf{. Select a new solution (new{\_}sol) in }

\textsf{the neighborhood of the curr{\_}sol}

\textsf{\textit{S}}$_{7}$\textsf{. Archive = Archive + {\{}new{\_}sol{\}};}

\textsf{\textit{S}}$_{8}$\textsf{. curr{\_}sol = new{\_}sol;}

\textsf{\textit{S}}$_{9}$\textsf{. t = t + 1;}

\textsf{\textit{S}}$_{10}$\textsf{. }\textsf{\textbf{endwhile}}\\

An important issue concerning the NFL algorithm described above is related 
to the step $S_{6}$ which selects a new solution that does not belong to the 
\textit{Archive}. This is usually done by mutating the current solution and keeping the 
offspring if the latter does not already belong to the \textit{Archive} (The actual 
acceptance mechanism is minutely described in section \ref{nfl1}). If the offspring 
belongs to the \textit{Archive} for a fixed number of mutations (stpng) it means that the 
neighborhood of the current solutions could be exhausted (completely 
explored). In this case, a new random solution is generated and the search 
process moves to another region of the search space. 

It is sometimes possible the generated solution to already belong to the \textit{Archive}. In this case, 
another random solution is generated over the search space. We assume that 
the search space is large enough and after a finite number of 
re-initializations the generated solution will not belong to the \textit{Archive}.

The algorithm for selecting a new solution which does not belong to the 
\textit{Archive} (the step $S_{6})$ is given below:\\

\textsf{\textit{SS}}$_{1}$\textsf{. nr{\_}mut = 0; // the number of mutations is set to 0}

\textsf{\textit{SS}}$_{2}$\textsf{. }\textsf{\textbf{Repeat}}

\textsf{\textit{SS}}$_{3}$\textsf{. new{\_}sol = 
}\textsf{\textit{Mutate}}\textsf{ (curr{\_}sol);}

\textsf{\textit{SS}}$_{4}$\textsf{. nr{\_}mut = nr{\_}mut + 1;}

\textsf{\textit{SS}}$_{5}$\textsf{. }\textsf{\textbf{until}}\textsf{ 
(nr{\_}mut = MAX{\_}MUTATIONS) and (new{\_}sol $ \notin 
$}\textsf{\textit{Archive}}\textsf{) }\textsf{\textbf{and}}\textsf{ 
}\textsf{\textit{Accepted}}\textsf{(new{\_}sol);}

\textsf{\textit{SS}}$_{6}$\textsf{. }\textsf{\textbf{while}}\textsf{ 
new{\_}sol $ \notin $Archive }\textsf{\textbf{do}}

\textsf{\textit{SS}}$_{7}$\textsf{. 
}\textsf{\textit{Initialize}}\textsf{(new{\_}sol); //we jump into another 
randomly chosen point of the search space}

\textsf{\textit{SS}}$_{8}$\textsf{. }\textsf{\textbf{endwhile}}\\

\section{Evolutionary Model and the Fitness Assignment Process}

Our aim is to find a test function for which a given algorithm $A$ performs 
better than another given algorithm $B$. The test function that is being 
searched for will be evolved by using Genetic Programming \cite{koza1} with steady 
state \cite{syswerda1}.

The quality of the test function encoded in a GP chromosome is computed in a 
standard manner. The given algorithms $A$ and $B$ are applied to the test 
function. These algorithms will try to optimize (find the minimal value of) 
that test function. To avoid the lucky guesses of the optimal point, each 
algorithm is run 500 times and the results are averaged. Then, the fitness 
of a GP chromosome is computed as the difference between the averaged 
results of the algorithm $A$ and the averaged results of the algorithm $B$. In the 
case of function minimization, a negative fitness of a GP chromosome means 
that the algorithm $A$ performs better than the algorithm $B$ (the values obtained 
by $A$ are smaller (on average) than those obtained by $B)$.

\section{Algorithms used for comparison}\label{nfl1}

We describe several evolutionary algorithms used for comparison purposes. 
All the algorithms described in this section are embedded in the NFL 
algorithm described in section \ref{nfl2}. More precisely, the considered algorithms 
particularize the solution representation, the mutation operator, and the 
acceptance mechanism (the procedure \textit{Accepted}) of the NFL algorithm described in 
section \ref{nfl2}. The mutation operator is the only search operator used for 
exploring the neighborhood of a point in the search space.\\

\textbf{\textit{A}}$_{1}$ - real encoding (the individuals are represented 
as real numbers using 32 bits), Gaussian mutation with $\sigma _{1}$ = 
0.001, the parent and the offspring compete for survival.\\

\textbf{\textit{A}}$_{2}$ - real encoding (the individuals are represented 
as real numbers using 32 bits), Gaussian mutation with $\sigma _{2}$ = 
0.01, the parent and the offspring compete for survival.\\

\textbf{\textit{A}}$_{3}$ - binary encoding (the individuals are 
represented as binary strings of 32 bits), point mutation with $p_{m}$ = 0.3, 
the parent and the offspring compete for survival.\\

\textbf{\textit{A}}$_{4}$ - binary encoding (the individuals are 
represented as binary strings of 32 bits), point mutation with $p_{m}$ = 0.1, 
the parent and the offspring compete for survival.

\section{Numerical experiments}

Several numerical experiments for evolving functions matched to a given 
algorithm are performed in this section. The algorithms used for comparison 
have been described in section \ref{nfl1}.

The number of dimensions of the space is set to 1 (i.e. one-dimensional 
functions) and the definition domain of the evolved test functions is [0, 
1].

The parameters of the GP algorithm are given in Table \ref{NFL2}.

\begin{table}[htbp]
\caption{The parameters of the GP algorithm used for numerical experiments}
\label{NFL2}
\begin{center}
\begin{tabular}
{|p{140pt}|p{135pt}|}
\hline
\textbf{Parameter}& 
\textbf{Value} \\
\hline
Population size& 
50 \\
\hline
Number of generations& 
10 \\
\hline
Maximal GP tree depth& 
6 \\
\hline
Function set& 
$F$ = {\{}+, -, *, \textit{sin}, \textit{exp}{\}} \\
\hline
Terminal set& 
$T$ = {\{}$x${\}} \\
\hline
Crossover probability& 
0.9 \\
\hline
Mutation& 
1 mutation / chromosome \\
\hline
Runs& 
30 \\
\hline
\end{tabular}
\end{center}
\end{table}

The small number of generations (only 10) has been proved to be sufficient for 
the experiments performed in this study.

Evolved functions are given in Table \ref{NFL3}. For each pair ($A_{k}$, $A_{j})$ is 
given the evolved test function for which the algorithm $A_{k}$ performs 
better than the algorithm $A_{j}$. The mean of the fitness of the best GP 
individual over 30 runs is also reported.

\begin{table}[htbp]
\caption{The evolved test functions.}
\label{NFL3}
\begin{center}
\begin{tabular}
{|p{64pt}|p{124pt}|p{98pt}|}
\hline
\textbf{Algorithms}& 
\textbf{Evolved Test Function}& 
\textbf{Averaged fitness} \\
\hline
($A_{1}$, $A_{2})$& 
$f_1 (x) = 0.$& 
0 \\
\hline
($A_{2}$, $A_{1})$& 
$f_2 (x) = - 6x^3 - x.$& 
-806.03 \\
\hline
($A_{3}$, $A_{4})$& 
$f_3 (x) = x - 2x^5.$& 
-58.22 \\
\hline
($A_{4}$, $A_{3})$& 
$f_4 (x) = - 4x^8.$& 
-34.09 \\
\hline
($A_{2}$, $A_{4})$& 
$f_5 (x) = 0.$& 
0 \\
\hline
($A_{4}$, $A_{2})$& 
$f_6 (x) = - 6x^3 - x.$& 
-1601.36 \\
\hline
\end{tabular}
\end{center}
\end{table}

Table \ref{NFL3} shows that the proposed approach made possible the 
evolving of test functions matched to the most of the given algorithms. The 
results of these experiments give a first impression of how difficult the 
problems are. Several interesting observations can be made:

The GP algorithm was able to evolve a function for which the algorithm 
$A_{2}$ (real encoding with $\sigma $ = 0.01) was better then the algorithm 
$A_{1}$ (real encoding with $\sigma $ = 0.001) in all the runs (30). However, 
the GP algorithm was not able to evolve a test function for which the 
algorithm $A_{1}$ is better that the algorithm $A_{2}$. In this case the 
function $f(x)$ = 0 (where both algorithms perform the same) was the only one to 
be found. It seems to be easier to find a function for which an algorithm 
with larger "jumps" is better than an algorithm with smaller "jumps" 
than to find a function for which an algorithm with smaller "jumps" is 
better than an algorithm with larger "jumps".

A test function for which the algorithm $A_{4}$ (binary encoding) is better 
than the algorithm $A_{2}$ (real encoding) was easy to find. The reverse 
(i.e. a test function for which the real encoding algorithm $A_{2}$ is better 
than the binary encoded algorithm $A_{4})$ has not been found by using the GP 
parameters considered in Table \ref{NFL2}.

\section{Conclusions and further work}

In this paper, a framework for evolving test functions that are matched to a 
given algorithm has been proposed. The proposed framework is intended to 
provide a practical evidence for the NFL theories. Numerical experiments 
have shown the efficacy of the proposed approach.

Further research will be focused on the following directions:

\begin{itemize}
\item[{\it (i)}] extending the function set ($F)$ and the number of space dimensions.

\item[{\it (ii)}] comparing other evolutionary algorithms for single and multiobjective 
optimization. Several test functions matched to some classical algorithms 
(such as standard GAs or ES) for function optimization will be evolved. In 
this case the problem is much more difficult since the number of distinct 
solutions visited during the search process could be different for each 
algorithm.

\item[{\it (iii)}] evolving test instances for algorithms used for solving other real-world 
problems (such as TSP, symbolic regression, classification etc).

\item[{\it (iv)}] finding test functions for which random search is better than other 
algorithms.

\item[{\it (v)}] finding the set of test functions for which an algorithm is better than the other.

\end{itemize}

\chapter{Conclusions and further work}

Three new techniques have been proposed during the course of this thesis: Multi Expression Programming, Infix Form Genetic Programming, and Traceless Genetic Programming.

Proposed techniques have been used in the following applications: symbolic regression, classification, designing digital circuits, evolving play strategies, evolving heuristics for NP-complete problems, evolving evolutionary algorithms, and evolving test problems.

Further work will be focused on developing new GP techniques, Designing digital circuits for reversible computers, evolving fast winning strategies for the end of the chess and othello games, evolving EAs with patterns, evolving heuristics for other NP-complete problems, applying the existing GP techniques to real-world prediction data, etc.

\end{document}